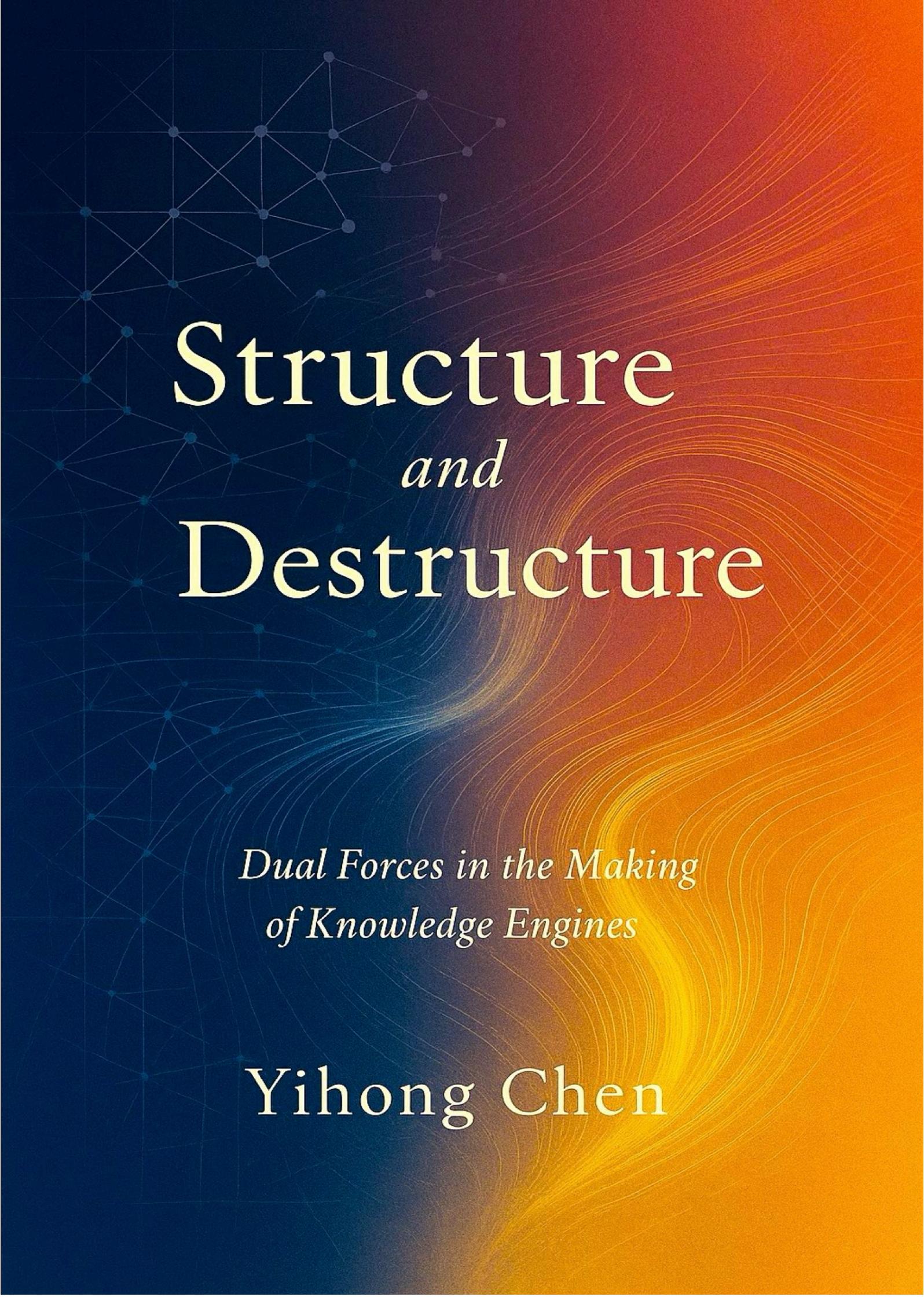

# Structure and Destructure

## Dual Forces in the Making of Knowledge Engines

### Yihong Chen

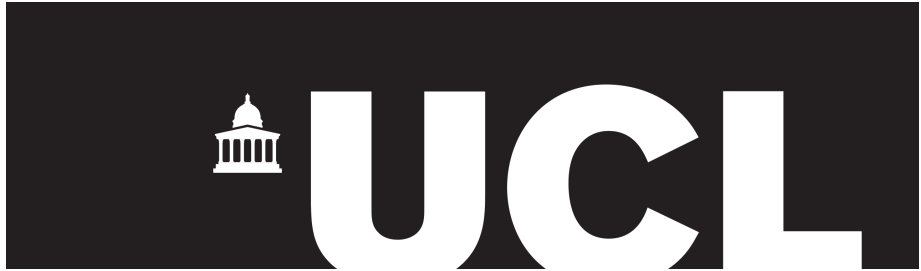

# Structure and Destructure
## Dual Forces in the Making of Knowledge Engines

**YIHONG CHEN**

UNIVERSITY COLLEGE LONDON

COMPUTER SCIENCE

Submitted to University College London (UCL) in partial fulfilment of the requirements for the award of the degree of Doctor of Philosophy.

Primary supervisor: Pontus Stenetorp
Secondary supervisor: Sebastian Riedel
Examining committee: John Shawe-Taylor, Mathias Niepert

# Declaration

I, Yihong Chen, confirm that the work presented in this thesis is my own. Where information has been derived from other sources, I confirm that this has been indicated in the thesis.

<div style="text-align: right;">
Yihong Chen  
London, United Kingdom  
December 2024
</div>

# Abstract

The making of knowledge engines in natural language processing has been shaped by two seemingly distinct paradigms: one grounded in structure, the other driven by massively available unstructured data. The structured paradigm leverages predefined symbolic interactions–such as knowledge graphs–as priors, and designs models to capture such priors. In contrast, the unstructured paradigm centres on scaling transformer architectures with increasingly vast data and model sizes, as seen in modern large language models. Despite their divergence, this thesis seeks to establish conceptual connections that bridge these two paradigms.

Two key connections are identified:

- **Structure Formation**: Self-supervised objectives, such as language modelling, induce structural patterns in model computation across both paradigms. These objectives support data graph reconstruction, facilitating link prediction in the structured paradigm and providing interpretability in the unstructured paradigm through extracted n-gram patterns.

- **Destructure for Plasticity**: Embeddings, a critical yet often overlooked component in both paradigms, cache message-passing computations over symbols during training. However, excessive caching can hinder generalization. *Embedding forgetting*, defined as the periodic reset of embedding weights, improves model plasticity and enables generalization to previously unseen scenarios, such as novel predicates or languages.

These connections form a new recipe for developing general knowledge engines, where the guidelines not only include modelling of the *seen* symbolic interactions but also modelling of the *unseen*, the latter being relatively underexplored. Efficiently mod-elling the seen necessitates structure formation, regardless of whether the data is inherently structured or not. Conversely, modelling the unseen benefits from active destructuring of the learned cache, which promotes robustness and adaptability.

By bridging the two paradigms, this thesis establishes **structure** and **destructure** as complementary forces in the design of knowledge engines that can support transparent, controllable, and adaptable intelligent systems.

# Impact Statement

Artificial intelligence (AI) systems are becoming deeply embedded in our life, as tools or companions. They shape how we search for information, receive recommendations, and interact with people across cultures and languages. These fundamentally change how we perceive the world, and acquire relevant knowledge for navigating in the world. Yet most AI models today still struggle to connect structured data (such as databases or knowledge graphs) with the unstructured natural language used in real-world communication. This thesis bridges structured and unstructured paradigms, providing a conceptual framework for developing general knowledge engines that back adaptable and controllable AI agents. Its impact is listed as follows.

**Academic Impact.** The thesis has introduced new methods, such as reinterpreting embeddings as message-passing caches and proposing active forgetting mechanisms. These methods enable models to adapt to unseen knowledge graphs and novel languages, enhancing generalization across domains. This influences various research communities:

- **Knowledge Base Completion:** The GitHub repository accompanying Chapter 2 has garnered over 100 stars, demonstrating its adoption by the community.

- **Language Model Interpretability:** Techniques for extracting n-gram patterns from transformers (Chapter 3) offer tools for auditing large-scale models, addressing critical needs for transparency and safety in AI.

- **Cognitive Science:** Chapter 5 also provides insights for studying human language acquisition for example the critical period [Constantinescu et al., 2024].

**Industry Impact.** Methods developed in the thesis can address practical challenges faced by the AI industry:

- **Recommender Systems and Search Engines:** Techniques for bridging structured and unstructured data integration (Chapter 2 and Chapter 3) are relevant for improving algorithms used by companies like Google and TikTok.

- **Language Model Plasticity:** Active forgetting enables more adaptable AI systems, which is critical for businesses operating in multilingual or dynamic environments. Microsoft [Aggarwal et al., 2024] applies active forgetting to GPT style models and Cambridge researchers [Zhao et al., 2024b, Iacob et al., 2024] explores forgetting inspired decoupled learning to privacy-preserving scenarios.

**Societal Impact.** The thesis has several societal implications:

- **Cognitive and Privacy Research:** Active forgetting mechanisms resonate with principles in cognitive science and privacy, informing discussions on memory management and data retention in AI systems.

- **AI Education and Policy:** Insights into model transparency contribute to designing AI systems that align with ethical standards and public accountability.

By bridging structured and unstructured paradigms, this thesis provides a pathway toward developing intelligent systems that are robust, transparent, and aligned with human values. These systems hold the potential to revolutionize how we interact with and benefit from AI across domains.

# Acknowledgements

This thesis is a small vessel that has carried many questions, friendships, frustrations, and unexpected joys. It is sewed together with codebases, papers, whiteboards, long walks, and quiet moments by the window. My deepest gratitude goes to everyone who has walked alongside me, for a step, or for miles.

To Sebastian, thank you for encouraging intuition and for guiding me toward a way of thinking grounded in gut feeling. Your support gave shape to my research before I knew what I was searching for. To Pontus, thank you for our pair-writing sessions, your curiosity across disciplines, and for showing me how to learn like a language model in a purely unstructured way. You fed my curiosity and never tried to tame it. To Pasquale, thank you for introducing me to knowledge graphs and sharing your enthusiasm for factorization models. To Luca, thank you for your clarity, humour, and for showing that rigorous thinking need not be rigid.

To Pushkar, Mikel, Kelly, and Patrick, thank you for your expertise and collaboration in multilingual and graph-based research. Mikel, your elegant experimental design and intuition often made things fall into place. To David, Jiayi, Eduardo, thank you for opening doors to low-resource translation, and to Roberta, thank you for your creativity and steady presence in the forgetting project. To Lena, Javier, Nicola, Karen, Igor, Naila, Shalini, Chunan, Lilian, Brooks, Wanru, Xinchi, William, Nicolas, Lisa, Giulia, and the many collaborators who shaped different parts of this journey, thank you for your conversations and your curiosity. To my master's students, Junyu, Ravi, Jie, Keyue, Hubert, Keenan, and Passawis, thank you for your energy, ideas, and trust. To Linqing, YingChen, Minqi, Mikayel, Nikita, Sten, Yali, and Revadee, thank you for meals, resilience, reflections, and late-pandemic companionship. To Shifu Leo and Fang Yuan, thank you for teaching me that the body holds its own intelligence and balance.

To the landscapes that held me, Regent's Canal, Armshire, Alicante, Brechin High-

lands, Xuanwu Lake, Jinshan, and Pingshan, thank you. To UCL FAI CDT, Meta FAIR, and ELLIS, thank you for the space to explore, fail, and grow.

To my family: Mom, thank you for your fierce resilience and gentle soul. Dad, thank you for walking beside me when I faltered. Yiyun, thank you for being my closest companion, from childhood to PhD. Yizhi, thank you for grounding me in the physical world and inspiring me with your circuits or carving. And to Bocan, your presence and insight into the inner path have been lovely transformative.

Finally, I would like to thank the examiners of this thesis, John and Mathias, for their thoughtful and detailed feedback on the chapters. Your input was invaluable in bringing this work to completion.

# Contents

# List of Figures

# List of Tables

# List of Algorithms

# Opening

# Chapter 1

# Introduction

## 1.1 Building General Knowledge Engines

Humans have long been captivated by the pursuit of intelligence: seeking to understand its emergence, improve it through training, slow its decline over time, and ultimately replicate it in machines. This endeavour is driven by a desire to extend our innate cognitive abilities across time and space, aiming to achieve more efficient and effective use of our intellectual resources – much like how the Industrial Revolution transformed our ability to automate and amplify our physical capabilities.

One of the defining characteristics of intelligence is its ability to process and manage knowledge about our realities. The human mind, as the faculty of intelligence, can function as a general knowledge engine, capable of acquiring information from diverse sources, consolidating it through abstraction, retrieving it for reasoning on relevant tasks, and updating it to address evolving environments. This knowledge engine supports us across a wide spectrum of tasks, ranging from routine activities – such as navigating daily commutes, managing personal schedules, or cooking meals – to complex decision-making, like formulating trading strategies, resolving political conflicts, diagnosing medical conditions, or writing a PhD thesis.

When developing artificial intelligence (AI), particularly with the aim of emulating human intelligence, replicating general knowledge engines becomes crucial. These knowledge engines can serve as the backbone for many of our most impactful digital infrastructure today, such as search engines, recommender systems, and conversational agents (e.g., virtual assistants and chatbots), supporting our daily digital activities, as depicted by Figure 1.1. However, building general knowledge engines is not an easy task. In fact, it has been a complicated subject and the focus of many areas of studies, spanning disciplines such as natural language processing, information retrieval, data mining, machine learning, and cognitive science. Profoundly, a core challenge lies in integrating *diverse* knowledge sources and updating them in *real time*.

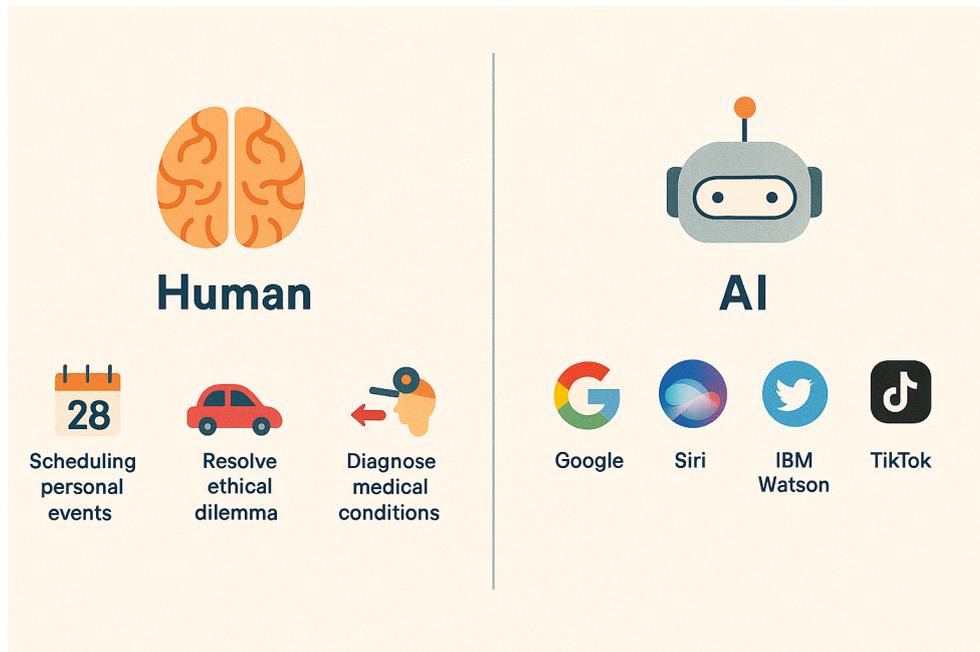

Figure 1.1: Illustration of how "knowledge engines" in human minds facilitate diverse human activities and how current digital knowledge engines underpin applications such as digital assistants, social media platforms, and recommendation systems.

To better understand this challenge, let us consider a concrete example: the development of an AI doctor designed to mimic a human physician. We can begin by examining the steps a human physician undergoes to acquire the necessary knowledge and skills.

> **Example: The Training of a Medical Doctor**
>
> **Consider Tom, a medical student, who progresses through various stages of learning to become a proficient doctor:**
>
> 1. **Childhood Curiosity:** As a child, Tom was attracted by the wonders of nature and the human body. His fascination deepened through stories shared by his grandfather, a seasoned doctor, who instilled in him a passion for healing.
>
> 2. **Formal Education:** In his school years, Tom immerses himself in medical textbooks, which provide organized and systematic knowledge in areas such as *biology, chemistry, anatomy, pathology, and pharmacology*. These resources act as the foundation of his medical expertise, enabling him to build clear connections between key concepts in the healthcare domain, forming structured knowledge that he can repeatedly use in his later profession life.
>
> 3. **Clinical Rotations:** During his clinical rotations, Tom observes senior doctors at work, engages in discussions about complex patient cases, and analyses unstructured clinical notes. These hands-on experiences and potentially unspoken knowledge teach him how to think critically about patient symptoms and interpret subtle contextual relationships among them.

We can see that Tom's mind operates as a knowledge engine, seamlessly blending structured knowledge sources (e.g., *drug-drug interactions*) for accurate recall with unstructured insights (e.g., *holistic symptom assessment notes*) to guide informed clinical decision-making. On the other hand, his natural curiosity, a form of open mindsets, continuously seeds the drive to refine, update, and expand his knowledge, ensuring that it evolves with the changing medical landscape. Similarly, an AI system aspiring to mimic such medical expertise must have a knowledge engine that can leverage both *structured* and *unstructured* sources to acquire, consolidate, apply, and update knowledge dynamically.

This thesis presents a scientific exploration aimed at understanding the approaches to develop knowledge engines for AI agents and how these seemingly disparate approaches can be unified into a framework for creating more general knowledge engines that can adapt to previously unseen environments. At a high level, there are primarily two exist-ing paradigms for building general knowledge engines, the **structured paradigm** and the **unstructured paradigm**, as detailed in Section 1.2. However, the dichotomy between these approaches diminishes, upon closer examination of their internal mechanisms during training and inference, as well as their shared limitations in generalizing to new, unseen environments. This convergence suggests a unified, integrated pathway for constructing general knowledge engines.

The remainder of this chapter will outline the motivation and context for such unification and integration (Section 1.2), the research objectives and questions (Section 1.3), a brief overview of the methodology (Section 1.4), and a roadmap of the thesis structure (Section 1.5).

## 1.2 The Dichotomy: Structured vs. Unstructured

The majority of human knowledge sources can be categorized into two forms: the *structured* and the *unstructured*. Historically, research on processing these two forms of knowledge for AI systems has largely been studied in separate streams.

The earlier waves of AI features expert systems proliferated in the 1980s [Hayes-Roth et al., 1983]. Expert systems were heavily backed by structured knowledge sources, such as curated knowledge graphs specifying relationships among entities. In contrast, contemporary AI advancements increasingly favour massive unstructured datasets – for instance web data – as the foundation for building state-of-the-art AI.

In this thesis, we will refer to these two paradigms as the structured paradigm and unstructured paradigm. We note that the transition from the structured data to unstructured data is not a binary division but rather along a spectrum of relative structuring. For example, from the grammar perspective, coding data is more semi-structured compared to natural language data; from the conceptual ogranisation perspective, textbook data is more structured and organized compared to texts coming from the internet. While acknowledging these intermediate forms, this thesis seeks to examine the archetypal structured and unstructured paradigms, as presented below.

## 1.2.1 The Structured Paradigm for Building Knowledge Engines: Exemplified by Knowledge Graphs

Structures are fundamentally about how different parts relate to each other and how they assemble to represent realities – whether physical or virtual. These structures are essential for humans to organize and understand the world around us. Particularly, our world is full of physical structures, such as molecular networks, protein folding patterns, and transportation routes. In this sense, structures allow us to efficiently *categorize* and *underpin* various manifestations of the physical world. On the other hand, structures can also be abstract or virtual, like social interactions, the laws governing rational reasoning or the hierarchical relationships among words. These types of structures help us *systematize* our understanding of abstract concepts and connections.

In the history of AI, structured knowledge sources have aimed to organize such information in predefined formats, such as knowledge graphs, databases, and other relational structures [Wang et al., 2017]. In these formats, symbols are arranged in fixed-length sequences governed by specific grammar, where each position holds a defined role. For instance, in a knowledge graph, a knowledge triplet consists of three components: the first position typically denotes the subject (or head entity), the second represents the predicate (or relation), and the third position corresponds to the object (or tail entity)[1]. To illustrate this, consider the following diagram of a knowledge triplet:

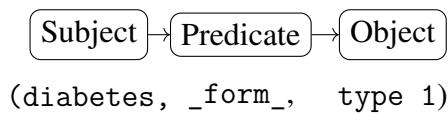

```
(diabetes, _form_,  type 1)
```

Where in this diagram:

- The **Subject** (or head entity) is `diabetes`.

- The **Predicate** (or relation) is `_form_`.

- The **Object** (or tail entity) is `type 1`.

A collection of such knowledge triples forms a knowledge graph. For example, the diagram in Figure 1.2 illustrates a portion of a widely used healthcare knowledge graph, SNOMED-CT, which is detailed in [Donnelly, 2006].

---

[1] In some cases, a relation defines a set of ordered pairs between subjects and objects.

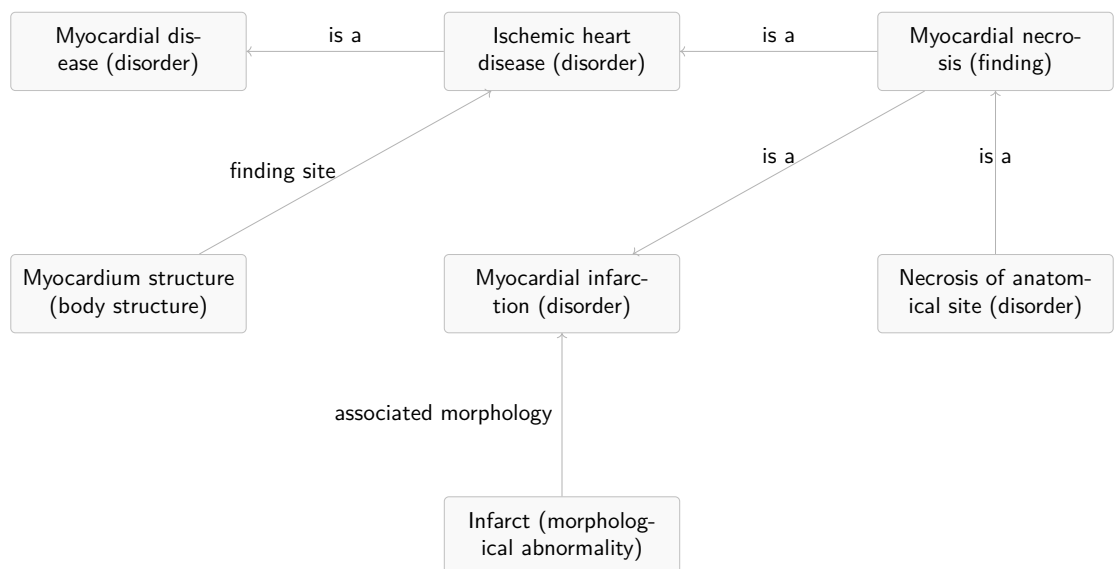

Figure 1.2: A medical knowledge graph showing relationships between myocardial diseases and associated conditions. The triples in the knowledge graph is drawn from SNOMED2Vec [Agarwal et al., 2019].

The structured paradigm is built around two key elements: data format and structural representation learning. Structured knowledge is typically represented through formats like multidimensional arrays, sparse graphs, or triplet databases, which allow for the explicit depiction of relationships and enable the analysis of logical properties such as transitivity, reflexivity, and antisymmetry. Representation learning in this context focuses on embedding these structures into model computations using approaches like factorization models (FMs) [Yang et al., 2016, Lacroix et al., 2018, Trouillon et al., 2016] and message-passing graph neural networks (GNNs) [Schlichtkrull et al., 2018, Vashishth et al., 2020, Zhu et al., 2021]. These models play a crucial role in both the automated construction of large-scale structured knowledge bases and in powering downstream tasks like question answering.

Knowledge engines built on structured paradigms excel in applications that require interpretability, consistency, and efficient reasoning. For example, they play a central role in serving as world models, which aim to represent reality comprehensively [LeCun, 2022]. Knowledge graphs, in particular, have been applied in a variety of domains, including commonsense reasoning [Hwang et al., 2021], digital twins [Akroyd et al., 2021], and text-based games [Ammanabrolu and Riedl, 2021]. These structured models also power some of the most widely used digital applications, such as:

- **Knowledge Bases**: Essential to expert systems (e.g., IBM Watson Medical).

- **Search Engines**: Enabling tools like Google Search.

- **Recommender Systems**: Underpinning platforms like YouTube.

- **Social Media**: Enhancing features on platforms like X.com and Instagram.

- **Intelligent Assistants**: Backing intelligent systems on edge devices like Siri.

### 1.2.2 The Unstructured Paradigm for Building Knowledge Engines: Exemplified by Pretrained Language Models

The latest wave of artificial intelligence, particularly generative AI, marks a significant shift toward an unstructured paradigm, exemplified by large language models. These models ingest vast amounts of unstructured text, moving away from the traditional reliance on structured knowledge sources. This paradigm shift was made possible by the Transformer architecture, which demonstrated that pretraining on large-scale unstructured datasets could lead to the generation of foundational representations [Devlin et al., 2019, Radford et al., 2019, Brown et al., 2020].

Following the advent of Transformer models, most algorithmic advancements have focused on improving computational efficiency, with an increasing emphasis on scaling model size and dataset diversity, rather than the structural intricacies of data or model architecture [Kaplan et al., 2020, Hernandez et al., 2021, Templeton et al., 2024]. The importance of preparing structured knowledge has diminished due to its high cost and complexity. In contrast, the process of crawling the web for diverse unstructured data has become a far more accessible and scalable alternative.

Unstructured data, in contrast to structured data, exists in free forms where the position of symbols within a sequence does not inherently define their role. For instance, in a sentence, the first word is not necessarily the subject, nor the last word the object. This type of knowledge is commonly referred to as corpus, corpora, or text, and is typically represented as sequences of variable lengths. Notable sources for pretraining large language models include:

- **Web Text**: One of the most commonly used web datasets is Common Crawl's petabyte-scale archive of web data since 2008 [Crawl, 2023]. Other similar datasets include CC100 [Conneau et al., 2020], OpenWebText [Contributors, 2019], and RedPajama [Computer, 2023].

- **Web Code Data**: Datasets like Starcoder [Project, 2023], which scrape repositories from GitHub and Stack Overflow.

- **High-Quality Referential Sources**: PeS2o [Soldaini and Lo, 2023] for academic data from Semantic Scholar, Project Gutenberg [Hart and Volunteers, 1971–2024] for books, and Wikipedia [authors, 2024] for encyclopedic knowledge.

The unstructured paradigm facilitates the development of large-scale language models that serve as alternative knowledge engines. These models are increasingly recognized as world models [Petroni et al., 2019, Li et al., 2021a, Hernandez et al., 2023], demonstrating exceptional performance in domains where structured data is sparse or unavailable. By processing unstructured data, these models have been shown to capture implicit relationships and context, enabling a broad range of capabilities, from answering questions to powering conversational AI systems like ChatGPT.

### 1.2.3 Comparing The Two Paradigms

The structured and unstructured paradigms of knowledge representation exhibit distinct features, as summarized in Table 1.1. Therefore, they also have different advantages and disadvantages as summarized by Table 1.2.

The structured paradigm offers significant *efficiency* benefits. It allows repetitive reuse of structured data, eliminating the need to compute solutions from scratch for recurring tasks. It also provides stable and consistent computational outcomes, particularly for logical reasoning tasks, such as deduction within knowledge graphs. Despite these benefits, structured paradigms face flexibility limitations. Particularly, structures can be restrictive, unable to fully accommodate the nearly infinite variability of real-world phenomena and vulnerable to missing entries.

The unstructured paradigm excels in its *flexibility*. It can represent and learn from diverse, unstructured data sources, capturing nuances that structured systems might miss.

The unstructured paradigm is particularly effective for tasks requiring generative capabilities, such as answering diverse questions flexibly or producing cartoon images based on given keywords. However, they have notable drawbacks: i) learning from unstructured data often requires starting from scratch, incurring high computational costs. ii) model generations can be hard to control, potentially containing biased or toxic content. iii) due to the black-box nature of end-to-end neural architectures commonly used in this paradigm, model generations are difficult to interpret and model internal mechanisms are less transparent to even their developers.

Table 1.1: Key distinctions between structured and unstructured paradigms in terms of data format, architecture, and learning objective.

|  | **Structured Paradigm** | **Unstructured Paradigm** |
| --- | --- | --- |
| **Data Format** | Knowledge Graphs | Free-form text |
| **Architecture** | FMs, GNNs | Transformer |
| **Learning Objective** | Entity Prediction | Language Modelling |

Table 1.2: Comparison of pros and cons between structured and unstructured paradigms for building knowledge engines.

|  | **Structured Paradigm** | **Unstructured Paradigm** |
| --- | --- | --- |
| **Pros** | • Controllable, easy to update, remove, or edit.<br>• Interpretable and consistent, supports reasoning and planning.<br>• Efficient for solving recurring and similar tasks. | • Flexible, solving diverse problems.<br>• Generative, responding without intermediate stages.<br>• Efficient ingestion, minimal data preprocessing. |
| **Cons** | • High construction cost for structured data.<br>• Lacks flexibility, vulnerable to missing data.<br>• High search cost for large knowledge bases. | • Expensive training and inference.<br>• Hard to control, prone to hallucination and toxicity.<br>• Lacks interpretability and transparency. |

## 1.3 Bridging Structured and Unstructured Paradigms

Despite the apparent differences between the two paradigms, this thesis seeks to bridge them in a mechanistic way, paving the path towards a unified framework for building general knowledge engines that can serve artificial intelligence agents in a dynamic environment.

Theoretically, unifying the two paradigms will deepen our understanding of their modeling principles, potentially revealing common techniques that can be applied across both structured and unstructured knowledge representations. Practically, both paradigms currently struggle with generalizing to unseen symbols. For instance, knowledge graph embedding models face challenges in generalizing to new entities, while pretrained language models often fail to generalize to unseen languages. A deep understanding of the mechanism underlying both paradigms allow us to develop new techniques that address the generalization issue.

Concretely, in this thesis, we ask:

1. What commonalities exist between structured and unstructured paradigms, given that both aim to build knowledge engines for AI agents? For example, can we identify and leverage shared techniques or methodologies that are effective across both paradigms?

2. How can we make the knowledge engines more universal? For example, how can we make models in both paradigms generalize to unseen environments faster?

## 1.4 Methodological Overview and Contributions

Our methodology begins by observing that mainstream models across both structured and unstructured paradigms share a common architectural design, which we refer to as the *Embedding Sandwich*. Specifically, these models are structured with embedding layers at both the input and output stages, enclosing a central processing module (referred to as the body of the model). The input embedding layer encodes initial data into dense, lower-dimensional representations where symbols of various granularities (e.g., words, characters, subwords, etc.) are represented as vectors. This encoded representation is then passed through the body (e.g., transformer layers, recurrent neural networks, or other architectures) that processes and transforms the information. Finally, the output embedding layer decodes the processed representation into the model's predicted output.

From there, our contributions are divided into two major research thrusts. The first focuses on *structure formation* within model computations, which naturally emerges from language modelling objectives, regardless of whether the input data is structured or unstructured. The second explores the opposite *force of destructuring*, wherein parts of the learned representation are periodically cleared to enable "model plasticity", the ability to allow the model to generalize effectively to unseen environments. These two research branches employ distinct methodologies. In Part I, we investigate the learning objective by reformulating models analytically and demonstrating how specific objectives can lead to equivalent tensor factorizations. In Part II, we focus on learning dynamics, introducing *active embedding forgetting* as a mechanism for resetting learned representations to promote adaptation in new environments.

Interestingly, while embeddings are often overlooked components or treated as yet another linear layer, our research highlights their critical role in learning symbolic relationships when using a language modelling objective. We show that a set of embeddings can store symbol interaction trajectories after trained with language modelling objectives, where parameterized inner-product computations can produce symbolic links. These symbolic interactions can subsequently be used to recover underlying global data structures (Chapter 2 and Chapter 3). We further propose a *message-passing reinterpretation* of embedding layers, where embeddings are not viewed in isolation but together with their gradient descent (GD) process (Chapter 4). GD over vector inner-products facilitates message-passing across neighbourhoods, and the vector embeddings store these accumulated relational signals.

Our theoretical analysis reveals that the generalization bottleneck stems from infinite message-passing within the training dataset. This insight suggests that *active forgetting* of embeddings mitigates this bottleneck by promoting destructuring, allowing the other parts of the model to focus on meaningful abstractions instead of being anchored to the noise in embedding initialisation (Chapter 5).

In summary, rather than focusing on surface-level distinctions such as data formats or specific model architectures, this thesis uncovers deeper conceptual connections between the two paradigms. These connections are framed along two core dimensions:

1. *Structure Formation*: This dimension depicts how symbolic relationships are encoded into model computations through language modelling objectives. The process applies to both structured and unstructured paradigms, enabling models to capture meaningful structures from different data formats, which are later useful either to complete missing entries in a knowledge engine or make a black-box knowledge engine transparent.

2. *Destructuring for Generalization*: This dimension addresses how regularly resetting learned embeddings – actively destructuring encoded structures – helps models overcome generalization bottlenecks and adapt to previously unseen symbols. The active destructuring helps models remain flexible and capable of continuous learning, regardless of whether the data is structured or unstructured.

Together, these insights reveal the mechanistic role of embeddings in the learning process, which are critical to practical tasks such as completing knowledge bases, interpreting large language models and enhancing their transparency, and addressing bottlenecks imposed by fixed vocabularies for both paradigms. These findings ultimately point toward building more *general knowledge engines* capable of adapting to new knowledge graphs, processing previously unseen languages, and potentially transferring across diverse tasks, tool usages and domains in the future.

## 1.5   Thesis Roadmap

The thesis will be organized into two main parts, Part I *Structure* and Part II *Destructure*, along with the opening and the closing. We will subsequently give an overview of these parts in the following table (Table 1.3).

Table 1.3: Overview of the thesis structure and chapter contributions.

| Part | Description |
| --- | --- |
| **Opening** | **Building Knowledge Engines.** Introduces general knowledge engines and the structured vs. unstructured paradigm divide. Presents the overarching research goal: bridging both paradigms. |
| **Part I** | **Structure – The Foundation of Knowledge Engines.** Language modelling objectives induce structure in both paradigms.<br>*Chapter 2: Language Modelling Completes Knowledge Graph Structures.* Reframes knowledge base completion as language modelling, showing how language models represent graph structure.<br>*Chapter 3: Uncovering Interpretable Structures in Pretrained Language Models.* Proposes a method to extract interpretable latent structures from LLMs using residual connections. |
| **Part II** | **Destructure – Addressing the Limits of Rigid Knowledge.** Introduces active forgetting to enhance model plasticity.<br>*Chapter 4: Inductive Knowledge Graph Learning with Active Forgetting.* Interprets factorization models as GNNs and proposes ReFactor GNNs for improved generalization.<br>*Chapter 5: Improving Language Model Plasticity with Active Forgetting.* Shows how forgetting improves adaptation in multilingual and out-of-domain settings. |
| **Closing** | **Toward General Knowledge Engines.** Summarizes findings, reflects on limitations, and outlines directions for future work. |

# Part I Structure,

*When modelling our surroundings, we construct concepts, judgments, and ourselves.*

One key characteristic of intelligent behaviours is their structured nature – acting on groups of similar objects in systematic and consistent ways, assembling small actions into larger tasks, and producing controllable outcomes. In human cognition, structures allow us to efficiently organize, remember, and recall information about the reality that is important and interesting to our survival and well-being in the world. With structures, we can make valid logical inferences in similar scenarios and generate conclusions reliably given the same premises. Likewise in intelligent systems, understanding structures is essential because it underpins efficient knowledge representation, which allows composing complex reasoning chains with smaller ones, and producing reliable outcomes given repeatable queries. Recognizing the importance of structures for both human and machine intelligence, we now turn to the two main paradigms for constructing AI knowledge engines – structured and unstructured – which, as their names suggest, approach structures in fundamentally different ways.

Structured approaches, exemplified by knowledge graphs and expert systems, predefine structures by manually specifying how entities (subjects or objects) relate to one another with different relations (predicates). This allows the same predicates to be applied consistently across similar subject-object pairs, enabling systematic manipulation of entities and relations. For instance, if we want to infer the function of a drug $A$, we can query a knowledge graph with something akin to "(?, `_is_the_function_of`, $A$)." This query pattern can systematically be reused to search for the functions of other drugs, such as $B$, $C$, $D$, etc., simply by substituting the query object. In a nutshell, the knowledge graph approach is equivalent to manually *structuring* the reality of an artificial intelligence agent: it segments the reality into discrete units, conceptualizes them as named entities, and organizes these entities into hierarchies or relational networks.

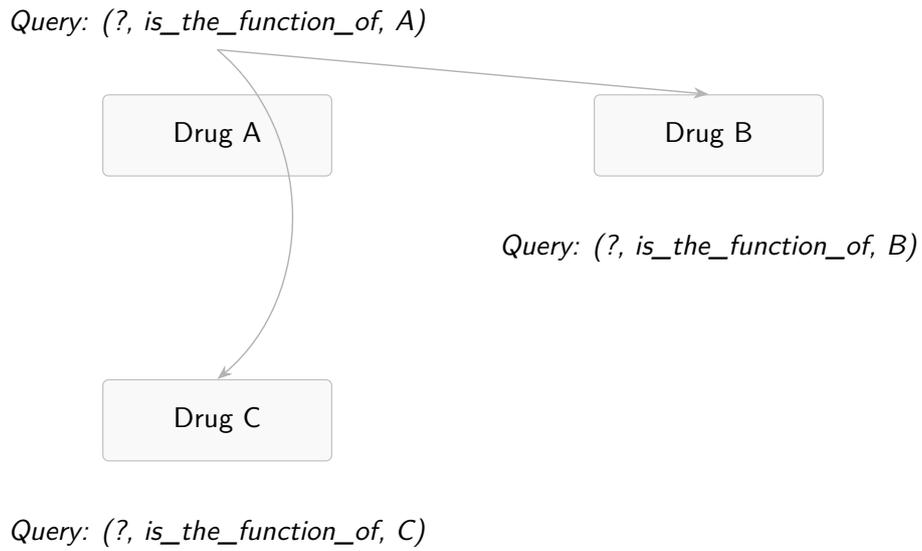

Figure 1.3: Knowledge graphs use predicates to define entity relationships, enabling the reuse of semantic structures like `is_the_function_of` across entities. This supports consistent inference over symbolic concepts.

On the other hand, the unstructured approaches, prominently featuring large language models (LLMs) trained on massive web corpora, do not prescribe any explicit structure. Continuing the drug search example above, in the unstructured approach, the relation "`_is_the_function_of`" can manifest in various forms with potential noise and lack a standarised reference identifier. Similarly, the drugs themselves do not adhere to a fixed or standardized naming structure. For instance, a drug search query posed to large language models might be expressed in natural and ambiguous terms: "Do you know when to use $A$-acid?". In this scenario, the model will retrieve information about the drug $A$ through internal implicit associations, instead of utilizing a predefined predicate. The LLM (unstructured) approach, therefore, does not seem to *structure* reality in advance or impose standardised relation predicates to its training data.

However, this part of the thesis demonstrates that, regardless of whether a structure is predefined or not, training with a language modelling objective induces inherent structures within the final embedding-encapsulated model, which serves as the bedrock for both structured and unstructured approaches. These emergent structures enable the model to recover meaningful relational patterns[2] – pertaining to the realities underlying the data – solely based on the model's learned parameters. The following chapters dive into the details for both structured knowledge completion with language modelling and completion and structural pattern emergence in large language models. Concretely,

1. **Chapter 2**: We show how a language modelling objective can help recover the structure of knowledge graphs and predict the missing links, thereby completing the knowledge base and supporting downstream knowledge queries. The language modelling objective can thus be a good alternative to classic structural recovery objectives, like entity prediction.

2. **Chapter 3**: We demonstrate how similar language modelling objectives also induce structural patterns in transformer computation, which can then be extracted and applied to enhance interpretability, transparency and safety within foundational large language models.

By examining the effects of language modelling objectives on structure formation, we find that even models from unstructured approaches acquire structures about the realities underlying the data, supporting themselves to store knowledge and to behave in the absence of predefined schematic structures. This connection between structured and unstructured approaches allows us to understand structure formation and highlight its particular role in achieving artificial intelligence, guiding our decisions on 1) when and why to incorporate structures – such as for the purpose of efficient knowledge representation, post-training interpretability, repeatable queries with consistent conclusions, and 2) how to induce structures – either explicitly through predefined structures in data for "shallow" model training or implicitly by training deep models on large corpora.

---

[2] Sequential patterns can be seen as hyper-relational edges on a graph.

# Chapter 2

# Language Modelling Completes Knowledge Graph Structures

*A version of this work was previously presented at a peer-reviewed conference. Please refer to [Chen et al., 2021] for full citation.*

Knowledge bases are one of the critical infrastructures empowering various common AI applications, including but not limited to expert systems (e.g. IBM Watson), search engines (e.g. Google Search), recommender systems (e.g. TikTok), social media (e.g. X.com) [Noy et al., 2019]. They represent the structured paradigm for building knowledge engines from curating highly structured data, e.g. knowledge graphs, that can serve various downstream applications. In this chapter, we show that a language modelling objective allows us to learn better multi-relational graph representations, leading to better structure recovery and thus can be used to complete the knowledge base automatically. Specifically, we extend the entity prediction (1vsAll) objective, which are the off-shelf choice for knowledge base completion, by incorporating *relation prediction*. The new training objective contains not only terms for predicting the subject and object of a given triple $(s, p, o)$, but also a term for predicting the relation type – predicting any symbol using its context i.e. its surrounding symbols in the triplet. This precisely matches the language modelling objective, in that we can treat the triplet as a sentence, the subject/object/predicate as the tokens, and predict the target token by modelling the context. We analyse how this language modelling objective impacts multi-relational learning for KBC: experiments on a variety of datasets and models show that the objective can significantly improve entity ranking, the most widely used evaluation task for KBC, yielding a 6.1% increase in MRR and 9.9% increase in Hits@1 on FB15k-237 as well as a 3.1% increase in MRR and 3.4% in Hits@1 on Aristo-v4. Moreover, we observe that the proposed objective is particularly effective on highly multi-relational datasets, i.e. datasets with many predicates, and generates better representations when larger embedding sizes are used. The code for our experiments is available at https://github.com/facebookresearch/ssl-relation-prediction.

## 2.1 Knowledge Base Completion as Language Modelling?

Aiming at completing missing entries, Knowledge Base Completion (KBC), also known as Knowledge Graph Completion (KGC), plays a crucial role in constructing large-scale knowledge graphs [Nickel et al., 2016a, Ji et al., 2020, Li et al., 2020]. In its essence, KBC is a task that require the model to learn the *structures* expressed in the data and thereby complete the missing entries. Over the past years, most research on KBC has been focusing on Knowledge Graph Embedding (KGE) models, which learn representations for all entities and relations in a Knowledge Graph (KG), and use them for scoring whether an edge exists or not [Nickel et al., 2016a]. Numerous models and architectural innovations have been proposed, including but not limited to translation-based models [Bordes et al., 2013], latent factorisation models [Nickel et al., 2011a, Trouillon et al., 2016, Balazevic et al., 2019], and neural network-based models [Dettmers et al., 2018, Schlichtkrull et al., 2018, Xu et al., 2020b]. Other more recent research has been making complementary efforts on analysing the evaluation procedures for these KBC models. For instance, Sun et al. [2020b] call for standardisation of evaluation protocols; Kadlec et al. [2017], Ruffinelli et al. [2020] and Jain et al. [2020a] highlight the importance of training strategies and show that careful hyperparameter tuning can produce more accurate results than adopting more elaborate model architectures; Lacroix et al. [2018] suggests that a simple model can produce state-of-the-art results when its training objective is properly selected.

Taking inspiration from these findings, we explore a language modelling style training objective, where the three symbols in a triplet are all treated equally, as tokens, and the target token is predicted by modelling the surrounding token. The main difference brought by this new objective is in that, aside from training models to predict the subject and object entities for triples in a knowledge graph, we also train them to predict the predicate, since now the predicate will simply be yet another token. This approach is akin to using a masked language model-like training objective [Devlin et al., 2019]. As we will elaborate, the simple change significantly improves multi-relational graph representation learning across several KBC models. Empirical evaluations on various models and datasets support the effectiveness of our new training objective: the largest improvements were observed on ComplEx-N3 [Trouillon et al., 2016] and CP-N3 [Lacroix et al., 2018] with embedding sizes between 2K and 4K, providing up to a $9.9\%$ boost in Hits@1 and a $6.1\%$ boost in MRR on FB15k-237 with negligible computational overhead. We further experiment on datasets with varying numbers of predicates and find that relation prediction helps more when the dataset is highly multi-relational, i.e. contains a larger number of predicates. Moreover, our qualitative analysis demonstrates improved prediction of some MANY-TO-MANY [Bordes et al., 2013] predicates and more diversified relation representations.

## 2.2 Literature Review: Design Space of Knowledge Base Completion

A Knowledge Graph $\mathcal{G} \subseteq \mathcal{E} \times \mathcal{R} \times \mathcal{E}$ contains a set of subject-predicate-object $\langle s, p, o \rangle$ triples, where each triple represents a relationship of type $p \in \mathcal{R}$ between the subject $s \in \mathcal{E}$ and the object $o \in \mathcal{E}$ of the triple. Here, $\mathcal{E}$ and $\mathcal{R}$ denote the set of all entities and relation types, respectively.

**Knowledge Graph Embedding Models** A Knowledge Graph Embedding model, also referred to as *neural link predictor*, is a differentiable model where entities in $\mathcal{E}$ and relation types in $\mathcal{R}$ are represented in a continuous embedding space, and the likelihood of a link between two entities is a function of their representations. More formally, KGE models are defined by a parametric *scoring function* $\phi_\theta : \mathcal{E} \times \mathcal{R} \times \mathcal{E} \mapsto \mathbb{R}$, with parameters $\theta$ that, given a triple $\langle s, p, o \rangle$, produces the likelihood that entities $s$ and $o$ are related by the relationship $p$.

**Scoring Functions**  KGE models can be characterised by their scoring function $\phi_\theta$. For example, in TransE [Bordes et al., 2013], the score of a triple $\langle s, p, o \rangle$ is given by $\phi_\theta(s, p, o) = -\|\mathbf{s} + \mathbf{p} - \mathbf{o}\|_2$, where $\mathbf{s}, \mathbf{p}, \mathbf{o} \in \mathbb{R}^k$ denote the embedding representations of $s$, $p$, and $o$, respectively. In DistMult [Yang et al., 2015a], the scoring function is defined as $\phi_\theta(s, p, o) = \langle \mathbf{s}, \mathbf{p}, \mathbf{o} \rangle = \sum_{i=1}^k \mathbf{s}_i \mathbf{p}_i \mathbf{o}_i$, where $\langle \cdot, \cdot, \cdot \rangle$ denotes the trilinear dot product. Canonical Tensor Decomposition [CP, Hitchcock, 1927] is similar to DistMult, with the difference that each entity $x$ has two representations, $\mathbf{x}_s \in \mathbb{R}^k$ and $\mathbf{x}_o \in \mathbb{R}^k$, depending on whether it is being used as a subject or object: $\phi_\theta(s, p, o) = \langle \mathbf{s}_s, \mathbf{p}, \mathbf{o}_o \rangle$. In RESCAL [Nickel et al., 2011a], the scoring function is a bilinear model given by $\phi_\theta(s, p, o) = \mathbf{s}^\top \mathbf{P} \mathbf{o}$, where $\mathbf{s}, \mathbf{o} \in \mathbb{R}^k$ is the embedding representation of $s$ and $p$, and $\mathbf{P} \in \mathbb{R}^{k \times k}$ is the representation of $p$. Note that DistMult is equivalent to RESCAL if $\mathbf{P}$ is constrained to be diagonal. Another variation of this model is ComplEx [Trouillon et al., 2016], where the embedding representations of $s$, $p$, and $o$ are complex vectors – i.e. $\mathbf{s}, \mathbf{p}, \mathbf{o} \in \mathbb{C}^k$ – and the scoring function is given by $\phi_\theta(s, p, o) = \Re(\langle \mathbf{s}, \mathbf{p}, \overline{\mathbf{o}} \rangle)$, where $\Re(\mathbf{x})$ represents the real part of $\mathbf{x}$, and $\overline{\mathbf{x}}$ denotes the complex conjugate of $\mathbf{x}$. In TuckER [Balazevic et al., 2019], the scoring function is defined as $\phi_\theta(s, p, o) = \mathbf{W} \times_1 \mathbf{s} \times_2 \mathbf{p} \times_3 \mathbf{o}$, where $\mathbf{W} \in \mathbb{R}^{k_s \times k_p \times k_o}$ is a three-way tensor of parameters, and $\mathbf{s} \in \mathbb{R}^{k_s}$, $\mathbf{p} \in \mathbb{R}^{k_p}$, and $\mathbf{o} \in \mathbb{R}^{k_o}$ are the embedding representations of $s$, $p$, and $o$. In this chapter, we mainly focus on DistMult, CP, ComplEx, and TuckER, due to their effectiveness on several link prediction benchmarks [Ruffinelli et al., 2020, Jain et al., 2020a].

**Training Objectives**  Another dimension for characterising KGE models is their *training objective*. Early tensor factorisation models such as RESCAL and CP were trained to minimise the reconstruction error of the whole adjacency tensor [Nickel et al., 2011a]. To scale to larger Knowledge Graphs, subsequent approaches such as Bordes et al. [2013] and Yang et al. [2015a] simplified the training objective by using *negative sampling*: for each training triple, a corruption process generates a batch of negative examples by corrupting the subject and object of the triple, and the model is trained by increasing the score of the training triple while decreasing the score of its corruptions. This approach was later extended by Dettmers et al. [2018] where, given a subject $s$ and a predicate $p$, the task of predicting the correct objects is cast as a $|\mathcal{E}|$-dimensional multi-label classification task, where each label corresponds to a distinct object and multiple labels can be assigned to the $(s, p)$ pair. This approach is referred to as KvsAll by Ruffinelli et al.

[2020]. Another extension was proposed by Lacroix et al. [2018] where, given a subject $s$ and a predicate $p$, the task of predicting the correct object $o$ in the training triple is cast as a $|\mathcal{E}|$-dimensional multi-class classification task, where each class corresponds to a distinct object and only one class can be assigned to the $(s, p)$ pair. This is referred to as 1vsAll by Ruffinelli et al. [2020].

Note that, for factorisation-based models like DistMult, ComplEx, and CP, KvsAll and 1vsAll objectives can be computed efficiently using GPUs [Lacroix et al., 2018, Jain et al., 2020a]. For example for DistMult, the score of all triples with subject $s$ and predicate $p$ can be computed via $\mathbf{E}(\mathbf{s} \odot \mathbf{p})$, where $\odot$ denotes the element-wise product, and $\mathbf{E} \in \mathbb{R}^{|\mathcal{E}| \times k}$ is the entity embedding matrix. In this chapter, we follow Lacroix et al. [2018] and adopt the 1vsAll loss, so as to be able to compare with their results, and since Ruffinelli et al. [2020] showed that they produce similar results in terms of downstream link prediction accuracy.

Recent work on standardised evaluation protocols for KBC models [Sun et al., 2020b] and their systematic evaluation [Kadlec et al., 2017, Mohamed et al., 2019, Jain et al., 2020a, Ruffinelli et al., 2020] shows that latent factorisation based models such as RESCAL, ComplEx, and CP are very competitive when their hyperparameters are tuned properly [Kadlec et al., 2017, Ruffinelli et al., 2020]. In this chapter, we show that using a language modelling like objective can further improve their downstream link prediction accuracy.

## 2.3 Transforming KBC Into Language Modelling Using Auxiliary Relation Prediction

We first recall 1vsAll, one of the typical training objectives used for learning a KBC model [Ruffinelli et al., 2020]. In 1vsAll, KBC models are trained by maximising the conditional likelihood of the subject $s$ (respectively the object $o$), given the predicate and the object $o$ (respectively the subject $s$) in the triple. More formally:

$$\arg\max_{\theta \in \Theta} \sum_{\langle s,p,o \rangle \in \mathcal{G}} [\log P_\theta(s \mid p, o) + \log P_\theta(o \mid s, p)]$$

$$\text{with} \quad \log P_\theta(o \mid s, p) = \phi_\theta(s, p, o) - \log \sum_{o' \in \mathcal{E}} \exp\left[\phi_\theta(s, p, o')\right] \quad (2.1)$$

$$\log P_\theta(s \mid p, o) = \phi_\theta(s, p, o) - \log \sum_{s' \in \mathcal{E}} \exp\left[\phi_\theta(s', p, o)\right],$$

where $\theta \in \Theta$ are the model parameters, including entity and relation embeddings, and $\phi_\theta$ is a scoring function parameterised by $\theta$. The terms $P_\theta(s \mid p, o)$ and $P_\theta(o \mid s, p)$ correspond to predicting the subject entity $s$ and the object entity $o$, respectively. These two terms align with the entity ranking task commonly used for evaluating KBC models. However, this purely discriminative formulation restricts prediction to only the first (subject) or third (object) positions, potentially overlooking structural signals that can be gained by modelling task-irrelevant postions in the triple.

On the other hand, transitioning to a generative paradigm enables the model to capture more universal patterns in the underlying data distribution, despite not directly tied to the evaluation task. To leverage the advantages of both paradigms for KBC, we follow the spirit of interpolating between generative and discriminative approaches [Bernardo et al., 2007]. Concretely, the joint distribution $P_\theta(s, p, o)$, central to generative modelling, can be factorised in three ways:

$$\begin{aligned} P_\theta(s, p, o) &= P_\theta(s, p) \underbrace{P_\theta(o \mid s, p)}_{\text{"object view"}}, \\ P_\theta(s, p, o) &= P_\theta(p, o) \underbrace{P_\theta(s \mid p, o)}_{\text{"subject view"}}, \\ P_\theta(s, p, o) &= P_\theta(s, o) \underbrace{P_\theta(p \mid s, o)}_{\text{"predicate view"}}. \end{aligned} \quad (2.2)$$

Each factorisation offers a distinct perspective on the dependencies among entities and relations. To benefit from fuller views on the joint distribution while maintaining the conditional modelling structure of 1vsAll, we propose incorporating the third view – predicate prediction – into the training objective.

Specifically, we introduce predicate (relation) prediction as an auxiliary task to ex-tend the standard 1vsAll training objective. The new training objective not only contains terms for predicting the subject and the object of the triple – $\log P(s \mid p, o)$ and $\log P(o \mid s, p)$ in Eq. 2.1 – but also a term $\log P(p \mid s, o)$ for predicting the predicate (relation typ) $p$:

$$\arg\max_{\theta \in \Theta} \sum_{\langle s,p,o \rangle \in \mathcal{G}} [\log P_\theta(s \mid p, o) + \log P_\theta(o \mid s, p) + \lambda \log P_\theta(p \mid s, o)]$$

$$\text{with} \quad \log P_\theta(p \mid s, o) = \phi_\theta(s, p, o) - \log \sum_{p' \in \mathcal{R}} \exp\left[\phi_\theta(s, p', o)\right], \quad (2.3)$$

where $\lambda \in \mathbb{R}_+$ is a hyperparameter that determines the contribution of the relation prediction objective; we assume $\lambda = 1$ unless otherwise specified.

This formulation can be viewed as a masked language modeling objective [Devlin et al., 2019] over symbolic triples, where each element – subject, predicate, or object – can be treated as a masked token predicted from the other two, with the triple functioning as a fixed-length sentence. While it remains discriminative (i.e., we do not model the full joint distribution or use autoregressive generation) in order to keep the strong classification performance, the new objective allows the model to learn contextual dependencies in all directions within a triple. This includes not only how entities depend on relation-context pairs, but also how likely a relation is to hold between a given subject-object pair. Compared to conventional approaches, the extra modelling on relation prediction helps the model better differentiate between predicates, particularly those with similar subjects or objects, or in knowledge graphs with many relation types. Section 2.4.3 will elaborate on how the new objective improves distinguishing predicates compared to the standard approach. Computation-wise, this new training objective adds very little overhead to the training process, and can be easily added to existing KBC implementations; PyTorch examples are included in Section A.1.1.

## 2.4 The Effects of Language Modelling on KBC Performance

In this section, we conduct several experiments to verify the effectiveness of the language modelling objective for KBC. We are interested in the following research questions:

**RQ1:** How does the new training objective impact the results on downstream knowledge base completion tasks across different datasets? How does the number of relation types on the datasets affect the performance of new training objective?

**RQ2:** How does the new training objective impact different models? Does it benefit all the models uniformly, or it particularly helps some of them?

**RQ3:** Does the new training objective produce better entity and relation representations?

**Datasets.** We use Nations, UMLS, and Kinship from [Kok and Domingos, 2007], WN18RR [Dettmers et al., 2018], and FB15k-237 [Toutanova et al., 2015], which are all commonly used in the KBC literature. As these datasets contain a relatively small number of predicates, we also experiment with Aristo-v4, the 4-th version of Aristo Tuple KB [Mishra et al., 2017], which contains more than $1,600$ predicates. Since Aristo-v4 has no standardised splits for KBC, we randomly sample $20,000$ triples for test and $20,000$ for validation. Table 2.1 summarises the statistics of these datasets.

Table 2.1: Dataset statistics, where $|\mathcal{E}|$ and $|\mathcal{R}|$ denote the number of entities and predicates.

| Dataset | $|\mathcal{E}|$ | $|\mathcal{R}|$ | #Train | #Validation | #Test |
| --- | --- | --- | --- | --- | --- |
| Nations | 14 | 55 | 1 592 | 100 | 301 |
| UMLS | 135 | 46 | 5 216 | 652 | 661 |
| Kinship | 104 | 25 | 8 544 | 1 068 | 1 074 |
| WN18RR | 40 943 | 11 | 86 835 | 3 034 | 3 134 |
| FB15k-237 | 27 395 | 237 | 272 115 | 17 535 | 20 466 |
| Aristo-v4 | 44 950 | 1 605 | 242 594 | 20 000 | 20 000 |
| CoDEx-S | 2 034 | 42 | 32 888 | 1 827 | 1 828 |
| CoDEx-M | 17 050 | 51 | 185 584 | 10 310 | 10 311 |
| CoDEx-L | 77 951 | 69 | 551 193 | 30 622 | 30 622 |

**Metrics** Entity ranking is the most commonly used evaluation protocol for knowledge base completion. For a given query $(s, p, ?)$ or $(?, p, o)$, all the candidate entities are ranked based on the scores produced by the models, and the resulting ordering is used to compute the *rank* of the true answer. We use the standard filtered Mean Reciprocal Rank (MRR) and Hits@$K$ (Hit ratios of the top-K ranked results), with $K \in \{1, 3, 10\}$, as metrics.

**Models** We use several competitive and reproducible [Ruffinelli et al., 2020, Sun et al., 2020b] models: RESCAL [Nickel et al., 2011a], ComplEx [Trouillon et al., 2016], CP [Lacroix et al., 2018], and TuckER [Balazevic et al., 2019]. To ensure fairness in various comparisons, we did an extensive tuning of hyperparameters using the validation sets, which consists of 41,316 training runs in total. For the main results on all the datasets, we tuned $\lambda$ using grid-search. For the ablation studies on the number of predicates and the choice of models, we set $\lambda$ to 1. This reduces computational overhead while still allowing us to examine the impact of these two factors. Details regarding the hyperparameter sweeps can be found in Section A.1.2.

### 2.4.1 RQ1: Language Modelling on Different KBC Datasets

How does the proposed language modelling training objective impact knowledge base completion for different datasets? To answer this question, we compare the performance of training with relation prediction (the language modelling objective) and training without relation prediction (the standard entity prediction objective) on several popular KBC datasets. For the smaller datasets (Kinship, Nations and UMLS), we selected the best model from RESCAL, ComplEx, CP, and TuckER. For larger datasets (WN18RR, FB15k-237, and Aristo-v4), due to a limited computation budget, we used ComplEx, which outperformed other models in our preliminary experiments.

Table 2.2 summarises the results for the smaller datasets, where  indicates training with relation (entity) prediction while  indicates training without relation (entity) prediction. We can observe that relation prediction brings a $2\% - 4\%$ improvement for MRR and Hits@1, as well as maintaining a competitive Hits@3 and Hits@10.

Table 2.3 summarises the results for the larger datasets. Including relation prediction as an auxiliary training objective brings a consistent improvement on the three datasets with respect to all metrics, except for Hits@10 on WN18RR. Particularly, relation prediction leads to increases of $6.1\%$ in MRR, $9.9\%$ in Hits@1, $6.1\%$ in Hits@3 on FB15k-237 and $3.1\%$ in MRR, $3.4\%$ in Hits@1, $3.8\%$ in Hits@3 on Aristo-v4. Compared to WN18RR, we observe a larger improvement for FB15k-237 and Aristo-v4. One potential reason is that on FB15k-237 ($|\mathcal{R}| = 237$) and Aristo-v4 ($|\mathcal{R}| = 1605$) there is a more diverse set of predicates than on WN18RR ($|\mathcal{R}| = 11$). The number of predicates $|\mathcal{R}|$ on WN18RR is comparatively small, and the model benefits more from distinguishing different entities than distinguishing different relations. In other words, using lower values for $\lambda$ (the weight of the relation prediction objective) is more suitable for datasets with fewer predicates but many entities. We include ablations on $|\mathcal{R}|$ in Section 2.4.1.

Table 2.2: Test performance comparison on Kinship, Nations, and UMLS. EP = Entity Prediction; RP = Relation Prediction. We conducted an extensive hyperparameter search over 4 different models, namely RESCAL, ComplEx, CP, and TuckER, where the model itself is also treated as a hyperparameter. Including relation prediction as an auxiliary training objective on these three datasets helps in terms of test MRR and Hits@1, while remaining competitive test Hits@3 and Hits@10. More details on the hyperparameter selection process are available in Section A.1.2.

| Dataset | EP | RP | MRR | Hits@1 | Hits@3 | Hits@10 |
|---|---|---|---|---|---|---|
| Kinship | ✗ | ✔ | **0.920** | **0.867** | **0.970** | **0.990** |
|  | ✔ | ✗ | 0.897 | 0.835 | 0.955 | 0.987 |
|  | ✔ | ✔ | 0.916 | 0.866 | 0.964 | 0.988 |
| Nations | ✗ | ✔ | 0.686 | 0.493 | 0.871 | 0.998 |
|  | ✔ | ✗ | 0.813 | 0.701 | **0.915** | **1.000** |
|  | ✔ | ✔ | **0.827** | **0.726** | **0.915** | 0.998 |
| UMLS | ✗ | ✔ | 0.863 | 0.795 | 0.914 | 0.979 |
|  | ✔ | ✗ | 0.960 | 0.930 | **0.991** | **0.998** |
|  | ✔ | ✔ | **0.971** | **0.954** | 0.986 | 0.997 |

Additionally, we conduct experiments using CoDEx, where datasets of varying sizes are created from the same data source. The results, summarized in Table 2.4, show that relation prediction consistently improves MRR and Hits@1 across the small, medium, and large datasets.

Table 2.3: Test performance on WN18RR, FB15k-237, and Aristo-v4 using ComplEx. EP = Entity Prediction; RP = Relation Prediction. Including relation prediction as an auxiliary training objective brings consistent improvements across the three datasets on all metrics except Hits@10 on WN18RR. On FB15k-237 and Aristo-v4, adding relation prediction yields larger improvements in downstream link prediction tasks. More details on the hyperparameter selection process are available in Section A.1.2.

| Dataset | EP | RP | MRR | Hits@1 | Hits@3 | Hits@10 |
|---|---|---|---|---|---|---|
| | ✗ | ✓ | 0.258 | 0.212 | 0.290 | 0.339 |
| WN18RR | ✓ | ✗ | 0.487 | 0.441 | 0.501 | **0.580** |
| | ✓ | ✓ | **0.488** | **0.443** | **0.505** | 0.578 |
| | ✗ | ✓ | 0.263 | 0.187 | 0.287 | 0.411 |
| FB15k-237 | ✓ | ✗ | 0.366 | 0.271 | 0.401 | 0.557 |
| | ✓ | ✓ | **0.388** | **0.298** | **0.425** | **0.568** |
| | ✗ | ✓ | 0.169 | 0.120 | 0.177 | 0.267 |
| Aristo-v4 | ✓ | ✗ | 0.301 | 0.232 | 0.324 | 0.438 |
| | ✓ | ✓ | **0.311** | **0.240** | **0.336** | **0.447** |

**Significance Testing**

To show that the improvements brought by relation perturbation are significant, we run the experiments with five random seeds and perform the Wilcoxon signed-rank test over the metrics obtained with and without relation prediction [Wilcoxon, 1992]. For simplicity, we select ComplEx as the base model, given its robust performance across multiple benchmark datasets. We evaluate the impact of relation prediction by computing the performance difference between ComplEx models trained with and without the auxiliary relation prediction objective. To assess statistical significance, we test the null hypothesis that the median of these differences is less than or equal to zero – i.e., that incorporating relation prediction does not improve performance over the standard 1vsAll objective.

Table 2.5 summarises the result. We can observe that almost all p-values are roughly 0.03, which means that we can reject the null hypothesis at a confidence level of about 97%. The new training objective that incorporates relation prediction as an auxiliary training objective significantly improves the performance of KBC models except for

Table 2.4: Test performance comparison on CoDEx-S, CoDEx-M and CoDEx-L using ComplEx. EP = Entity Prediction; RP = Relation Prediction. Relation prediction improves most metrics. Details in Section A.1.2.

| Dataset | EP | RP | MRR | Hits@1 | Hits@3 | Hits@10 |
|---------|----|----|------|--------|--------|---------|
| CoDEx-S | ✔ | ✘ | 0.487 | 0.441 | 0.501 | **0.580** |
|         | ✔ | ✔ | **0.488** | **0.443** | **0.505** | 0.578 |
| CoDEx-M | ✔ | ✘ | 0.366 | 0.271 | 0.401 | 0.557 |
|         | ✔ | ✔ | **0.388** | **0.298** | **0.425** | **0.568** |
| CoDEx-L | ✔ | ✘ | 0.301 | 0.232 | 0.324 | 0.438 |
|         | ✔ | ✔ | **0.311** | **0.240** | **0.336** | **0.447** |

Table 2.5: Wilcoxon signed-rank test for ComplEx-N3 on several datasets. For each dataset and metric, we report the corresponding statistics – i.e. the sum of ranks of positive differences – and the p-value as (statistics, p-value).

| Dataset | MRR | Hits@1 | Hits@3 | Hits@10 |
|---------|-----|--------|--------|---------|
| WN18RR | (15.0, 0.03125) | (15.0, 0.03125) | (15.0, 0.03125) | (3.0, 0.76740) |
| FB15k-237 | (15.0, 0.03125) | (15.0, 0.03125) | (15.0, 0.03125) | (15.0, 0.03125) |
| Aristo-v4 | (15.0, 0.03125) | (15.0, 0.03125) | (15.0, 0.03125) | (15.0, 0.03125) |

Hits@10 on WN18RR.

**Ablation on the Number of Predicates**

As previously discussed, relation prediction brings different impacts to WN18RR, FB15k-237, and Aristo-v4. Since a notable difference between these datasets is the number of predicates $|\mathcal{R}|$ (1,605 for Aristo-v4 and 237 for FB15k-237, while only 11 for WN18RR), we would like to determine the impact of perturbing relations with various $|\mathcal{R}|$. In order to achieve this, we construct a series of datasets with different $|\mathcal{R}|$ by sampling triples containing a subset of the predicates from FB15k-237. For example, to construct a dataset with only five predicates, we first sampled five predicates from the set of 237 predicates and then extracted triples containing these five predicates as the new dataset. In total, we have datasets with $|\mathcal{R}| \in [5, 25, 50, 100, 150, 200]$ predicates. To address the noise introduced in predicate sampling during datasets construction, we experimented with three random seeds. For convenience, we set the weight of relation prediction $\lambda$ to 1 and performed a similar grid-search over the regularisation and other hyperparameters to ensure that the models were regularised and trained appropriately with the different amounts of training and test data points.

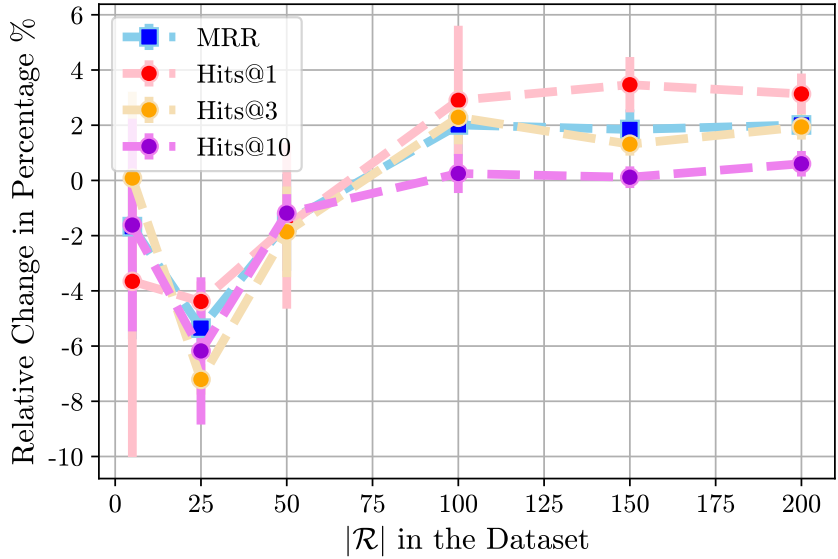

Figure 2.1: Relative changes between ComplEx trained with and w/o Relation Prediction on datasets with varying numbers of predicates $|\mathcal{R}|$. We experimented with 3 random seeds. Larger bars mean more variance. Relative changes were computed with $(m_+ - m_-)/m_-$, where $m_+$ and $m_-$ denote the metric values with and w/o relation prediction. A clear downward trend can be observed for datasets with $|\mathcal{R}| < 50$ while $2\% - 4\%$ clear increases in MRR, Hits@1, and Hits@3 are shown where $|\mathcal{R}| > 50$.

Results are summarised in Figure 2.1. As shown in the right portion of Figure 2.1, predicting relations helps datasets with more predicates, resulting in a 2%–4% boost in MRR, Hits@1, and Hits@3. For datasets with fewer than 50 predicates, there is considerable fluctuation in the relative change as shown in the left portion of the figure – but a clear downward trend. These results verify our hypothesis that relation prediction brings benefits to datasets with a larger number of predicates. Note that we did not tune the weight of relation prediction objective $\lambda$ (and fixed it to 1), and this choice might have been suboptimal on datasets with a fewer number of predicates.

### 2.4.2 RQ2: Language Modelling on Different KBC Models

Table 2.6: Test performance comparison on FB15k-237 across 4 different models: CP, ComplEx, RESCAL, and TuckER. We set the weight of relation prediction to 1 and performed a grid search over hyperparameters. More details are available in the appendix. While relation prediction seems to help all 4 models, it brings more benefit to CP and ComplEx compared to TuckER and RESCAL.

| Model | Relation Prediction | MRR | Hits@1 | Hits@3 | Hits@10 |
|---|---|---|---|---|---|
| CP | ✗ | 0.356 | 0.262 | 0.392 | 0.546 |
|    | ✓ | **0.366** | **0.274** | **0.401** | **0.550** |
| ComplEx | ✗ | 0.366 | 0.271 | 0.401 | 0.557 |
|         | ✓ | **0.382** | **0.289** | **0.419** | **0.568** |
| RESCAL | ✗ | 0.356 | 0.266 | 0.390 | 0.532 |
|        | ✓ | **0.359** | **0.271** | **0.395** | **0.533** |
| TuckER | ✗ | 0.351 | 0.260 | 0.386 | 0.532 |
|        | ✓ | **0.354** | **0.264** | **0.388** | **0.535** |

To measure how incorporating relation prediction (to induce a language modelling objective) influences the downstream prediction accuracy of KBC models, we run experiments on FB15k-237 with several models – namely ComplEx, CP, TuckER, and RESCAL. For simplicity, we set the weight of relation prediction $\lambda$ to 1. As shown in Table 2.6, including relation prediction as an auxiliary training objective brings consistent improvement for all models. Notably, up to a 4.4% and a 6.6% increase in Hits@1 can be observed respectively for CP and ComplEx. For TuckER and RESCAL, the improvements brought by relation perturbation are relatively small. This may be due to the fact that we had to use smaller embedding sizes for TuckER and RESCAL, since these models are known to suffer from scalability problems when used with larger embedding sizes. The ablation on embedding sizes of the models follows after this paragraph. As for the computational cost, the primary overhead arises from calculating $P(p \mid s, o)$. This increases the total computation to approximately $1.5\times$ that of the original objective, which only involves $P(s \mid p, o)$ and $P(o \mid s, p)$. When using a GPU, the dominant cost typically lies in matrix multiplications over all entities in the vocabulary, which is largely determined by the choice of model. For instance, models such as TuckER and RESCAL are more computationally intensive than CP and ComplEx. As a result, the overall training time remains largely unchanged after incorporating relation prediction. In our experiments, adopting the new loss led to only a 2% average increase in per-epoch training time, although more epochs may be needed to reach convergence.

**Ablations on Embedding Size**

In our experiments, increasing the embedding size of the model leads to better performance. However, there might exist a saturation point where larger embedding sizes stop boosting the performance. We are interested in how perturbing relations will impact the saturation point and which embedding sizes benefit most from it. Figure 2.2 shows the relationship between the embedding size and the MRR for CP on FB15k-237. At small embedding sizes, perturbing relations makes little difference. However, it does help CP with larger embedding sizes and delays the saturation point. As we can see, the slope of the blue curve is steeper than the red one, which bends little between an embedding size of 1,000 and an embedding size of 4,000. We can thus observe that perturbing relations leaves more headroom to improve the model by increasing its embedding sizes.

## 2.4.3 RQ3: Qualitative Analysis of Entity and Relation Representations

In our experiments, we observe that relation prediction improves the link prediction accuracy for MANY-TO-MANY predicates, which are known to be challenging for KBC models [Bordes et al., 2013]. Table 2.7 lists the top 10 predicates that benefit most from relation prediction. We rank the predicates by averaging the associated MRR of $(s, p, ?)$ and $(?, p, o)$ queries. Table A.7 and Table A.8 list the top 20 queries of $(s, p, ?)$ and $(?, p, o)$ that are improved most by relation prediction. We can see that relation prediction helps the queries like *"Where was film Magic Mike released?"*, *"Where was Paramount Pictures founded?"*, *"Which person appear in the film The Dictator 2012?"*, *"Which places are located in UK?"*, and *"Which award did Vera Drake win?"*.

To intuitively understand why the objective helps with these predicates, we ran t-SNE over the learned entity and predicate representations. Reciprocal predicates are also included in the t-SNE visualisations. We set the embedding size to 1,000, and use N3 regularisation. Hyperparameters were chosen based on the validation MRR. We run t-SNE for 5,000 steps with 50 as perplexity. As we can see from Figure 2.3, there are more predicate clusters in the t-SNE visualisation for relation prediction compared to without relation prediction. This demonstrates relation prediction helps the model distinguish between different predicates: Most predicates are separated from the entities (the pink region) while some predicates with similar semantics or subject-object contexts form a cluster (the red region); There are also a few predicates, which are not close to their predicate counterparts but instead close to highly related entities (the green region). Table 2.8 lists three example predicates for each region. Though there can be information loss during the process of projecting high-dimensional embedding vectors into two-dimensional space, we hope this visualisation suggests how relation prediction helps to learn more diversified predicate representations.

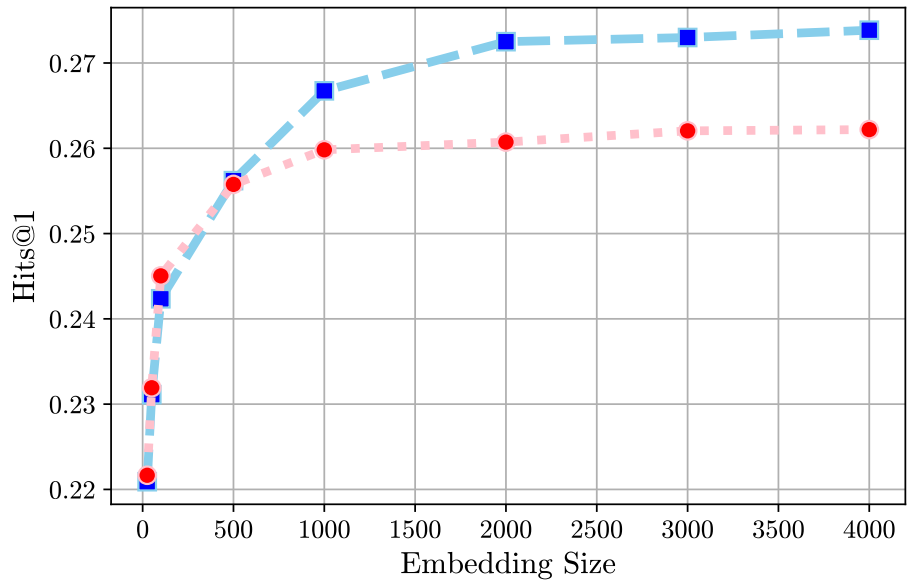

Figure 2.2: Hits@1 versus embedding size for CP on FB15k-237, each point represents a model trained with some specific embedding size with (blue) / -out (red) perturbing relations. The smallest embedding size is 25.

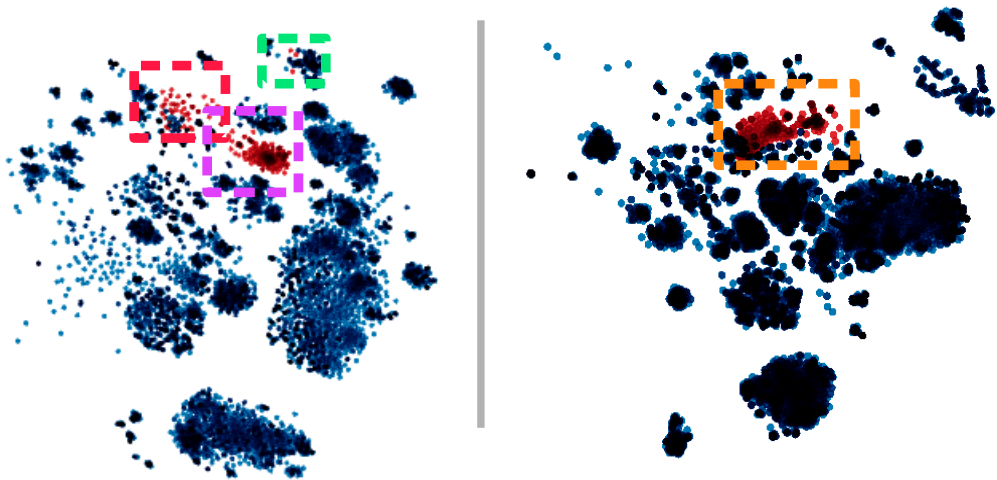

Figure 2.3: t-SNE visualisations for ComplEx embeddings, trained with relation prediction (left) and without relation prediction (right). Red points and blue points correspond to predicates and entities respectively. Dashed boxes highlight different clusters.

Table 2.7: Top 10 predicates that are improved most by relation prediction.

| |
|---|
| /ice_hockey/hockey_team/current_roster./sports/sports_team_roster/position |
| /sports/sports_team/roster./baseball/baseball_roster_position/position |
| /location/country/second_level_divisions |
| /tv/tv_producer/programs_produced./tv/tv_producer_term/program |
| /olympics/olympic_sport/athletes./olympics/olympic_athlete_affiliation/olympics |
| /award/award_winning_work/awards_won./award/award_honor/honored_for |
| /music/instrument/family |
| /olympics/olympic_games/sports |
| /base/biblioness/bibs_location/state |
| /soccer/football_team/current_roster./soccer/football_roster_position/position |

## 2.5 Discussion

**Limitations.** We mainly focus on simple factorisation-based models. Future work should consider analysing the proposed objective for more complex KBC models, such as graph neural network-based KBC models, and on more datasets. Another direction is to analyse the language modelling objective on broader downstream applications beyond link prediction.

Table 2.8: Three example predicates in each region of the t-SNE plot.

| |
|---|
| **Pink Region** |
| /base/schemastaging/organization_extra/phone_number./base/schemastaging/phone_sandbox/contact_category |
| /location/statistical_region/places_exported_to./location/imports_and_exports/exported_to |
| /sports/sports_league/teams./sports/sports_league_participation/team |
| **Red Region** |
| /people/person/nationality |
| /people/person/religion |
| /soccer/football_team/current_roster./sports/sports_team_roster/position |
| **Green Region** |
| /education/educational_institution/students_graduates./education/education/student |
| /common/topic/webpage./common/webpage/category |
| /education/educational_institution/students_graduates./education/education/major_field_of_study |

**Summary.** This chapter proposes to use a language modelling like training objective for training KBC models - by simply incorporating *relation prediction* into the commonly used 1vsAll objective. Experiments show that this new learning objective is significantly helpful to various KBC models. It brings up to $9.9\%$ boost in Hits@1 for ComplEx trained on FB15k-237, even though the evaluation task of entity ranking might seem irrelevant to *relation prediction*. The results suggest that language-modelling-like, self-supervised objectives can help models acquire structural knowledge. Moreover, even though these objectives focus solely on local contexts – i.e., the immediate surroundings of a predictive target – the induced model weights are still able to robustly recover the global structures of the knowledge graphs.

# Chapter 3

# Uncovering Interpretable Structures in Pretrained Language Models

*Parts of this work were previously presented in a preprint. Please refer to [Chen et al., 2024] for the full citation.*

In the previous chapter, we observed that language modelling objectives effectively complete knowledge graphs, indicating that these objectives can embed structural patterns in their model weights. At its core, a language modelling objective uses a token's local context to predict itself. Remarkably, this local approach enables models to infer broader, global structures within structured data, such as knowledge graphs, particularly when there is high contextual variety.[1] This prompts a natural question: *Can language modelling objectives capture global structures in any dataset, or are they limited to explicitly organized data like knowledge graphs?*

To answer this question, we study transformer based large language models (LLMs)[2] trained on unstructured texts. Typically, LLMs are trained using autoregressive language modelling objectives, where each token is predicted based on the model's analysis of all its preceding tokens in the context. We hypothesize that this local modelling in LLMs allows them to capture global structures, as factorization models do, *even* when trained on unstructured, potentially noisy datasets like web text. Accordingly, this chap-ter seeks to uncover these latent global structures within LLMs. Our method decomposes the transformer's monolithic computations into an ensemble of atomic computational paths, where each path resembles a factorization model, enabling structure recovery as in knowledge graph completion (see Chapter 2). In factorization models and knowledge graph completion, structures are typically limited to trigrams, whereas here they can potentially span n-grams with sufficient compute budget.[3] Using this method, we uncover and reconstruct structures embedded within LLMs that reflect patterns from their unstructured training data – such as common English phrases and domain-specific keywords from programming. Thus, despite training on unorganized texts, i.e. data without any structures, large language models ultimately learn and encode meaningful structures underlying the data through language modelling objectives. Since these structures are intrinsic to the trained model, they provide a basis for interpreting LLM behaviour without requiring external benchmarks, enabling data-free interpretability and transparency. We explore several applications of these intrinsic structures for language models.

---

[1] For example, when there is many diverse predicates in the knowledge graph.
[2] Also known as foundation models for their general intelligence capabilities and applications across diverse tasks.

- **Symbolic Interface.** Constructing symbolic interfaces for neural language models by sketching their (or their components') computation with the n-gram structures embedded in the model weights.

- **Behaviour Search.** Searching key n-grams in the model internal to locate and measure specific behaviours of interest, providing a deeper, structural profiling of model behaviour beyond surface-level probing.

- **Model Diff.** Enabling data-free comparison of models by analyzing differences in their n-gram structures, e.g., before and after fine-tuning.

Our case studies establish initial evidence for these applications with a few new interpretations of LLM behaviours.

- Some feedforward networks (FFNs) appear to handle simple grammatical tasks, such as adding the suffix "-ly" to preceding tokens, complementing recent findings that FFNs store factual knowledge [Geva et al., 2021, 2022].

- LLMs acquire different bigram structures at varying speeds during pretraining. In

---

[3]We leave as future work scaling the method and finding n-gram structures for $n > 3$.

*OLMo*, unique 1-to-1 bigrams like (`&`, `amp`) are acquired quickly while many-to-many bigrams like (`at`, `least`)[4] are initially promoted and later down-weighted.

- Vertical (downstream) finetuning, such as finetuning for coding tasks, raises the ranking of coding-related n-gram structures within the LLMs.

- Alignment finetuning through RLHF [Bai et al., 2022] conceals toxic n-gram structures from the surface-level outputs. Yet significant portions of toxic n-gram structures still reside within the model, making it susceptible to "jail breaking".

These findings contribute insights toward the responsible and transparent use of LLMs.

## 3.1 Interpreting LLMs by Uncovering Hidden Structures

Large language models (LLMs) are becoming increasingly prevalent as the universal knowledge engine, supporting a wide range of tasks, especially generative applications [Wei et al., 2021, Radford et al., 2019, Brown et al., 2020, Touvron et al., 2023a,b]. Despite their impressive capabilities, their opaque nature raises questions about their inner workings and the need for attribution to understand model behaviour. Mechanistic interpretability (MI) has emerged as an alternative to traditional attribution methods [Lundberg, 2017], focusing on tracing model behavior to internal structures rather than to the input [Bereska and Gavves, 2024, Ferrando et al., 2024].

Most MI research seeks to reveal the learned "algorithms" embedded within model computations, often using a hypothesis-and-dataset-driven approach. This approach typically involves forming a hypothesis, selecting a probing dataset, applying techniques like path patching [Wang et al., 2022] or causal tracing [Meng et al., 2022], iteratively refining the hypothesis in response to findings. Although valuable, this hypothesis-driven MI approach may restrict open-ended exploration, which is crucial for uncovering global behavior as did in human behavior studies [Skinner, 1965, Simon et al., 1990, Zipf, 2016], mapping model knowledge, and indexing behaviors to computation. Ultimately, MI aims to uncover and label structures within the monolithic computations described by the large neural models, with which users can index, associate and attribute various model behaviours to distinct aspects of the model operations.

---

[4]Many-to-many refers to the fact that there are rich continuations after the token `at` and precedings before the token `least`.

As we see in Chapter 2, factorization-based models (FMs) with language modelling objectives demonstrate that, after training, recovering structures can be as straightforward as computing (parameterized) inner products between embedding matrices [Trouillon et al., 2016, Lacroix et al., 2018, Balazevic et al., 2019] – revealing that these embedding matrices, derived from language modelling optimization, often store patterns aligning with underlying structures in the data, if we query them through proper operations e.g. relational weighted inner products. Given that large language models (LLMs) are similarly composed as an embedding-encapsulated system – an embedding layer, a central transformer "body", and an unembedding layer – trained using language modelling objectives, we hypothesize that similar structures latent in the model may also emerge in these large language models. We are interested in finding the structures and investigate whether such structures could facilitate mechanistic interpretability in LLMs.

To achieve this goal, this chapter introduces a method for uncovering latent structures by decomposing a transformer's computation into a set of distinct input-to-output computational paths, each of which begins with an embedding layer and ends with an unembedding layer – mirroring factorization-based models for knowledge base completion. By isolating these paths and systematically evaluating them in the input space, our method reveals n-gram structures embedded in the model's computations, analogous to how FMs reveal relational patterns in knowledge graphs.

We further discuss the relationship between such decomposition and approximating the original computation using Taylor Expansion. Despite not fully approximating the original transformer computation, the identified n-gram structures are useful for interpreting large language models as we will elaborate in our case studies. Figure 3.1 illustrates the workflow. We present a set of case studies on several autoregressive large language models (LLMs) from *Llama* and *OLMo* families with varying sizes. Our case studies illustrate that these isolated computational paths and the $n$-grams they retrieve offer valuable tools for interpreting LLM in multiple scenarios:

- i) revealing inner workings of LLMs where we identify specific functions of FFNs and attention heads, such as adding "-ing" suffixes (Section 3.5.1);

- ii) analysing pretraining dynamics where we observe distinct learning patterns for various bigrams e.g., "at least" is initially promoted and later suppressed in *OLMo* (Section 3.5.2);

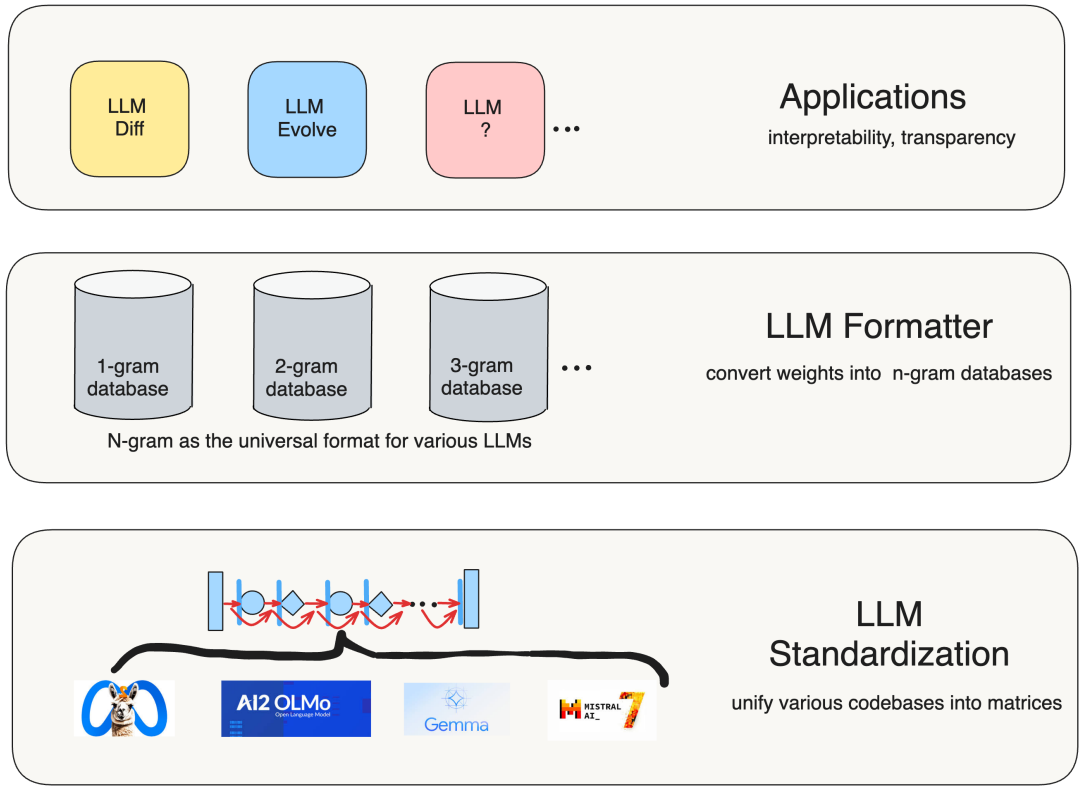

Figure 3.1: The uncovered n-gram structures can be seen as a reformatting of the corresponding large language models. These n-gram structures are derived from decomposing the transformer computations into smaller units, from where we can recompose matrix factorizations. And the identified semantic structures can support applications in interpretability and transparency.

- iii) assessing finetuning effects where we reveal model knowledge via domain-specific $n$-grams with applications in quantifying toxicity levels, finding, perhaps unexpectedly, that reinforcement learning from human feedback (RLHF) alignment [Bai et al., 2022] does not completely eliminate toxicity (Section 3.5.3). These findings support the development of more interpretable, transparent and responsible applications of LLMs.

## 3.2 Literature Review: Transformers and N-grams

**Interpreting transformers.** There has been much effort in interpreting the inner computations of transformer models. In particular, *mechanistic interpretability* [Ferrando et al., 2024] focuses on reverse-engineering such computations by identifying, clustering and labelling model behavior [Shah et al., 2024, Meng et al., 2022, Bricken et al., 2023] in human understandable terms and attributing them with certain model components, e.g., MLPs [Geva et al., 2021, 2022], or typical "circuits" [Conmy et al., 2023, Ferrando and Voita, 2024]. Recent work discussed limitations of currents approaches to MI. For example, Templeton et al. [2024] found it generally hard to conclude neuron-level interpretabilities, compared with feature representations; while Bolukbasi et al. [2021], Goldowsky-Dill et al. [2023] points out that conclusions drawn are generally limited to the chosen data distribution. As our approach focuses on manipulating functions, it does not require extra datasets that are used for probe fitting in methods such as Belrose et al. [2023] nor sampling, as needed by [Conmy et al., 2023, Ferrando and Voita, 2024, Voita et al., 2024]. On a high level, allowing singling out any portion of compute from the original monolithic transformer, our expansions abstract and generalize previous characterizations of the computational paths [Veit et al., 2016, Elhage et al., 2021], where non-linear components with significant roles, e.g. layernorm and MLPs, are either ignored or over-simplified for the ease of analysis. Additionally, zero ablations (or knock out) [Olsson et al., 2022] and direct logits attributions [Wang et al., 2022] are linked to particular instantiations of zeroth-order jet expansions [Chen et al., 2024].

**The resurgence of $n$-gram models.** The early applications of $n$-gram models for languages dates back to [Shannon, 1948], where $n$-grams were used to model the statistics of English. In essence, these $n$-grams captured structure underlying the English data they modeled: which words usually go together and which do not. The $n$-gram based approaches have since then been vital in natural language processing, particularly for general language modelling [Goodman, 2001] with applications like machine translation [Brants et al., 2007]. Recently, there have been regained interests in combining $n$-gram with neural network based approaches [e.g. Liu et al., 2024b]. Several recent works have also explored the relationships between LLMs and $n$-gram language models, such as analysing the representational capacity of transformers to emulate $n$-gram

LMs [Svete and Cotterell, 2024], and measuring the agreement between LLM predictions and curated $n$-gram rule sets [Nguyen, 2024].

## 3.3 Decomposing Transformers for Structural Recovery

Large language models are often based on the transformer architecture [Vaswani et al., 2017]. The transformer, in its original formalization, was optimized for leveraging the SIMD (single instruction multiple data) paradigm offered by the GPU for fast parallel processing sequences. Despite its efficiency, this formalization is not designed for underpinning any human-understandable structures embedded in the model. To enable structural recovery similar to how a factorization model does on a knowledge graph (Chapter 2), we need to decompose the transformer computation into smaller and easier-to-analyse units. A straightforward way is to cluster activation patterns on external datasets and treat components reacting similarly to a group of data points as a unit [Voita et al., 2024, Ferrando and Voita, 2024, Ferrando et al., 2024]. However, the recovered structures will heavily depend on the choice of data in this case, undesirable for understanding the model's global behaviour.

Luckily, transformers, despite consisting of complicated modules like self-attention, follow a simple recursive residual paradigm, where multiple identical architected residual blocks [He et al., 2016] are stacked together. We can exploit this fact to decompose computations into a set of atomic paths, each of which behave like a factorization model and enable latent structure recovery. Notation-wise, we operate at the granularity of residual blocks (e.g., self-attention or MLP blocks). This notational choice simplifies our presentation, while aligning with previous literature [Veit et al., 2016], and maintains practical relevance given the prevalence of residual computation for real-world applications [Dosovitskiy et al., 2020, Touvron et al., 2023a,b].

### 3.3.1 Neural Networks with Recursive Residual Links

We start by reviewing the archetypal computational structure of recursive residual nets, which feature transformers prominently. Specially, we focus on neural network architectures where the main body comprises multiple recursive residual blocks, with input and output managed respectively by an encoding and a decoding module. Such models fall into the same category of embedding-encapsulated models as the factorization models do, where the body is "sandwiched" between two embedding layers (see Figure 3.2).

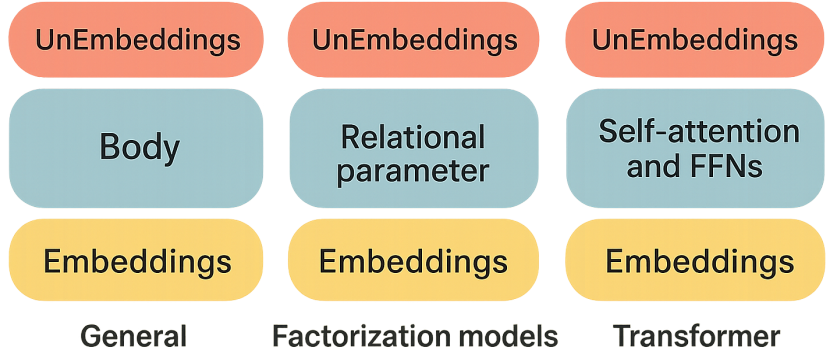

Figure 3.2: Embedding "sandwiches" are typical architectures for dealing with discrete and finite inputs to the neural networks. For example, the factorization based models for knowledge graph completion and the transformer for textual sequence completion.

Formally, let $\mathcal{Z}$ be an input space. For example, this can be sequences of tokens. Denote $c \in \mathbb{N}^+$ as the number of classes, such as the vocabulary size in a language model. Define $\mathcal{Y} = \mathbb{R}^c$ as the space of output logits, which correspond to the unnormalised over the $c$ classes. Let $d \in \mathbb{N}^+$ represent the dimensionality of the hidden representations. We are concerned with functions $q : \mathcal{Z} \to \mathcal{Y}$ described as follows:

$$q = v \circ h_L \circ \eta, \quad \text{where } h_L : \mathbb{R}^d \to \mathbb{R}^d, \ h_L = \bigcirc_{l=1}^L \beta_l, \tag{3.1}$$

where $L \in \mathbb{N}^+$ is the number of residual blocks (e.g. recursive depth), $\eta : \mathcal{Z} \to \mathbb{R}^d$ is an input encoding module (e.g. token embedding layer), $\bigcirc$ denotes repeated functional composition, and

$$\begin{aligned} &\beta_l : \mathbb{R}^d \to \mathbb{R}^d, &&\text{for } l \in [L], \\ &\beta_l = \mathrm{id} + \gamma_l, &&\gamma_l : \mathbb{R}^d \to, \mathbb{R}^d \end{aligned} \tag{3.2}$$

$$\begin{aligned} &v : \mathbb{R}^d \to \mathcal{Y}, &&v(x) = U \cdot \gamma_{L+1}(x), \\ &U \in \mathbb{R}^{c \times d}, &&\gamma_{L+1} : \mathbb{R}^d \to \mathbb{R}^d \end{aligned} \tag{3.3}$$

are respectively residual blocks with non-linearities $\gamma_l$'s (e.g., input-normalized causal self-attentions or MLPs), and the output decoding module (e.g., an unembedding projection $U$ after a layer normalization $\gamma_{L+1}$); id is the identity map. We leave all parameters *implicit* and assume all functions are infinitely differentiable $C^\infty$.

For transformer based language models, the model is optimized with a language modelling objective, where the next token is predicted based on analysing all the prior tokens in the local context. The function $q$ therefore outputs unnormalised conditional probabilities (or logits) in that

$$\mathbb{P}_q(\text{``}z \text{ belongs to class } i\text{''}|z) = \text{Softmax}[q(z)]_i, \text{ for } z \in \mathcal{Z}.$$

The recursive residual links are the critical ingredient that manages the information flow in the transformer. By carrying forward the outputs from each layer along with the embedded input, the recursive residual connections enable each subsequent layer to access not only the immediate computations of the previous layer but also the aggregated results from all prior layers. The recursive residual links thus facilitate the "storage" of computations from all preceding blocks along with the embedded input, leading to the accumulation of information across the model's depths.

### 3.3.2 Rewriting Residual Computation for Various Purposes

Although residual links have mainly been visualized as arrows connecting stacked modules in the mainstream expression of Eq. 3.1, we note that this is a perspective that renders their role in easing the training of deep networks. Such an expression of Eq. 3.1, suited for developing and training the deep residual nets, might not be suitable for analysing and interpreting them. Therefore, rewriting them in other ways become necessary for post training analysis and interpretability. Figure 3.3 summarizes several rewritings for different purposes.

**Nested update accumulation**   Notably, as visualized in Figure 3.3 (b), we can rewrite the recursive computation of Eq. 3.1 by accumulating all the prior block outputs up to

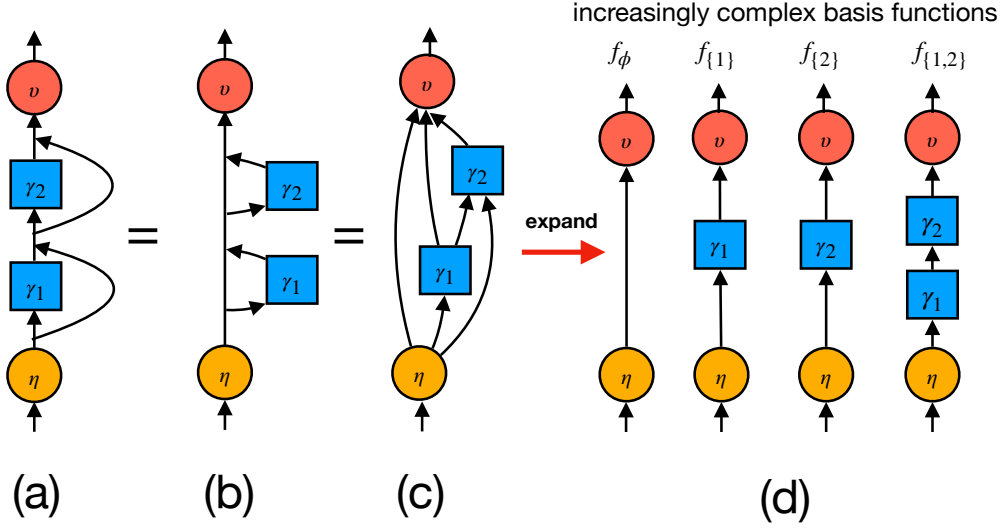

Figure 3.3: Various expressions of residual stream, each emphasizing a different aspect. (a) a visual expression adapted from [He et al., 2016, Vaswani et al., 2017], highlighting the identity shortcuts which ease the training of very deep models. (b) a visual expression adapted from [Elhage et al., 2021, nostalgebraist, 2021], highlighting the updates being written into the residual stream which serve as a communication channel. (c) a visual expression adapted from [Veit et al., 2016], highlighting the unrolling of all the residual links (d) a visualization highlighting our proposed decomposition in Section 3.3.3 into separated input-to-output computational paths which are useful for interpretability. For a linear residual net, (a)-(d) are equivalent expressions.

block $l \in [L]$, assuming $h_0 = \eta$:

$$\begin{aligned} h_l &= \left( \bigcirc_{j=1}^{l} \beta_j \right) \circ \eta = \eta + \sum_{j=1}^{l} \gamma_j \circ h_{j-1} \\ q &= \upsilon \circ \eta + \sum_{l=1}^{L} \upsilon \circ \gamma_l \circ h_{l-1}. \end{aligned}$$

(3.4)

Elhage et al. [2021] introduces the term *residual stream* to describe $h_l$, while similar concepts like "residual bus" can be traced back to Hochreiter and Schmidhuber [1997] and Srivastava et al. [2015]. Such rewritings of recursive residual links have been widely applied in the mechanistic community [Elhage et al., 2021, nostalgebraist, 2021], highlighting the updates produced by each block (e.g. the self-attention block or the FFN block in the standard transformer) being written into the residual stream which serve as a communication channel.

**Gradient paths**   Similarly, Veit et al. [2016] describe and study the unrolled structure of the final residual stream expressed as $h_L = \eta + \sum_{j=1}^{L} \gamma_j \circ h_{j-1}$, which reveals a number of paths from the input to the decoder (rather than the output), growing *linearly* with the network depth $L$. This expansion is illustrated by the three pathways (black arrows) leading to the node $v$ (red circle) in Figure 3.3 (c) for a case of two-layer residual architecture. Because the differentiation is a linear operator, this kind of rewriting is useful for analysing the gradient flow during backpropagation, where one can track common issues in training deep neural networks, such as gradient vanishing and gradient ensembling from different paths. However, this rewriting alone does not lend itself directly to analysing the model's intrinsic input-output functional relationships. To "mechanistically" understand the model's behaviour, a further decomposition is needed to reflect the internal structure underpinning the model's knowledge possession.

### 3.3.3   Rewriting Recursive Residual Networks into Factorizations

So far, we have described several rewritings of a recursive residual computation graph, each for a different purpose. For instance, Eq. 3.4 decomposes the original computational graph into a series of additive terms. Each term builds incrementally on the previous ones, forming a hierarchical structure. Despite resembling a series expansion (e.g., a Fourier Expansion), the terms in this rewriting are not sufficiently "atomic" – the interdependency among terms and their intertwined roles complicate direct interpretation.

**Decomposing recursive residual networks into $2^L$ input-output paths**   To systematically decompose the nested terms in Eq. 3.4, we observe that each $\gamma_l$ takes as input a sum of upstream terms. Let us consider a sum $x_1 + x_2$ as the input signal. If $\gamma_l$ preserves addition, i.e. it is an additive map [Reed and Simon, 1980], then $\gamma_l(x_1 + x_2) = \gamma_l(x_1) + \gamma_l(x_2)$, naturally expanding the nested terms into distinct chains of dependencies that trace back to the input when applied at all residual links. The original computational graph can then be expanded as a sum of $2^L$ unique paths. Each path applies $L$ transformations, where each transformation is either $\gamma_l$ or id. Formally, we can rewrite $q$ by

$$\begin{aligned}
q &= \upsilon \circ \{\bigcirc_{l=1}^{L}(\text{id} + \gamma_l)\} \circ \eta \\
&= \upsilon \circ \left( \sum_{s \in \{0,1\}^L} \bigcirc_{l=1}^{L} \gamma_l^{s_l} \right) \circ \eta \\
&= \sum_{s \in \{0,1\}^L} \upsilon \circ (\bigcirc_{l=1}^{L} \gamma_l^{s_l}) \circ \eta \\
&= \sum_{s \in \{0,1\}^L} f_s.
\end{aligned} \quad (3.5)$$

Here $s = (s_1, s_2, ... s_L)$ is an $L$-bit binary vector in the set of $\{0,1\}^L$, indicating a unique path configuration. $s_l = 1$ represents the path using the $\gamma_l$ transformation. $s_l = 0$ represents the path using the identity transformation id. $\bigcirc_{l=1}^{L} \gamma_l^{s_l}$ is the sequential composition used by the path according to $s$. This rewriting reveals that the original recursive residual computation behaves as an ensemble of $2^L$ increasingly complex input-to-output computational paths $f_s : \mathcal{Z} \to \mathcal{Y}$ sharing $L$ core components. The complexity of a path is determined by the number of non-identity transformations it involves. Thus the hierarchy of the paths implies interesting properties of the recursive residual computation. For example, simpler paths with fewer $\gamma_l$ terms might capture broad and abstract data patterns while more complex paths might capture finer details and potentially nuanced noise. Moreover, these paths include "non-continuous", where one path can skip one or several blocks and directly go to the later portion of the computation graph.

**Linear recursive residual networks as an ensemble of factorization models** In the real domain, linear $\gamma$'s are additive maps. So if we assume all $\gamma$'s are linear, such that $\gamma_l(x) = A_l x$, for $l \in [L]$, and assume the encoder $\eta(x) = Ex$ and the decoder $\upsilon(x) = Ux$ then the result of the above decomposition turns out to be an ensemble of factorization models:

$$q = \sum_{S \in 2^{[L]}} U \left( \prod_{l \in S} A_l \right) E \quad (3.6)$$

where $2^{[L]}$ is the power set of $[L]$ which contain $2^L$ elements, meaning $S$ could for example be $\{1\}$ or $\{1,2\}$ etc. Let us denote $W_S = \prod_{l \in S} A_l$, which is a $d \times d$ projection matrix, and $f_S(x) = W_S x$ denotes the mapping of the selected path. So we have

$$q = \sum_{S \in 2^{[L]}} U W_S E^\top$$

which is exactly a generalized factorization models where $U \in \mathbb{R}^{c \times d}$, $E \in \mathbb{R}^{c \times d}$ are the two embedding matrices wrapping the $W_S$ matrix. From this we can see that a linear transformer boils down to an ensembling of $2^L$ weighted matrix factorization $U W_S E^\top$, where $W_S \in \mathbb{R}^{d \times d}$ is the weighting matrix between $U$ and $E$. Akin to how predicates (relations) weight the subject embeddings and the object embeddings, here $W_S$ plays a similar role as a special kind of global predicates (and self-attention might act as local predicates as our ongoing work shows). And most importantly, the outcomes from these individual factorization models $D_S = U W_S E^\top \in \mathbb{R}^{c \times c}$ becomes a database storing the $c \times c$ interactions between the $c$ tokens, resembling how a factorization model based scoring function stores the links on a knowledge graph. These direct readouts from the individual input-output paths thus recover the latent input-output structure underlying the model computation. When applied to language models, we are equivalently converting a large language model into a set of factorization models and thus into their associated token interaction databases – a symbolic reformatting into a set of bigram databases, where high-scoring entries reflect meaningful information structures about the training dataset. Figure 3.4 illustrates this process.

**Non-linearity in $\gamma_l$'s**   In practical residual architectures, however, $\gamma_l$ are typically non-linear and do not preserve addition – meaning $\gamma_l(x_1 + x_2)$ can not be expanded into separate terms associated with each individual upstream input $x_i$. As a result, nested terms in Eq. 3.4 are retained and the decomposition into $2^L$ paths is not immediately possible. However, we show that we can still single out any target computational path from the super exponential set of block combinations as we do for the above linear $\gamma_l$ case and empirically obtain meaningful structural recovery as we show in Section 3.5 Despite the practical transformer's non-linearity, we argue that this simple method resembling the factorization based models enable meaningful structure recovery, of which the effectiveness is validated with our case studies. In addition, the rewriting error can be reduced via higher-order expansions with jets as we present the method in a follow-up work of this chapter [Chen et al., 2024], where we propose to use jets expansions to handle non-linearities.

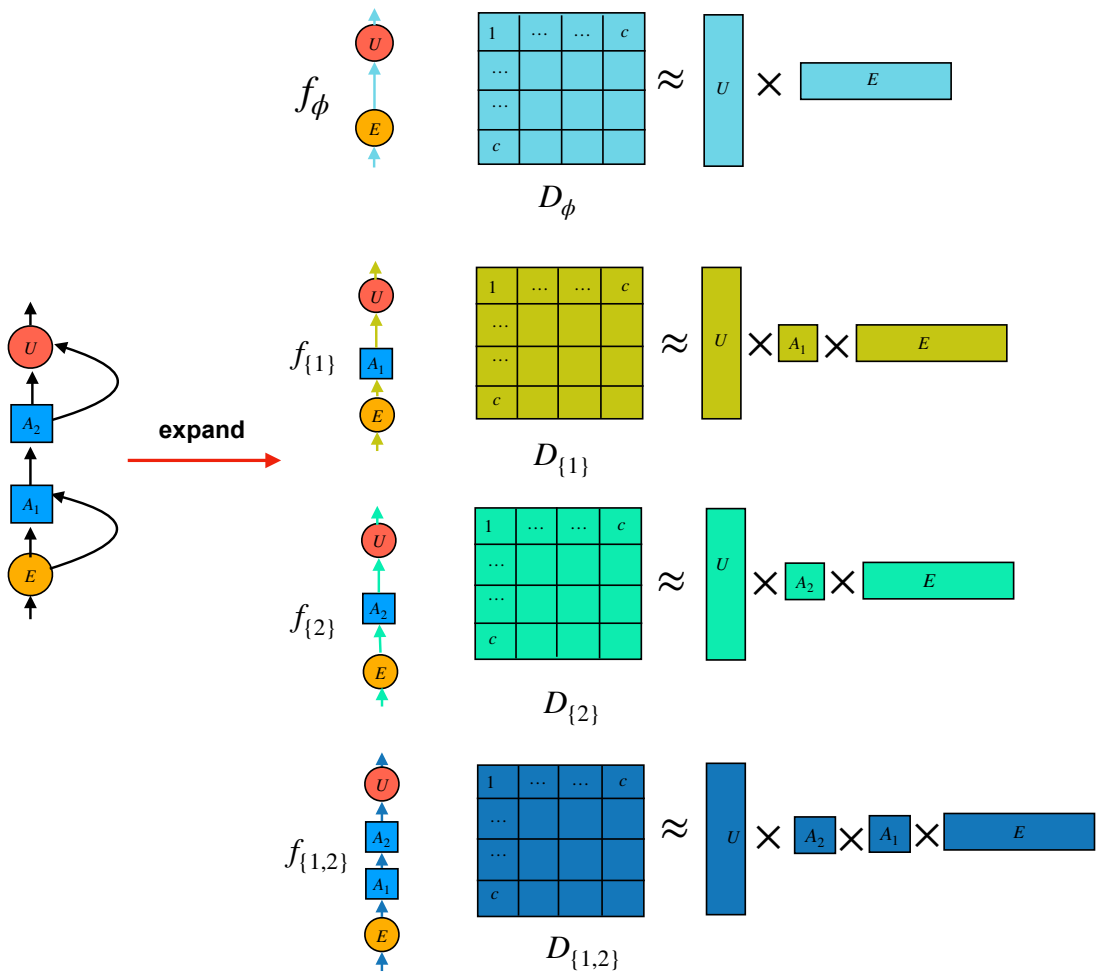

Figure 3.4: Cartoon of the process of deriving bigram databases $D_S$ from the embedded factorization model in each expanded input-output path $f_S$ for a two-layer recursive residual net. For example, $D_{\{1\}}$ is derived from the path $f_{\{1\}}$. These bigram databases can be used to depict their corresponding paths to a certain extent.

## 3.4 Extracting N-gram Structures from Pretrained Language Models

Now that we have established that factorization models can be pinpointed within (linear) transformers, we can extract symbolic knowledge bases systematically from pretrained language models. These knowledge bases, represented in n-gram formats, can be used to analyse structural information captured by large language models, thus bridging the gap between arithmetic computations (e.g. matrix multiplications) and interpretable structures (e.g. domain keywords or other semantically meaningful units). As stated above, the practical transformer contains non-linear components such as normalization function before input to each module. Implementation-wise, we chose to incorporate these normalization functions into the input-output paths, and empirically we find these non-linearities improve the quality of the extracted bigrams compared to using purely linear paths [Elhage et al., 2021].

This section details our algorithms for extracting n-gram knowledge bases from the factorization models embedded in transformer-based LLMs, specifically on unigrams, bigrams, and trigrams. Due to computational constraints, higher-order ngrams with $n > 3$ are left for future work. Positional embeddings and the discussion on their choices (absolute learnable positional embeddings v.s. relative positional embeddings) are also excluded to avoid additional complexities beyond this study's scope.

### 3.4.1 Bigrams

We focus on bigrams, as they are the first studied in the literature [Elhage et al., 2021]. Algorithm 1 outlines our approach to computing pairwise token interaction scores for bigrams using token embeddings ($E$), an unembedding matrix ($U$), and paths through selected network components. The algorithm can be extended to accommodate any computational path among the $2^L$ possible paths through the transformer blocks. In this study, we consider the following path options and use OLMo [Groeneveld et al., 2024] as a demonstrative model in the algorithm:

1. **Direct Path**: This path processes embeddings directly without intermediate transformations, as described in Elhage et al. [2021]. Additionally, our algorithm incorporates the non-linearities presented in the OLMo architecture. The token embeddings ($E$) are normalized using RMS normalization (RMSNorm), and the normalized embeddings are projected onto the unembedding space to compute the interaction scores, represented as $D_{T,T+1}$. This bigram database corresponds to the path represented as $f_\phi$.

2. **Single FFN Path**: This path includes a single feed-forward network (FFN) block into the direct path. The token embeddings are first normalized using RMSNorm, passed through the FFN, and normalized again. The resulting embeddings are projected onto the unembedding space to compute the interaction scores. This bigram database corresponds to the path represented as $f_{\{FFN_i\}}$.

3. **Merged Path with Multiple FFNs**: This option allows merging a list of selected FFNs along with the direct path. This bigram database corresponds to the path represented as $f_{\{FFN_{i_1},...,FFN_{i_m}\}}$. For this path:

   (a) An accumulation tensor ($e$) is initialised with the normalized embeddings ($e \leftarrow \text{RMSNorm}(E, \epsilon)$).

   (b) For each selected FFN in the set, embeddings are normalized, processed through the FFN, and normalized again. The FFN outputs are accumulated into $e$.

   (c) After processing all selected FFNs, the final interaction score is computed as $D_{T,T+1}$, normalized by the number of FFNs plus one direct path ($|\mathcal{C}| + 1$).

In all paths, a SoftMax operation is applied to the unnormalised scores $D_{T,T+1}$ along the first dimension, ensuring interpretability as probabilities. In essence, the algorithm evaluates these paths over the vocabulary space by *wrapping* the selected components *with* the token embeddings ($E$) and the unembedding matrix ($U$). The final output is a 2D tensor $D_{T,T+1}$ that captures the pairwise interactions between tokens $T$ and $T+1$. This tensor serves as a quantitative approximation of a bigram statistic $\mathbb{P}_q(z_{T+1}|z_T,\dots)$, revealing the token interaction dynamics embedded in the selected path(s). This bigram algorithm can be extended to encompass the full residual computation rather than focusing on partial computations. We refer to the results derived from this specific path choice as naive bigrams. However, naive bigrams have limitations: they cannot describe arbitrary paths of interest, nor do they facilitate the analysis of path contributions to model behaviour. Therefore, we skip them in the empirical study.

**Algorithm 1:** Bi-gram Score. Compute 2-gram token interaction graph embedded in embeddings, unembeddings and FFNs. Applicable to the OLMo architecture with vanilla attention and non-parametric RMSNorm

**Input:** Token embeddings $E$, unembedding matrix $U$, path option $p$, a set of components $\mathcal{C}$ along the specified path
**Output:** $D_{T,T+1}$, a 2D tensor of pairwise token interactions
**Function** bigram($E, U, p, \mathcal{C}$):
    **if** $p$ is *direct path* **then**
        $x \leftarrow$ RMSNorm($E, \epsilon$) ;        // Apply RMS normalization
        $D_{T,T+1} \leftarrow xU^\top$ ;        // Project onto unembeddings
    **else if** $p$ is *single FFN path* **then**
        $x \leftarrow$ RMSNorm($E, \epsilon$);
        $x \leftarrow$ FFN($x$);
        $x \leftarrow$ RMSNorm($x, \epsilon$);
        $D_{T,T+1} \leftarrow xU^\top$;
    **else if** $p$ *includes Feed-Forward Networks (FFNs)* **then**
        $e \leftarrow$ RMSNorm($E, \epsilon$) ;        // Initialize accumulation
        **foreach** *FFN* $\in \mathcal{C}$ **do**
            // Normalize embeddings for FFN computation
            $x \leftarrow$ RMSNorm($E, \epsilon$);
            // Perform FFN computation
            $x \leftarrow$ FFN($x$);
            // Normalize FFN output and accumulate
            $x \leftarrow$ RMSNorm($x, \epsilon$);
            $e \leftarrow e + x$;
        // Compute final interaction score across layers
        $D_{T,T+1} \leftarrow eU^\top$;
        $D_{T,T+1} \leftarrow \frac{D_{T,T+1}}{|\mathcal{C}|+1}$;
    Apply softmax on $D_{T,T+1}$ along dimension 1;
    **return** $D_{T,T+1}$

### 3.4.2 Extension to Unigrams

Unigrams can be obtained via finding the stable state of the Markov transition equation defined via the bigrams conditional probability (Algorithm 2). The algorithm calculates unigram scores by first deriving the Markov transition matrix from bigram probabilities using the direct path, then performing an eigendecomposition to identify the steady-state eigenvector ($\lambda = 1$), which represents the unigram probabilities, and finally returning this as the unigram score.

---
**Algorithm 2:** Unigram Score. Applicable to the OLMo architecture with vanilla attention and non-parametric RMSNorm.

---
**Input:** Embeddings $E$, Unembeddings $U$, RMSNorm constant $\epsilon$
**Output:** $D_{T+1}$, a 1D tensor storing individual token score, representing their prominence within the model.
**Function** unigram($E, U, \epsilon$):
    Obtain transitions $D_{T,T+1} \leftarrow$ bigram($E, U,$ *direct path*, $\emptyset$);
    Initialize the steady state $D_{T+1}$ as a 1D zero tensor;
    Compute eigenvalues and eigenvectors
      $\{\lambda_i\}, \{\mu_i\} \leftarrow$ eigen_decompose($D_{T,T+1}$);
    // Loop over eigenvalues to identify the stable state
    **foreach** $\lambda_i, \mu_i$ *in* $\{\lambda_i\}, \{\mu_i\}$ **do**
      **if** $\lambda_i == 1$ **then**
        $D_{T+1} \leftarrow \mu_i$;
    **return** $D_{T+1}$;

---

### 3.4.3 Extension to Trigrams

Calculating trigrams or skip n-grams becomes more nuanced because it requires unpacking the mechanism of **self-attention modules**.

**Self-Attention: Beyond Immediate Tokens**    Self-attention enables a model to attend to tokens beyond just the immediate neighbours (e.g., bigrams). By applying one self-attention layer, the model collects information from tokens farther away in the sequence.

For instance:

- **Predicting Token** $T+1$**:** Using the representation at position $T$, one self-attention allows the model to attend to any previous token $k$ ($k < T$). The information flow can be represented as:

$$T+1 \underbrace{\leftarrow}_{\text{time step}} T \underbrace{\leftarrow}_{\text{time step}} k$$

Here, $T$ passes relevant context from $k$ to $T+1$, creating a chain of dependencies over time steps.

**Skip N-Grams: Information Steps** The above equation uses time steps as the coordinates for a stream of tokens. However, a different coordinate axis will reveal more informative reliance among tokens. Skip n-grams view the same information flow from an **information step** perspective, rather than a time step. For instance, the skip trigram process looks like this:

$$n+1 \underbrace{\leftarrow}_{\text{information step}} n \underbrace{\leftarrow}_{\text{information step}} n-1$$

In this view:

- $n$ carries relevant context from $n-1$ to $n+1$.

- This contrasts with bigrams, where $n-1$ passes information directly to $n+1$ without intermediary steps.

Identifying such patterns embedded in the model can be useful to understand what kind of knowledge is being stored in the model.

**Example: Skip N-Grams in a Sentence** Consider the sentence: "Lemma (Properties of Jets) Let s be the function to be approximated." If there is a sufficient number of similar sentences in the training dataset, for example the training dataset contains heavy portion of maths texts, then the model would capture skip-trigrams like:

- Token $z_{n-1}$: "Lemma"

- Token $z_n$: "Let"

- Token $z_{n+1}$: "s"

**Connecting Self-Attention with Skip Trigrams**   We can obtain skip trigram statistics relating to $\mathbb{P}_q(z_n|z_{n-1},\ldots,z_{n-2},\ldots)$, where dots indicate any number of interceding tokens, by focusing on paths that contain one self-attention module and possibly filtering out all paths that involve more than one self-attention. In general, paths with more self-attentions will have higher $n$.

Algorithm 3 describes in detail how we obtain the trigrams. During the calculation of the attention score between token $T$ and $k$, the current token $T$ becomes a bucket for storing several contextual token $k$ along with their weightings, and pass them later to the target token $T+1$ with weighting. The big 3D tensor for describing triplet interactions among $(k, T, T+1)$ is decomposed into matrices from two steps $T \to k$ and $k \to T+1$. In other words, we trace the indirect influence of each context token $k$'s onto the $(T, T+1)$ pairings by performing a non-contracted tensor product[5] between the $T \to k$ messaging matrix and the $k \to T+1$ messaging matrix.

Such $n$-gram statistics extracted directly from large language models can serve as a *data-free* tool to sketch LLMs via casting them into (symbolic) $n$-gram databases. Thus, they allow us to perform symbolic model comparison between *any* two models that share a common vocabulary, as opposed to taking differences in the parameter space, which is harder to interpret and only possible for models with the same architecture.

---

[5]It is interesting to see the non-contracted tensor products become the key operators for unpacking transformer computation and derive interpretable structures. Its contracted version, matrix products, works well when training deep neural networks on GPUs, where the SIMD paradigm prefers massive parallel ALU computation and accumulating the intermediate computation results rather than caching them all in memory and sequencing the computation. However, when we move to the interpreting neural network phase, it seems that accumulating the intermediate results all the way forward, i.e. the "deep" computation, can be less relevant compared to the "wide" computation, where non-contracted tensor product can keep track of all combinations of the indices – in language models indices correspond to tokens – without reducing them via summation. With "wide" operators like non-contracted tensor product, we can capture global information flow inside the entire vocabulary space, without collapsing higher-order token interactions. The drawback is that it requires large amounts of memory to store all the interactions. We foresee that there is a hardware lottery [Hooker, 2021] for language models interpretability akin to how training deep language models favors GPUs. For example, in this chapter, we do not use any GPUs but adopt CPUs with 1 TB memory.

**Algorithm 3:** Trigram Score. Compute 3-gram token interaction graph embedded in a self-attention layer via sparsely joining all attention heads. Applicable to the OLMo architecture with vanilla attention and non-parametric RMSNorm

    **Input:** embeddings $E$, unembeddings $U$, attention weights $W_q, W_k, W_v, W_o$, RMSNorm constant $\epsilon$, head size $D_h$, target head indices `heads`,

    **Output:** $D_{T,k,T+1}$: a sparse 3D tensor storing interactions

    $e \leftarrow$ `RMSNorm`$(E, \epsilon)$;

    Initialize $D_{T,k,T+1}$ as zero tensor;

    **for** $h \in$ *heads* **do**

        Obtain current head dimensions $H = [hD_h : (h+1)D_h]$;

        Obtain QK matrix $W \leftarrow W_{q\,[:,H]}^T W_{k[H,:]}$;

        Obtain OV matrix $V \leftarrow W_{v\,[:,H]}^T W_{o[H,:]}$;

        Compute QK message $D_{T,k} \leftarrow \frac{eWe^T}{\sqrt{D_h}}$;

        Apply softmax normalization on $D_{T,k}$ along dimension 1;

        Sparsify $D_{T,k}$ based on threshold to obtain sparse tensor $\tilde{D}_{T,k}$;

        Compute $D_{k,T+1} \leftarrow$ `RMSNorm`$(eV, \epsilon) \cdot U^T$;

        Apply softmax normalization on $D_{k,T+1}$ along dimension 1;

        Sparsify $D_{k,T+1}$ based on threshold to obtain sparse tensor $\tilde{D}_{k,T+1}$;

        Compute $D_{T,k,T+1}^{(h)} \leftarrow$ `non_contracted_tsr_prod`$(\tilde{D}_{T,k}, \tilde{D}_{k,T+1})$;

        Accumulate $D_{T,k,T+1} \leftarrow D_{T,k,T+1} + D_{T,k,T+1}^{(h)}$ ;

    `// weighting trigrams with bigrams`

    Compute $D_{T,T+1} \leftarrow$ `bigram`$(E, U, \epsilon)$;

    Compute $D_{T,k,T+1} \leftarrow 32 D_{T,k,T+1} + D_{T,T+1}$;

    **return** $D_{T,k,T+1}$

---

**Algorithm 4:** Non-Contracted Tensor Product $A_{i,j} B_{j,k} = C_{i,j,k}$

    **Input:** Two tensors $A$ and $B$

    **Output:** A 3D tensor $C$

    **Function** `non_contracted_tsr_prod`$(A, B)$:

        **for** *each index $i$ and $k$* **do**

            `// if vectorized, an outer product` $A_{[i,:]} \otimes B_{[:,k]}$

            **for** *each index $j$* **do**

                Compute $C_{i,j,k} = A_{i,j} \times B_{j,k}$;

        **return** $C$

## 3.5 Case Studies: Latent Structures for Interpreting Language Models

In this section, we explore applications of the uncovered n-gram latent structures. We present several case studies where we utilize the identified structures for understanding and interpreting large language models. To showcase the generality of the structure-revealing method, we conduct experiments with popular open-source large language model families: *Llama* [Touvron et al., 2023a,b, Rozière et al., 2024] and *OLMo* [Groeneveld et al., 2024]. Our experiments run on servers with 1 TB of memory and 128 CPUs. Unlike traditional mechanistic interpretability studies, our method does not rely on GPUs or external datasets for collecting network activation patterns, making it more accessible to resource-constrained communities.

### 3.5.1 Use Case 1: Analysing LLM Inner Workings

Large language models are notorious for their lack of interpretability [Zhao et al., 2024a]. The lack of interpretability is due to their inherent model complexity and size, made worse by the usual opaque training process and unknown training data. Understanding their inner workings, for example the roles of different components, can help calibrate trust for users to use them appropriately. We showcase how the bigrams and trigrams extracted along user-selected computational paths can help us discover and locate learned associations akin to studies in mechanistic interpretability [Templeton et al., 2024], but without any additional training or inference on external datasets.

**Paths of individual components.** By examining the representative bigrams that are captured by each MLP path, we find MLPs that might perform special linguistic functions. For example, in *OLMo-7B*, the path which passes through the 3rd MLP promotes the addition of the "-ing" suffixes to the current token. Similar MLPs with certain linguistic functions are listed in Table 3.1. Note that the relationship between functions and components are not necessarily one-to-one mappings. Particularly we find that the paths through multiple MLPs might work together to complete one linguistic function e.g. MLP 6 and MLP 18 in *Llama-2-7B* can add "-ing" suffix. One MLP might also do multiple linguistic jobs e.g. MLP 1 in OLMo 7B adding "-ly" and "-_else" suffixes.

Table 3.1: MLPs in *OLMo-7B* and *Llama-2-7B* performing linguistic functions based on jet bi-grams extracted from the corresponding jet paths. Logit values are computed after intervention.

| | *OLMo-7B* | | | *Llama-2-7B* | |
|---|---|---|---|---|---|
| **MLP** | **Role** | **Δ logit** | **MLP** | **Role** | **Δ logit** |
| 1 | `-ly, -_else` | $-4.19, -3.35$ | 6 | `-ing` | $-14.61$ |
| 3 | `-ing` | $-0.58$ | 7 | `-es` | $-3.55$ |
| 9 | `-'t` | $-9.73$ | 18 | `-ing, -ity` | $-9.69, -11.93$ |
| 17 | `-_than` | $-4.26$ | 19 | `-ly` | $-9.14$ |
| 19 | `-s` | $-7.42$ | | | |

This echos work on circuit discovery [Conmy et al., 2023, Ferrando and Voita, 2024] and superposition [Elhage et al., 2022], where the role of each component can not easily be dissected and multiple components collaborate to fulfil a function. Table 3.2 reports a role identification study on attention heads in the first self-attention of *OLMo-7B* using trigrams. Specifically, we find heads associated with maths and programming, e.g. head 1 on Maths/latex; heads promoting digits and dash composition into dates, e.g. head 25; and heads constituting phrase templates, e.g. head 15 managing a "for $x$ purposes", where $x$ is a placeholder. To verify the roles we revealed, we further perform preliminary intervention experiments where we ablate MLPs or attention heads and compute variations in model logits. After the interventions, the logits drop consistently for all cases, suggesting our $n$-grams indeed can help identify roles for selected components. Varying impact on logit differences is likely due to overdetermination [Mueller, 2024] and our partial selection of paths (e.g. for trigrams we only selected encoding-attention-decoding paths, excluding any MLP).

### 3.5.2 Use Case 2: Analysing Pretraining Dynamics

Pretraining an LLM is usually highly resource-intensive. Therefore, it is crucial to monitor the progress of a pretraining run to prevent wasting of time and compute. In this section, we show how bigrams can serve as an effective signalling tool to trace the pretraining dynamics, providing insights about the model's maturity. Such signals are especially useful to understand what happens with the model when the pretraining loss

Table 3.2: Several attention heads in the first residual block of *OLMo-7B* and their roles identified with jet trigrams extracted from corresponding jet paths. We also include an example trigram captured by each head.

| Head Index | Role | Example 3-gram | $\Delta$logit |
|---|---|---|---|
| 2 | Maths/latex | (_Lemma, _let, _s) | -0.1570 |
| 16 | "for...purposes" | (_for, _use, _purposes) | -0.0019 |
| 26 | Date composition | (20, 23, _-) | -0.0093 |
| 30 | "into account..." | (_into, _account, _possible) | -0.0001 |

Table 3.3: Bi-gram evolution across pretraining steps for OLMo 7B. Each column represents a distinct step, while each row corresponds to a different rank. The table entries are the bi-grams at each step for each rank. The number of tokens seen in association with the pretraining steps is also annotated. The model gradually picks up meaningful bi-grams after starting from random bi-grams (due to random initialization).

| Rank | 0K [#steps] / 0B [#tokens] | 100K / 442B | 200K / 885B | 300K / 1327B | 400K / 1769B | 555K / 2455B |
|---|---|---|---|---|---|---|
| 0 | immortal | 's | at least | & | & | & |
| 1 | ICUirling | at least | 's | at least | its own | its own |
| 2 | ords architect | its own | & | its own | their own | their own |
| 3 | yaml Adam | okerly | your own | your own | at least | his own |
| 4 | 231 next | VENT thanks | its own | their own | your own | make sure |
| 5 | clonal | iums | iums | more than | his own | your own |
| 6 | Charg@{ | you're | you're | can't | 2nd | 2nd |
| 7 | avoir careless | Everything v | 2nd | his own | more than | at least |
| 8 | HOLD worsening | erna already | you guys | 2nd | make sure | more than |
| 9 | Horse dismant | 'my | more than | make sure | can't | iums |

shows marginal improvements and fails to reflect the changes inside the model.

**Identifying the top bigrams.** To assess the model's progression, we extracted bigrams from *OLMo-7B* model checkpoints across 555K pretraining steps. Table 3.3 presents a summary of the top 10 bigrams at different stages of training. Due to space constraints, we only show the top 10 bigrams every 100K steps. Initially, the network exhibits non-sensical bigrams, such as "ICUirling". As training advances, it gradually learns more meaningful combinations, like "at least". This process of acquiring sensible bigrams stabilizes around step 200K, indicating that the model is reaching a level of maturity where the top 10 bigrams capture common meaning.

**Analysing bigram learning speed.** To evaluate the learning speed of these bigrams, we consider the bigrams at the final training step (555K) as the ground-truth. We then chart the hit ratios of these ground-truth bigrams at each pretraining step, as illustrated in Figure 3.5. Interestingly, even though the pretraining loss (the blue curve) shows only minor improvements after the initial 50K steps, the model's acquisition of effective bigrams continues to progress in a steady, consistent manner. This observation aligns with known phenomena in neural network training, such as double-descent and grokking, which highlight the model's ability to improve generalization capabilities even when the loss appears to stagnate [Zhang et al., 2021, Power et al., 2022]. In addition, Figure 3.6 characterizes the total pseudo-joint probability mass of top 1K bigrams from empirical data [Liu et al., 2024b]. We derive a pseudo-joint bigram probability using statistical unigrams from [Liu et al., 2024b]. We observe that the model gradually accumulates probability mass that aligns with the real corpus data distribution. Interestingly, although the overall trend is upward, the mass initially rises sharply from zero, then undergoes two noticeable dips before continuing to increase. This non-monotonic behaviour likely reflects distinct stages in the model's learning dynamics. Early in training, the model quickly captures high-frequency bigrams, resulting in the initial surge. As training progresses, it explores a broader range of token combinations, including less frequent or less relevant bigrams, temporarily redistributing probability mass away from the top 1K bigrams and causing the first dip. The second dip may result from further rebalancing, overfitting to mid-frequency patterns, or transient noise in gradient updates. Contributing factors may include optimization dynamics and noise in the training data, which we leave for future investigation. Eventually, the model reallocates probability mass more accurately and converges toward the empirical distribution, resuming its upward trajectory.

**Learning schemes for different bigrams.** To understand if there are any differences between the learning schemes of different bigrams, we can trace the progression of the bigram scores for selected bigrams. Figure 3.8 provides a visual comparison of how different bigrams are promoted or suppressed during the pretraining process. We analyse bigrams that exhibit different mapping relationships between the first and second tokens, inspired by the one-to-one, one-to-many, and many-to-many relational analysis in the knowledge graph literature [Lacroix et al., 2018]. For example, "at least" is a few-to-many bigram: there are many possible tokens that can follow "at", but relatively few that commonly precede "least". The different slopes and levels of the lines indicate varying rates of learning for the respective bigrams. We observe that, the model first acquires random bigrams due to random parameter initialisation. These random bigrams, like "`ICUirling`" and "`VENT thanks`", are quickly suppressed in the early steps and never regain high scores. In contrast, few-to-many bigrams like "`at least`" are first promoted to very high scores but then get suppressed perhaps due to the model seeing more of the scope of the token "at". One-to-one bigrams like "`&`" (HTML code) are gradually promoted and stabilize. Many-to-many bigrams like "`make sure`" takes the most time to learn, and the scores are still increasing even at the end of pretraining. Our findings suggest that the training process effectively promotes certain "good" bigrams, but at different paces, where they might be suppressed later depending on their occurrences and linguistic nature. These insights could inform future training strategies, such as targeted training on more relevant bigrams or adjusting the training data to improve the pretraining speed.

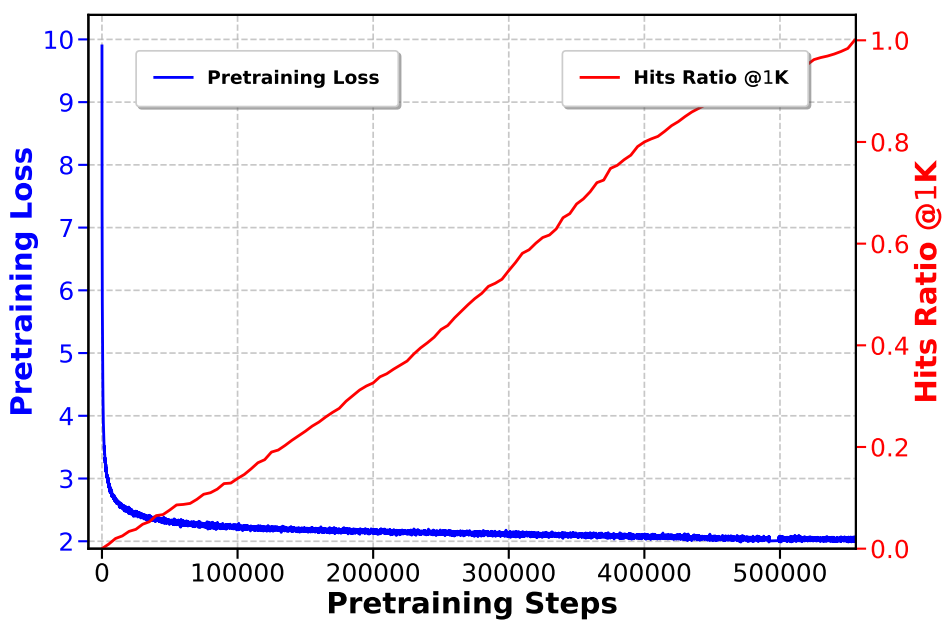

Figure 3.5: Top 1K bigram hit ratios w.r.t. the final step.

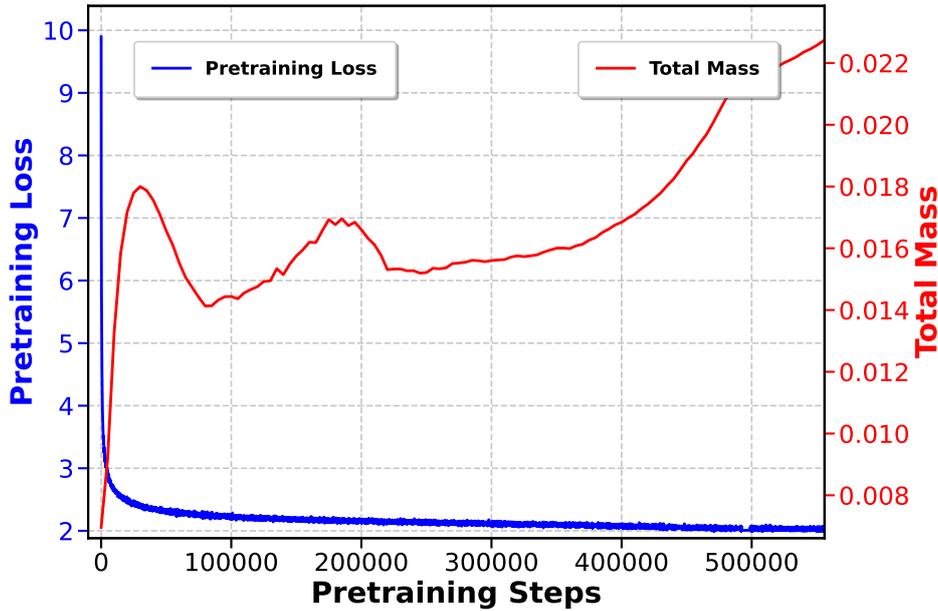

Figure 3.6: Top 1K bigram mass w.r.t. empirical data.

Figure 3.7: Analysis of *OLMo-7B*'s pretraining dynamics by measuring its bigram progression.

### 3.5.3 Use Case 3: Analysing Finetuning Effects

Finetuning is an important phase where the raw pretrained LLMs are guided to perform particular tasks. We would like to understand how the model inner knowledge changes during finetuning processes. While "parameter diff" can be a straightforward solution, n-grams provides an alternative approach, where the diffs are human-readable and directly reflect the change of knowledge retained by the LLMs, similar to how a `diff` command would work in Linux platforms. Such insights would allow us to better decide the mixture of data for finetuning, and the number of steps for finetuning, which are currently a mix of heuristics and trial-and-error.

**Code finetuning promotes coding-relevant bigrams.** We analyse the changes due to code finetuning via *diffing* bigrams extracted from *Llama-2-7B* and its finetuned versions, *Codellama-7B* and *Codellama-Python-7B*. As highlighted in Table 3.4 with orange coloring, the bigram comparison reveals coding-relevant keywords, such as "`**kwargs`", "`getters`" and "`Assertion`", suggesting bigrams can be a tool for verifying if finetun-ing is effective in acquiring relevant knowledge.

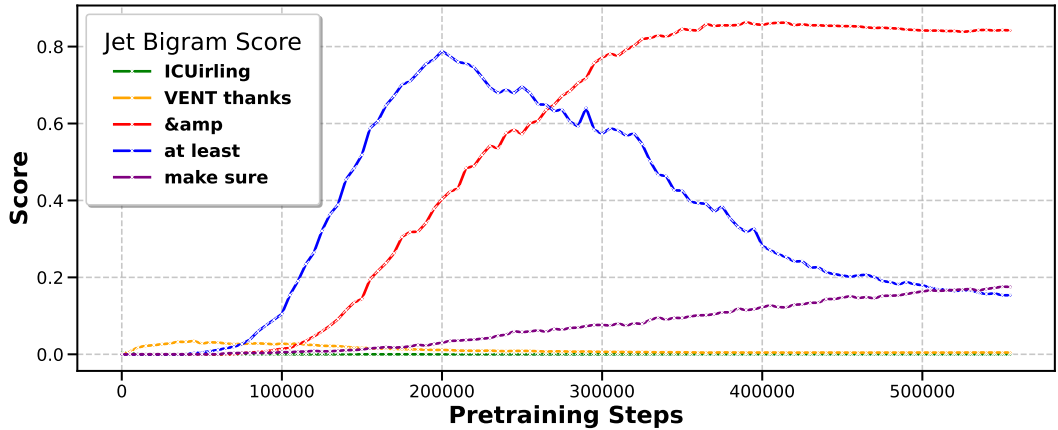

Figure 3.8: Visualization of *OLMo-7B*'s promotion and suppression dynamics of bigrams scores.

**Does RLHF finetuning remove toxicity?** We compare the raw pretrained model, *Llama-2-7B*, with its RLHF version, *Llama-2-7B-Chat*. RLHF alignment [Bai et al., 2022] is widely believed to detoxify LLMs, as indicated by *ToxiGen* scores [Hartvigsen et al., 2022]. However, it remains easy to prompt LLMs to bypass this alignment and produce toxic content [Yi et al., 2024]. In Table 3.5, we demonstrate this with dataset-based toxicity scores on a subset of challenging prompts in the *RealToxicityPrompts* (RTP) dataset [Gehman et al., 2020]: the gap in toxicity potential between the two models *narrows* as we prepend to RTP prompts increasingly "explicit" (short) context. Specifically, for hard context, *Llama-2-7B-Chat* shows a $84\%$ probability of producing toxic content, close to that of *Llama-2-7B*. This suggests that the RLHF model is not completely detoxified but rather hides the toxicity knowledge from the "surface", which however can be easily triggered by specific contexts. To quantify the toxicity knowledge embedded in these models, we use bigram probability scores and calculate the cumulative conditional probability mass for a set of "toxic" bigrams, which are combinations of tokens associated with toxic meanings from a predefined list of keywords. Interestingly, we observe a small change in mass from $0.03445$ to $0.03377$ after RLHF. Thus, although *ToxiGen* score may suggest that the model has been effectively detoxified, the bigram mass reflects retention of toxic knowledge after RLHF, aligning with the scores obtained by introducing medium or hard explicit context and computing a toxicity score (via a second scorer model, [Hanu and Unitary team, 2020]) on *RealToxicityPrompts* dataset [Gehman et al., 2020]. This showcases a potential application of bigrams in constructing *data-free* indices that reveal embedded knowledge, offering complimentary views beyond traditional data-driven benchmark evaluations.

Table 3.4: The bi-grams before and after code fine-tuning. For space constraints, we only show the bi-grams at every 50 ranks among the top 1,000 bi-grams. We highlight the bi-grams that are relevant to coding, such as "**kwargs" a keyword in Python programming. This demonstrates that our method has the capability to extract representative bi-grams that reflect fine-tuning quality.

| Rank | LLAMA2-7B | CodeLLAMA-7B | CodeLLAMA-Python-7B |
|---|---|---|---|
| 0 | (_more, _than) | (_like, wise) | (_like, wise) |
| 50 | (_Now, here) | (_just, ification) | (_Like, wise) |
| 100 | (_system, atically) | (_in, _case) | (_all, udes) |
| 150 | (_all, erg) | (_get, ters) | (_no, isy) |
| 200 | (_on, ions) | (któber, s) | (output, ted) |
| 300 | (_other, world) | (_all, ud) | (Object, ive) |
| 350 | (_Just, ified) | (gebiet, s) | (_as, cii) |
| 400 | (_trust, ees) | (_Protest, s) | (_can, nab) |
| 450 | (_at, he) | (_deploy, ment) | (_transport, ation) |
| 500 | (_book, mark) | (Class, room) | (Tag, ging) |
| 550 | (_from, ) | (_access, ory) | (_personal, ized) |
| 600 | (_WHEN, ever) | (_In, variant) | (_excess, ive) |
| 650 | (_where, about) | (_I, _am) | (_Add, itional) |
| 700 | (ag, ged) | (add, itionally) | (_**, kwargs) |
| 750 | (_he, he) | (_invalid, ate) | (name, plates) |
| 800 | (_all, anto) | (div, ision) | (_select, ive) |
| 850 | (_Tom, orrow) | (_process, ors) | (_Assert, ions) |
| 900 | (_for, ays) | (_Program, me) | (blog, ger) |
| 950 | (_Bach, elor) | (_set, up) | (_can, cellation) |

## 3.6 Discussion

**Limitations.** Isolating partial computations out of the original transformer computation graph can be seen as a truncated Taylor approximation problem, where the center is the portion we want to single out and the variate is the rest of the computation [Chen et al., 2024]. This chapter does not dive into the details of such approximation but rather choose to present the parallel with factorization models, where latent structures can be surfaced similarly as in knowledge base completion, echoing Chapter 2. Besides, the structures we consider are fragments of natural languages, rather than factually meaningful entities or relations. There are substantial evidences that LLMs encode real-world factual structures, for example [Petroni et al., 2019] and [Yang et al., 2024], use curated benchmarks to show pretrained language exhibit certain factual reasoning capability. We would explore similar factual structures in our approach in the future. Additionally, the $n$ in the n-gram structures is bounded by the number of self-attention layers to unfold. For example, when no self-attention is used, we observe $n = 2$; adding a single self-attention layer increases this to $n = 3$. We speculate that there exists a systematic relationship between $n$ and the number of self-attention layers, potentially exponential in nature. Finally, we plan to verify the relationship between the found structures and the pretraining data distribution, which requires large computing resources.

Table 3.5: Toxicity indexes for *Llama-2-7B* and *Llama-2-7B-chat* using different methods: *ToxiGen*, jet bi-grams, and *RealToxicityPrompts* challenge prompting. Higher numbers indicate higher toxicity scores on the corresponding benchmarks and higher toxic knowledge possession for jet bi-grams.

| **Metric** | *Llama-2-7B* | *Llama-2-7B-chat* |
| --- | --- | --- |
| *Standard Benchmarking* | | |
| ToxiGen Score [Hartvigsen et al., 2022] | 21.25 | 0.00 |
| *Prompt-based Benchmarking with RTP Challenging Prompting* [Gehman et al., 2020] | | |
| No Prompt | 38% | 23% |
| Very Mild | 49% | 35% |
| Medium | 64% | 64% |
| Hard | 88% | 84% |
| *Data-free Benchmarking* | | |
| Jet Bi-gram Mass | 0.03445 | 0.03377 |

**Summary.** Large language models are sometimes seen as the victory symbol for the unstructured learning paradigm, where structure curation seems no longer necessary for building a powerful artificial intelligence agent – scaling model sizes on larger unstruc-tured textual corpora is the way. This chapter, however, shows that structures are still the critical ingredients even in the large language models and exposing them is helpful for profiling the knowledge within each model checkpoint. Overall, this chapter provides initial evidence that language modelling objectives, though focused on local context and trained on unstructured data, can recognize and encode structural patterns into the transformer model weights. The key in exposing these inherent structures is to observe that transformers, the typical architecture for large language models, contain portions of computations that resemble factorization based models (FMs). Once trained with LM objectives, these portions of computations capture latent structures in the training data. To expose these structures, this chapter dissects these FMs from the monolithic computation graph of the transformer and derive their corresponding bigrams and trigrams. Akin to how structures help recover the knowledge graph in knowledge base completion, this chapter demonstrates that the uncovered n-gram structures in LLMs help reconstruct the linguistic functions acquired via the models, offering an alternative angle to interpret LLMs in a data-free way. Our case studies demonstrate the potential of using extracted n-gram patterns to debug pretraining progress, verify fine-tuning effects, and detect model toxicity. Looking ahead, LLMs could expose two complementary interfaces: a neural interface for training and prediction, and an n-gram-based symbolic interface for inspection, analysis, and control.

**Implications.** This chapter demonstrates that the same computation, if examined under a new perspective, can lead to new insights that are invisible in the original lens. Using transformers as an example, one view (let us call it the neuron view) is to see it as a special organization of neurons into stacked self-attention and FFNs plus embeddings on both ends; this view allows easy implementations for training on GPUs. Another view (let us call it the behavior view), which is more helpful to interpretability, is to see it as an ensemble of n-gram models describing token transition behavior. Although the neuron view is useful when building the model and training it, it might not be the best level of abstraction for understanding and interpreting model behavior due to the issue of polysemy [Elhage et al., 2022]. We believe that to understand the model better, channelling both the neuron and behaviour view is necessary. Our method provides an initial attempt to do this by reorganizing the neural computations into FMs, which brings structures in behaviours. This new lens enable new findings such that LLMs do not "digest" data points equally – some structures are acquired fast, but the others are always in learning or first learned and then suppressed. These new findings are relevant in the ongoing discourse on AI transparency and trustworthiness.

# Summary of *Structure*

Structures are necessary components for us human beings to grasp physical and artificial worlds[6] with limited representational capacity[7]. The opposite would be instead keeping all, potentially infinite, manifested instances in our memory without any abstraction. With structures, we can effectively perceive reality as a collection of elements and their assembling with relations or hierarchy. By this cognitive process, our minds allow us to know the world and make sense of what is happening in it, ideally leading to consistent and rational behaviour.

Similarly, when building knowledge engines for intelligent agents, structure formation emerges as a key factor. While the structured paradigm explicates structures through data and trains models to capture them, the unstructured paradigm seems to directly rely on specific computational graphs (e.g. deep transformers) and massive pretraining corpora without any predefined schematic structures. This sharp difference in the treatment of structures has led to the tension between the two paradigms, with the current fashion in favour of the unstructured paradigm. Scaling language models on larger datasets tends to be the preferred answer to artificial intelligence while knowledge graphs being left out in the cold.

In this part, however, we reveal a deep connection between the two paradigms. Specifically, we have demonstrated that training models with language modelling objectives picks up structural patterns in embedding-encapsulated models, regardless of whether we are working within a structured or unstructured paradigm. In other words, structures naturally arise when the learning objective attempts to model the surrounding environment through a self-supervised, language modelling approach, as illustrated in Figure

---

[6]E.g. a fictional universe or societal traditions.

[7]The brain operation is bounded by its energy supply from the body and can be sensitive to working environments e.g. the psychological contexts.

3.9.

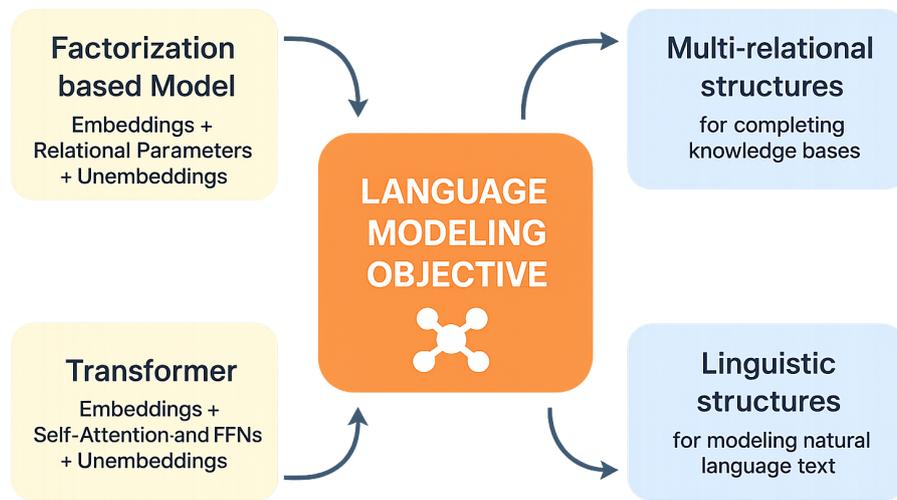

Figure 3.9: Language modelling objective over embedding-encapsulated models drives the acquisition of structures for both structured and unstructured learning paradigms.

When we consider such a structure formation process induced by training with language-modeling objectives, the surficial tension between structured and unstructured paradigms – as exemplified separately with knowledge graphs and large language models – thus shrinks: they both rely on structures, be they in data or in the models, and language modeling helps both form meaningful structures in the models; it is just that one encodes structures deeper in the models via training with sometimes arbitrary or even "magic" corpora choice, while the other uses structures more straightforward. These forming[8] and formed structures are fundamental to both paradigms, and crucial for downstream tasks such as text or graph completion and for broader representation learning.

To conclude this part of the thesis, we would like to prompt the readers' brain with a question on structures in artificial intelligence and perhaps for our own brains: Are structures always helpful for the artificial intelligence agent's performance? And to what extent do these embedded structures advance or constrain model generalization across scenarios? These structures definitely offer efficiency in knowledge representation and retrieval, but at the cost of demanding queries of more formal terms, usually needing to align with the language used to form the structures Can they be that flexible when dealing with informal knowledge in unseen formats or new knowledge?

---

[8] At this moment, someone may be training a large language model, which is forming some structures about our physical world or about some artificial reality.

# Part II  Destructure.

*In the wild, reality is closer to the fresh eyes, where things are observed as they are.*

Structures are the cornerstones for building knowledge engines. Both structured and unstructured paradigms inherently induce global structures into model computations through language modelling objectives as shown in Part I. These global structures may manifest in different forms, such as relationships between drugs or common syntax patterns. For instance, in the structured paradigm, drug-drug interactions can be encoded using tensor factorization models (Chapter 2). And in the unstructured paradigm, linguistic patterns like suffixes may be captured in the embedding-FFN-unembedding computational path in the transformer models (Chapter 3).

However, not all structures contribute positively to intelligent behaviours (Figure 3.10). For a given task, useful structures tend to be specific and concrete. Therefore, an excess of irrelevant structures may reduce efficiency, as identifying and retrieving the useful ones becomes computationally costly. Existing useful structures may become outdated or mismatched with our evolving world. For example, factual knowledge [Petroni et al., 2019, Roberts et al., 2020], such as "*who is the current US president,*" may change after every election. Corresponding structures once embedded in neural weights will stay static unless updated through human intervention. Additionally, new terminologies, like "*social distancing,*" did not exist before the COVID-19 pandemic, and models trained prior to such events may struggle to understand them in the new situations. Moreover, problematic structures can be induced unintentionally due to noise in the training data. For example, gender-biased content in the internet corpora can be captured in the transformer model weights if not filtered from the training data [Bender et al., 2021].

Model editing [Meng et al., 2022, Ilharco et al., 2023], model unlearning [Yao et al., 2024, Liu et al., 2025], retrieval-augmented generation (RAG) [Lewis et al., 2020b], and reinforcement learning from human feedbacks (RLHF) [Christiano et al., 2017, Ouyang et al., 2022] have emerged as remedies to patch these unwanted structures. However, they have certain limitations. The behaviour changes might be local (limited to chosen datasets) and temporary (the biased structures still live in model weights), making them prone to adversarial attacks [Xue et al., 2024] and requiring substantial centralised human steering [Wang et al., 2023].

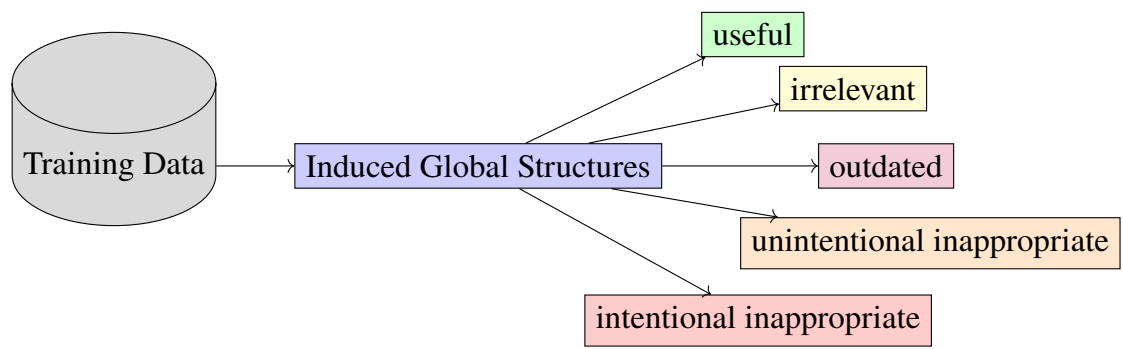

Figure 3.10: Not all structures encoded in the model are of positive roles. Some can be irrelevant, unintentionally inappropriate, intentionally inappropriate, and outdated.

This thesis instead explores an alternative approach, where we want to improve the model's ability to rewire itself rather than relying on external patches. If a model can transition from outdated structures to new ones in a sample-efficient manner, rewiring it with desired behaviours becomes significantly easier when unwanted structures are identified. The core idea is pretty simple: a model can learn to rewire its internal structure by frequently exposing itself to controlled structural dismantling during training. Hence, Part II shifts the focus from structuring, the topic of Part I, to its opposite force – *destructuring*,[9] and seeks to examine its impact on both the structured and unstructured paradigms for building knowledge engines.

**Lessons from natural intelligence**

At first glance, the concept of destructuring in artificial intelligence may appear counter-intuitive. Why would one discard the outcomes of prior learning experiences, especially when these outcomes were achieved through substantial investment, such as prolonged training hours and massive datasets?

To answer this question, we can draw a parallel to natural intelligence, particularly how the brain memory. Memory consists of four primary operations: acquisition, consolidation, forgetting, and retrieval [Berry et al., 2024]. Acquisition and consolidation work together to form stable structures in the memory. Retrieval, in turn, use these structures in combined with contextual clues to surface relevant knowledge.

---

[9] Definition of "destructure" as a verb can be found in wiktionary, meaning dismantling with etymology from *de-* + *structure*. For human cognition, destructuring can be conceived as the act to acknowledge the epistemic uncertainty, admit the state of unknowing about the ever-changing present, and reassess the reality with an open perspective.

However, the structures in human brains are not fixed, but remain *flexible* even into old age. Neural circuits, the brain's primary structures, can be rewired by both internal and external experiences, allowing adaptation to new environments [Park and Huang, 2010], acquiring new skills [Green and Bavelier, 2008], recovery from psychological trauma [Kays et al., 2012], and even compensating for past physical brain damages [Kleim and Jones, 2008]. These phenomenons have been known under the umbrella term of neuroplasticity in neuroscience and other relevant subjects [Fuchs and Flügge, 2014, Costandi, 2016]. While the exact mechanisms behind neuroplasticity remain unclear, recent work suggest that mechanisms at molecular [Bennett et al., 1964], cellular [Rosenzweig, 1996] and network levels [Leuner and Gould, 2010], contribute to the brain's functional flexibility. *Active forgetting*, in particular, has been found to be a key ingredient to neuroplasticity [Anderson and Hulbert, 2021, Berry et al., 2024], which enables memory adaptation to suit particular cognitive and emotional objectives.

When mirroring natural intelligence to artificial intelligence, we find that the operations of acquisition, consolidation, and retrieval are well understood and modeled in both the structured and unstructured AI paradigms. For instance, acquisition and consolidation are achieved through gradient-based optimization guided by self-supervised objectives and subsequent finetuning [Devlin et al., 2019, Radford et al., 2019, Brown et al., 2020], while retrieval relies on training dense passage classifiers [Karpukhin et al., 2020, Lewis et al., 2020b, Reichman and Heck, 2024]. In contrast, forgetting, especially its positive role, has been receiving less attention in AI systems, despite its potential to mimic the neuro-plastic processes that allow for continuous adaptation and learning.

**Inductive inference and generalization to the unseen**

Philosophically, "majority" learning fails to encompass the full spectrum of intelligence. Intelligence is not solely about excelling in frequent or common patterns, often arising from habitualisation, but equally about thriving in long-tailed, atypical behaviours. Throughout human history, individuals those residing the long tail of thoughts, have sometimes been closer to the truth. For instance, Copernicus's heliocentric model, which proposed that Earth orbits the Sun rather than the opposite, defied the dominant belief structures of his time but proved to be a closer representation of reality. Such break-throughs highlight how intelligence can *transcend* the constraints of prior knowledge structures and venture into the unknown.

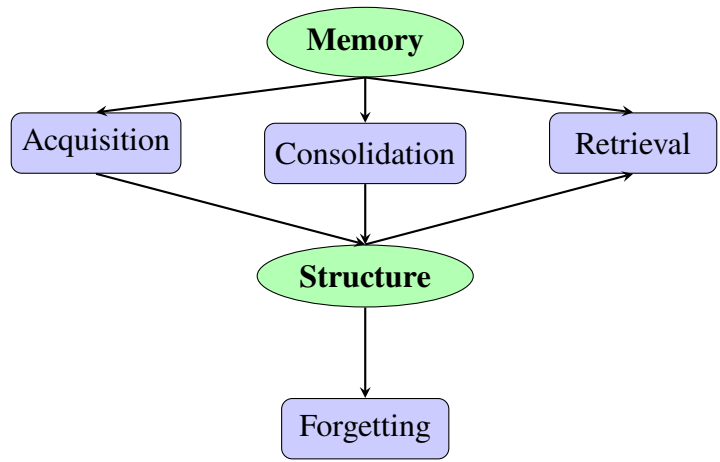

Figure 3.11: Memory consists of four main operations: acquisition, consolidation, retrieval, and forgetting.

From a computational perspective, while modern AI systems achieve superhuman performance on frequently encountered training data, they experience significant degradation when handling long-tailed or unseen data [Buda et al., 2018], regardless of whether they operate in structured or unstructured paradigm. Substantial studies [Finn et al., 2017, Arjovsky et al., 2019, Schölkopf et al., 2021] have been conducted to investigate this issue. The performance gap is often attributed to: (i) the model's tendency to pick up nuances or perform shortcut learning [Geirhos et al., 2020], preventing it from capturing more abstract and generalizable patterns during training; and (ii) at test-time, an over-reliance on computational structures learned from past examples coupled with limited capacity to dynamically incorporate immediate contextual information from the input [Geirhos et al., 2020].

An ideal artificial intelligence system capable of transcending experience diverges fundamentally from deduction systems, which applies established general rules to specific cases (e.g., querying a classic ontological knowledge base [Horrocks et al., 2003, Horrocks, 2005, World Wide Web Consortium (W3C), 2024] in the structured paradigm). Instead, it aligns more closely with *inductive inference*, a form of reasoning that extrapolates beyond immediate evidence. While the rationality of inductive inference remains to be philosophically justified [Hume, 1999], its *amplicative* nature allows bringing in new information into old systems, thereby reshaping an established knowledge landscape. In the unstructured paradigm, large language models exhibits some form of inductive inference through in-context learning, where they solve a task with given new inputs using the prompt composed of a task description and a few example input-output pairs [Brown et al., 2020]. However, this extrapolation is difficult to control, verify, and validate its outcome [Lee et al., 2025], rendering it susceptible to adversarial attacks, prompt choices, and hallucination.[10] Consequently, while promising, current large language models fail to perform systematic inductive inference in a robust and scalable way [Thomm et al., 2024, Wang et al., 2024]. We refer to this shortcoming as a **lack of model plasticity**, which we equate with inadequate inductive inference capacity throughout this thesis. This challenge is particularly acute in larger models, such as foundation models, due to combined factors including their immense model sizes, vast datasets, and high cost of training.

In Part II, we explore whether and how destructuring can improve model plasticity and help build universal knowledge engines that are capable of powering AI in diverse environments. Given the embedding-centric nature of mainstream architectures, where embeddings (and unembeddings) are crucial to encode global structures (Part I), we hypothesize that analysing the embedding learning process can provide deeper insights into structure formation. A core finding of this study is the reinterpretation of embedding learning as a sequence of message-passing operations (Chapter 4). In essence, embeddings function as caches for the outcomes of traversals over structures. However, these cached structures can become overly fixed, limiting generalization. Consequently, the simplest approach to destructure the model, is to clean overly fixed structures in the embeddings regularly, allowing other model components to learn more abstract and generalizable patterns. This new perspective motivates our proposed approach *active forgetting*, a learning mechanism that periodically resets embeddings. By acting as a targeted intervention in standard training, active forgetting improves model plasticity and facilitates inductive inference.

Concretely, we explore active forgetting and its impact in both the structured paradigm

---

[10]The lack of controllability stems from limited understanding of why in-context learning emerges in transformer-based language models. We speculate that it may be linked to deeper structural formation facilitated by self-attention mechanisms, where the toolkit discussed in Chapter 3 could aid in uncovering these processes in the future.

and unstructured paradigm for constructing knowledge engines:

1. **Chapter 4:** In the structured paradigm, we study factorization models, a leading approach in knowledge base completion. While FMs demonstrate strong performance in transductive settings, they fail in inductive scenarios. Our theory into structure formation reveals that entity embeddings in FMs cache infinite rounds of message-passing over the knowledge graph. This insight explains FMs' failure in inductive scenarios, as their reliance on rigid structures (ingrained in the embeddings) overfits to the original graph and impedes generalization to new graphs. To address this, we propose incorporating *active forgetting* into FMs, periodically flushing old values and reloading new ones in embeddings to mitigate the effects of overly rigid structures. This derives a new model that bridges FMs and the message-passing graph neural networks – ReFactor GNNs. Experiments show that ReFactor GNNs improves generalization to novel graphs with unseen nodes.

2. **Chapter 5:** In the unstructured paradigm, we study pretrained language models, the mainstream approach for building universal knowledge engines from purely unstructured data. Pretrained language models often lack plasticity and require large datasets to relearn embeddings for extending to new languages. However, data is scarce for low-resource languages. To address this, we examine cross-lingual transfer in a low-data regime, aiming to pretrain language models that quickly adapt to reasoning tasks for new languages with limited data. Each language represents a distinct environment, requiring inductive generalization. *Active forgetting* again proves effective and practical to incorporate into large scale pretraining processes. Models pretrained with this technique exhibit significantly better adaptation to previously unseen languages with limited data.

Both the structured and unstructured paradigms rely on structure formation to learn representations. By rethinking the procedures underlying structure formation, this part suggests that AI's struggles with generalization stem from overly fixed internal structures. To address this, Part II proposes that destructuring can serve as a way to counterbalance the rigidity introduced by these overly-fixed structures embedded in model computations. Given the widespread use of embedding layers in structured and unstructured paradigms, *active forgetting* emerges as a simple yet powerful mechanism for implementing destructuring, where destructuring becomes as easy as resetting embedding values periodically. This approach proves effective across both the structured and unstructured paradigms, providing a unified framework to improve model plasticity and enable models to perform well in dynamic environments. Our results demonstrate that universal knowledge engines, which must achieve the plasticity and adaptability necessary for robust performance in diverse, real-world settings, will benefit significantly from incorporating *active forgetting* or similar destructuring techniques.

# Chapter 4

# Inductive Knowledge Graph Learning with Active Forgetting

*A version of this work was previously presented at a peer-reviewed conference. Please refer to [Chen et al., 2022] for full citation.*

Knowledge graphs form the backbone of modern knowledge engines, enabling AI systems to organize, retrieve, and reason over structured information. Among the tools that enrich and sustain these knowledge graphs, Factorization Models (FMs), such as DistMult, have emerged as a cornerstone in Knowledge Graph Completion (KGC), a task focused on predicting missing relationships between entities. In transductive scenarios, Factorization Models (FMs) often surpass Graph Neural Networks (GNNs), emerging as indispensable pillars of knowledge graphs, completing them and elevating their utility as a foundational source of knowledge for downstream tasks.

However, FMs struggle in inductive scenarios, where they can not generalize to unseen nodes or incorporate node features effectively. To transfer FM's transductive performance to inductive scenarios, we observe that FMs' structure formation rely highly on the embeddings. These embeddings, when optimized through gradient descent, can be reinterpreted as a sequence of message-passing rounds across the knowledge graph. In other words, embeddings essentially act as a historical cache of node states, tracing structural traversals over the knowledge graph.

This perspective reveals a fundamental limitation about FMs: when trained to convergence, FMs tend to capture excessive global graph structures through infinite rounds of implicit message-passing, often far exceeding the graph's natural radius ($L \to \infty$). While extensive structuring yields strong transductive performance, it also results in overly constrained representations that hinder generalization from training graphs to new, unseen graphs. To destructure rigid representations, we propose a simple yet powerful mechanism: *active forgetting*. By periodically clearing and reloading new node embeddings, this operator truncates the infinite rounds of message-passing, resetting the model's memory of past computations over the nodes. This reset forces the model to focus on the local neighbourhood information, which enables inductive reasoning for previously unseen or forgotten nodes. Mathematically, this approach synthesizes the strengths of FMs and GNNs into a unified framework, which we call REFACTOR GNNs.

Evaluations across standard KGC benchmarks demonstrate that REFACTOR GNNs maintain the transductive performance of FMs while achieving state-of-the-art inductive performance with significantly fewer parameters. REFACTOR GNNs bridge the gap between FMs and GNNs, providing a unified architecture for robust knowledge graph representation learning, supporting AI agents' dynamic knowledge needs in the wild.

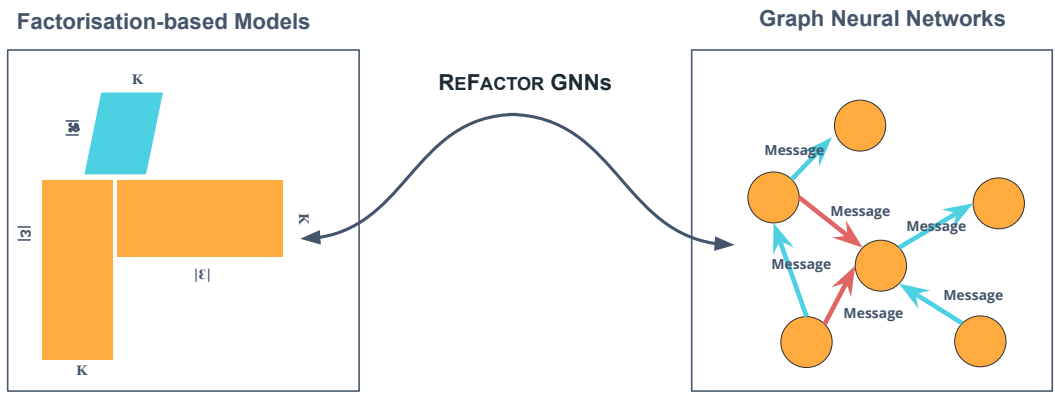

Figure 4.1: REFACTOR GNN bridges factorization-based models and graph neural networks by reformulating gradient descents over entity embeddings as message-passing rounds.

## 4.1 Factorization Meets Message-Passing

In recent years, machine learning on graphs has attracted significant attention due to the abundance of graph-structured data and developments in graph learning algorithms. Graph Neural Networks (GNNs) have demonstrated state-of-the-art performance for many graph-related problems, such as node classification [Kipf and Welling, 2016] and graph classification [Gilmer et al., 2017]. Their main advantage is that they can easily be applied in an inductive setting: generalising to new nodes and graphs without re-training. However, despite many attempts at applying GNNs for multi-relational link prediction tasks such as Knowledge Graph Completion [Nickel et al., 2016c], there are still few positive results compared to more traditional factorisation-based models (FMs) [Yang et al., 2015b, Trouillon et al., 2016]. As it stands, GNNs, after resolving reproducibility concerns, either deliver significantly lower performance [Nathani et al., 2019, Sun et al., 2020a] or yield negligible performance gains at the cost of highly sophisticated architecture designs [Xu et al., 2020b]. A notable exception is NBFNet [Zhu et al., 2021], but even here improvements come at the price of high computational inference costs compared to FMs. Furthermore, it is unclear how NBFNet could incorporate node features, which, as we will see in this work, leads to remarkably lower performance in an inductive setting. On the other hand FMs, despite being a simpler architecture, have been found to be very accurate for knowledge graph completion when coupled with appropriate training strategies [Ruffinelli et al., 2020] and training objectives [Lacroix et al., 2018, Chen et al., 2021]. However, they also come with shortcomings in that they, unlike GNNs, can not be applied in an inductive setting.

Given the respective strengths and weaknesses of FMs and GNNs, we wonder *whether we can bridge these two seemingly different model categories* so that we can develop knowledge graph completion models that generalize to unseen graphs. While exploring this question, we make the following contributions:

- By reformulating gradient descent on node embeddings using message-passing primitives, we show a practical connection between FMs and GNNs, in that: FMs can be treated as a special instance of GNNs, but with infinite neighbourhood, layer-wise training and a global normaliser.[1]

---

[1]The traditional view is that *the transductive nature of FMs stem from their need to retrain on new*

- Based on this connection, we propose a new family of architectures, referred to as REFACTOR GNNs, which interpolates between FMs and GNNs. In essence, REFACTOR GNNs inductivise FMs by using a *finite* number of message-passing layers, and incorporating node features.

- Through an empirical investigation across 15 well-established inductive and transductive benchmarks, we find that REFACTOR GNNs achieve state-of-the-art inductive performance and comparable transductive performance to FMs, despite using an order of magnitude fewer parameters than GNNs.

## 4.2 Literature Review: Multi-relational Graph Learning, FMs, and GNNs

**Multi-Relational Graph Representation Learning** Multi-relational graph representation learning concerns graphs with various edge types. Another relevant line of work would be representation learning over heterogeneous graphs, where node types are also considered. Previous work on multi-relational graph representation learning focused either on FMs [Nickel et al., 2011b, Trouillon et al., 2016, Yang et al., 2015b, Lacroix et al., 2018, Nickel et al., 2016c, Dettmers et al., 2018, Nguyen et al., 2018, Chen et al., 2021] or GNN-based models [Schlichtkrull et al., 2018, Xu et al., 2020a, Zhang et al., 2020, Li et al., 2021b]. Similar to a recent finding in a benchmark study over heterogeneous GNNs [Lv et al., 2021], where the best choices of GNNs for heterogeneous graphs seem to regress to simple homogeneous GNN baselines, the progress of multi-relational graph representation learning also mingles with FMs, the classic multi-relational link predictors. Recently, FMs were found to be significantly more accurate than GNNs for KGC tasks, when coupled with specific training strategies [Ruffinelli et al., 2020, Jain et al., 2020b, Lacroix et al., 2018]. While more advanced GNNs [Zhu et al., 2021] for KBC are showing promise at the cost of extra algorithmic complexity, little effort has been devoted to establishing links between plain GNNs and FMs, which are strong multi-relational link predictors despite their simplicity. Our work aims to *align* GNNs with FMs so that we can combine the strengths from both families of models.

---

*nodes*, a view which we further underpin by also observing that *FMs are not inductive due to the need for infinite layers of on-the-fly message-passing*.

**Relationships between FMs and GNNs**   A very recent work [Srinivasan and Ribeiro, 2020] builds a theoretical link between structural GNNs and node (positional) embeddings. However, on one end of the link, the second model category encompasses not merely factorisation-based models but also many practical graph neural networks, between which the connection is unknown. Our work instead offers a more practical link between positional node embeddings produced by FMs and positional node embeddings produced by GNNs, while at the same time focusing on KGC. Beyond FMs in KGC, using graph signal processing theory, Shen et al. [2021] show that matrix factorisation (MF) based recommender models correspond to ideal low-pass graph convolutional filters. They also find infinite neighbourhood coverage in MF although using a different approach and focusing on a different domain in contrast to our work.

**Message-passing**   Message-passing is itself a broad terminology, it is generally discussed under two different contexts. Firstly, as a computational technique, message passing allows recursively decomposing a global function into simple local, parallelisable computations [MacKay, 2003], thus being widely used for solving inference problems in a graphical model. Specifically, we note that message passing-based inference techniques were proposed for matrix completion-based recommendation [Kim et al., 2010] and Bayesian Boolean data decomposition [Ravanbakhsh et al., 2016] in the pre-deep-learning era. Secondly, as a paradigm of parameterising learnable functions over *graph-structured data*, message-passing has recently been used to provide a unified reformulation [Gilmer et al., 2017] for various GNN architectures, including Graph Attention Networks [Veličković et al., 2018], Gated Graph Neural Networks [Li et al., 2016], and Graph Convolutional Networks [Kipf and Welling, 2016]. In this work, we show that FMs can also be cast as a special type of message-passing GNNs by considering the gradient descent updates [Bottou, 2012] over node embeddings as message-passing operations between nodes. To the best of our knowledge, our work is the first to provide such connections between FMs and message-passing GNNs. We show that FMs can be seen as instances of GNNs, with a characteristic feature about the nodes being considered during the message-passing process: our REFACTOR GNNs can be seen as using an *Augmented Message-Passing* process on a dynamically re-wired graph [Veličković, 2022].

## 4.3 Formalizing FMs and GNNs for KGC

Knowledge Graph Completion (KGC) [Nickel et al., 2016b], also known as knowledge base completion (KBC), is a canonical task of multi-relational link prediction. The goal is to predict missing edges given existing edges. Formally, a knowledge graph contains a set of entities (nodes), $\mathcal{E} = \{1, \ldots, |\mathcal{E}|\}$, a set of relations (edge types) $\mathcal{R} = \{1, \ldots, |\mathcal{R}|\}$, and a set of typed edges between the entities $\mathcal{T} = \{(v_i, r_i, w_i)\}_{i=1}^{|\mathcal{T}|}$, where each triplet $(v_i, r_i, w_i)$ indicates a relationship of type $r_i \in \mathcal{R}$ between the *subject* $v_i \in \mathcal{E}$ and the *object* $w_i \in \mathcal{E}$. Given a node $v$, we denote its *outgoing* 1-hop neighbourhood as the set of relation-object pairs $\mathcal{N}_+^1[v] = \{(r, o) \mid (v, r, o) \in \mathcal{T}\}$, its *incoming* 1-hop neighbourhood as the set of subject-relation pairs $\mathcal{N}_-^1[v] = \{(r, s) \mid (s, r, v) \in \mathcal{T}\}$, and its total neighbourhood as the union of the two $\mathcal{N}^1[v] = \mathcal{N}_+^1[v] \cup \mathcal{N}_-^1[v]$. We denote the neighbourhood of $v$ under a specific relation $r$ as $\mathcal{N}_\pm^1[r, v]$. Entities may come with features $X \in \mathbb{R}^{|\mathcal{E}| \times K}$ for describing them, such as textual encodings of their names and/or descriptions. Given a (training) knowledge graph, KGC is evaluated by answering $(v, r, ?)$-style queries i.e. predicting the object given the subject and relation in the triplet. And queries like $(?, r, v')$ are answered using inverse queries $(v', r^{-1}, ?)$ in this work, following [Lacroix et al., 2018].

Following the 1vsAll setting used in Chapter 2 and Ruffinelli et al. [2020], multi-relational link prediction models can be trained via maximum likelihood, by fitting a parameterized conditional categorical distribution $P_\theta(w \mid v, r)$ over the candidate objects of a relation, given the subject $v$ and the relation type $r$:

$$P_\theta(w | \mathbf{v}, \mathbf{r}) = \frac{\exp \Gamma_\theta(\mathbf{v}, \mathbf{r}, w)}{\sum_{u \in \mathcal{E}} \exp \Gamma_\theta(\mathbf{v}, \mathbf{r}, u)} = \text{Softmax}(\Gamma_\theta(\mathbf{v}, \mathbf{r}, \cdot))[w]. \qquad (4.1)$$

Here $\Gamma_\theta : \mathcal{E} \times \mathcal{R} \times \mathcal{E} \to \mathbb{R}$ is a *scoring function*, which, given a triplet $(v, r, w)$, returns the likelihood that the corresponding edge appears in the knowledge graph.

We illustrate our derivations using DistMult [Yang et al., 2015b] as the score function $\Gamma$ and defer extensions to general score functions, e.g. ComplEx [Trouillon et al., 2016], to the appendix. For DistMult, the score function $\Gamma_\theta$ is defined as the trilinear dot product of the vector representations corresponding to the subject, relation, and object of the

triplet:

$$\Gamma_\theta(v, r, w) = \langle f_\phi(v), f_\phi(w), g_\psi(r) \rangle = \sum_{i=1}^{K} f_\phi(v)_i f_\phi(w)_i g_\psi(r)_i, \qquad (4.2)$$

where $f_\phi : \mathcal{E} \to \mathbb{R}^K$ and $g_\psi : \mathcal{R} \to \mathbb{R}^K$ are learnable maps parameterised by $\phi$ and $\psi$ that encode entities and relation types into $K$-dimensional vector representations, and $\theta = (\phi, \psi)$. We will refer to $f$ and $g$ as the entity and relation *encoders*, respectively. If we define the data distribution as $P_D(x) = \frac{1}{|\mathcal{T}|} \sum_{(v,r,w) \in \mathcal{T}} \delta_{(v,r,w)}(x)$, where $\delta_{(v,r,w)}(x)$ is a Dirac delta function at $(v, r, w)$, then the objective is to learn the model parameters $\theta$ by minimising the expected negative log-likelihood $\mathcal{L}(\theta)$ of the ground-truth entities for the queries $(v, r, ?)$ obtained from $\mathcal{T}$:

$$\arg\min_\theta \mathcal{L}(\theta) \quad \text{where} \quad \begin{aligned} \mathcal{L}(\theta) &= -\mathbb{E}_{x \sim P_D}[\log(P_\theta(w|v,r)] \\ &= -\frac{1}{|\mathcal{T}|} \sum_{(v,r,w) \in \mathcal{T}} \log P_\theta(w|v,r). \end{aligned} \qquad (4.3)$$

During inference, we use $P_\theta$ for determining the plausibility of links not present in the training graph.

### 4.3.1 Factorisation-based Models for KGC

In factorisation-based models, which we assume to be DistMult, the entity encoder $f_\phi$ and the relation encoder $g_\psi$ are simply parameterised as look-up tables, associating each entity and relation with a continuous distributed representation:

$$f_\phi(v) = \phi[v], \; \phi \in \mathbb{R}^{|\mathcal{E}| \times K} \quad \text{and} \quad g_\psi(r) = \psi[r], \; \psi \in \mathbb{R}^{|\mathcal{R}| \times K}. \qquad (4.4)$$

The corresponding score function is then given by

$$\Gamma_\theta(v, r, w) = \langle \phi[v], \phi[w], g(r) \rangle = \sum_{i=1}^{K} \phi[v]_i \phi[w]_i \psi[r]_i. \qquad (4.5)$$

### 4.3.2 GNN-based Models for KGC

GNNs were originally proposed for node or graph classification tasks [Gori et al., 2005, Scarselli et al., 2009]. To adapt them to KGC, previous work has explored two different paradigms: *node-wise entity representations* [Schlichtkrull et al., 2018] and *pair-wise entity representations* [Teru et al., 2020, Zhu et al., 2021]. Though the latter paradigm has shown promising results, it requires computing representations for all pairs of nodes, which can be computationally expensive for large-scale graphs with millions of entities. Additionally, node-wise representations allow for using a single evaluation of $f_\phi(v)$ for multiple queries involving $v$, resulting in faster batch evaluation.

Models based on the first paradigm differ from pure FMs only in the entity encoder and lend themselves well for a fair comparison with pure FMs. We will therefore focus on this class and leave the investigation of pair-wise representations to future work. Let $q_\phi : \mathcal{G} \times \mathcal{X} \to \bigcup_{S \in \mathbb{N}^+} \mathbb{R}^{S \times K}$ be a GNN encoder, where $\mathcal{G} = \{G \mid G \subseteq \mathcal{E} \times \mathcal{R} \times \mathcal{E}\}$ is the set of all possible multi-relational graphs defined over $\mathcal{E}$ and $\mathcal{R}$, and $\mathcal{X}$ is the input feature space, respectively. Then we can set $f_\phi(v) = q_\phi(\mathcal{T}, X)[v]$. Following the standard message-passing framework [Gilmer et al., 2017, Battaglia et al., 2018, Hamilton] used by the GNNs, we view $q_\phi = q^L \circ ... \circ q^1$ as the recursive composition of $L \in \mathbb{N}^+$ layers that compute intermediate representations $\{h^l\}$ for $l \in \{1, \ldots, L\}$ with $h^0 = X$ for all entities in the KG. Each layer $q^l$ producing representation $h_l$ is made up of the following three functions:

1. A *message function* $q_\text{M}^l : \mathbb{R}^K \times \mathcal{R} \times \mathbb{R}^K \to \mathbb{R}^K$ that computes the message along each edge. Given an edge $(v, r, w) \in \mathcal{T}$, the message function $q_\text{M}^l$ not only makes use of the node states $h^{l-1}[v]$ and $h^{l-1}[w]$ (as in standard GNNs) but also uses the relation $r$; denote the message as

$$m^l[v, r, w] = q_\text{M}^l \left( h^{l-1}[v], r, h^{l-1}[w] \right) ;$$

2. An *aggregation function* $q_\text{A}^l : \bigcup_{S \in \mathbb{N}} \mathbb{R}^{S \times K} \to \mathbb{R}^K$ that aggregates all messages from the 1-hop neighbourhood of a node; denote the aggregated message as

$$z^l[v] = q_\text{A}^l \left( \{ m^l[v, r, w] \mid (r, w) \in \mathcal{N}^1[v] \} \right) ;$$

3. An *update function* $q_U^l : \mathbb{R}^K \times \mathbb{R}^K \to \mathbb{R}^K$ that produces the new node states $h^l$ by combining previous node states $h^{l-1}$ and the aggregated messages $z^l$:

$$h^l[v] = q_U^l(h^{l-1}[v], z^l[v]).$$

Different parametrisations of $q_M^l$, $q_A^l$, and $q_U^l$ lead to different GNNs. For example, R-GCNs [Schlichtkrull et al., 2018] define the $q_M^l$ function using per-relation linear transformations $m^l[v, r, w] = \frac{1}{\mathcal{N}^1[r,v]} W_r^l h^{l-1}[w]$, where $W_r^l$ denotes the weight matrix associated with relation $r$ and $\mathcal{N}^1[r, v]$ represents the degree of $v$ under relation $r$; $q_A^l$ is implemented by a summation and $q_U^l$ is a non-linear transformation $h^l[v] = \sigma(z^l[v] + W_0^l h^{l-1}[v])$, where $\sigma$ is the sigmoid function. For each layer, the learnable parameters are $\{W_r^l\}_{r \in \mathcal{R}}$ and $W_0^l$, all of which are matrices in $\mathbb{R}^{K \times K}$. Sometimes applying GNNs over an entire graph might not be feasible due to the size of the graph. Hence, in practice, $f_\phi(v)$ can be approximated with sampled sub-graphs [Hamilton et al., 2017, Zou et al., 2019, Zeng et al., 2020], such as $L$-hop neighbourhood around node $v$ denoted as $\mathcal{N}^L[v]$:

$$f_\phi(v) = q_\phi(\mathcal{T}_{\mathcal{N}^L[v]}, X_{\mathcal{N}^L[v]})[v]. \qquad (4.6)$$

## 4.4 Implicit Message-Passing in FMs

The sharp difference in analytical forms might give rise to the misconception that GNNs incorporate message-passing over the neighbourhood of each node (up to $L$-hops), while FMs do not. In this work, we show that by explicitly considering the training dynamics of FMs, we can uncover and analyse the hidden message-passing mechanism within FMs. In turn, this will lead us to the formulation of a novel class of GNNs well suited for multi-relational link prediction tasks (Section 4.5). Specifically, we propose to interpret the FMs' optimisation process of their objective as the entity encoder. After randomly initialising the parameters $\phi$ of the look-up table, FMs are typically trained to minimise the loss $\mathcal{L}$ (Equation 4.3). If we consider, for simplicity, a gradient descent training dynamic, then the entity encoder operating on a given node $v$, $f_{\phi^t}(v)$, can be rewritten as the outcome of a series of gradient descent steps:

$$\begin{aligned}
f_{\phi^t}(v) &= \phi^t[v] \\
&= \mathrm{GD}^t(\phi^{t-1}, \mathcal{T})[v] \\
&= \mathrm{GD}^t \circ \mathrm{GD}^{t-1}(\phi^{t-2}, \mathcal{T})[v] \\
&= \underbrace{\mathrm{GD}^t \circ \cdots \circ \mathrm{GD}^1}_{t\,\text{gradient steps}}(\phi^0, \mathcal{T})[v]
\end{aligned} \qquad (4.7)$$

where $\phi^t$ is the embedding vector at the $t$-th step, $t \in \mathbb{N}^+$ is the total number of training iterations, and $\phi^0$ is a random initialisation of the look-up table. GD is the gradient descent operator, which we can expand by substituting in the objective $\mathcal{L}$ (Equation 4.3):

$$\mathrm{GD}(\phi, \mathcal{T}) = \phi - \beta \nabla_\phi \mathcal{L} \qquad (4.8)$$

$$= \phi + \alpha \sum_{(v,r,w)\in\mathcal{T}} \frac{\partial \log P(w|v,r)}{\partial \phi}, \qquad (4.9)$$

where $\alpha = \beta\,|\mathcal{T}|^{-1}$ with a learning rate $\beta > 0$. We now dissect Equation 4.8 in two different but equivalent ways. In the first, which we dub the *edge view*, we separately consider each addend of the gradient $\nabla_\phi \mathcal{L}$. In the second, we aggregate the contributions from all the triplets to the update of a particular node. With this latter decomposition, which we call the *node view*, we can explicate the message-passing mechanism at the core of the FMs. While the edge view suits a vectorised implementation better, the node view further exposes the information flow among nodes, allowing us to draw an analogy to message-passing GNNs.

### 4.4.1 The Edge View

Each addend of Equation 4.8 corresponds to a single edge $(v, r, w) \in \mathcal{T}$ and contributes to the update of the representation of all nodes. The update on the representation of the subject $v$ contributed by this edge can be written as:

$$\begin{aligned}
\text{GD}(\phi, \{(v, r, w)\})[v] &= \phi[v] + \alpha \frac{\partial \log P(w|v, r)}{\partial \phi[v]} \\
&= \phi[v] + \alpha \frac{\partial \log \frac{\exp \Gamma(v,r,w)}{\sum_{u \in \mathcal{E}} \exp \Gamma(v,r,u)}}{\partial \phi[v]} \\
&= \phi[v] + \alpha \left( \frac{\partial \Gamma(v, r, w)}{\partial \phi[v]} - \sum_{u \in \mathcal{E}} P(u|v, r) \frac{\partial \Gamma(v, r, u)}{\partial \phi[v]} \right) \\
&= \phi[v] + \alpha \left( \underbrace{g(r) \odot \phi[w]}_{w \to v} - \underbrace{\sum_{u \in \mathcal{E}} P_\theta(u|v, r) g(r) \odot \phi[u]}_{u \to v} \right).
\end{aligned}$$

Step two follows by substituting the softmax expression for the conditional probability (Equation 4.1) and take gradients of the log softmax, where the critical part is the treatment of the gradient of the log partition function:

$$\begin{aligned}
\frac{\partial \log \sum_u \exp(\Gamma(\cdot, u))}{\partial \phi[v]} &= \frac{1}{\sum_u \exp(\Gamma(\cdot, u))} \left[ \sum_u \exp(\Gamma(\cdot, u)) \frac{\partial \Gamma}{\partial \phi[v]} \right] \\
&= \sum_u \frac{\exp(\Gamma(\cdot, u))}{\sum_u \exp(\Gamma(\cdot, u))} \frac{\partial \Gamma}{\partial \phi[v]} = \sum_u P(u|\cdot) \frac{\partial \Gamma}{\partial \phi[v]}.
\end{aligned}$$

Step three results from taking the gradient of the score function $\Gamma$ (Equation 4.5):

$$\frac{\partial \Gamma(v, r, w)}{\partial \phi[v]} = \frac{\langle \phi[v], \phi[w], g(r) \rangle}{\partial \phi[v]} = g(r) \odot \phi[w].$$

We discuss the meaning underlying this decomposition. The $w \to v$ term represents information flow from $w$ (a positive neighbour of $v$) to $v$, thereby increasing the score of the gold triplet $(v, r, w)$. In contrast, the $u \to v$ term captures information flow from global pseudo-negative nodes $\{u \in \mathcal{E}\}$, which serves to decrease the scores of triplets $(v, r, u)$. Fundamentally, the term $u \to v$ is induced by the partition function in the denominator of the conditional probability (Equation 4.1). Due to the 1vsAll setting, the conditional probability $P_\theta(w \mid v, r)$ is computed over all entities in $\mathcal{E}$. As a result, the model incorporates signals from pseudo-negative edges linking $v$ across the entire vocabulary $\{u \in \mathcal{E}\}$ when updating the representation of the subject $v$. This negative contribution can be seen as a global repulsion to ensure that truly informative neighbours maintain strong influence. Note that the "negative" here is about the non-existing edges that are automatically considered due to the 1vsAll loss. This is different from the negative neighbourhood, which is from existing edges. There negative sign $\mathcal{N}^1_-[\mathbf{v}] = \{(r, s) \mid (s, r, \mathbf{v}) \in \mathcal{T}\}$ means the in-coming as opposed to outgoing.

Similarly, for the object $w$, we have

$$\text{GD}(\phi, \{(v, r, w)\})[w] = \phi[w] + \alpha \underbrace{(1 - P_\theta(w|v, r))\, g(r) \odot \phi[v]}_{v \to w},$$

where, again, the $v \to w$ term indicates information flow from the neighbouring node $v$. Finally, for the nodes other than $v$ and $w$, we have

$$\text{GD}(\phi, \{(v, r, w)\})[u] = \phi[u] + \alpha \left( \underbrace{-P_\theta(u|v, r)\phi[v] \odot g(r)}_{v \to u} \right).$$

### 4.4.2 The Node View

To fully uncover the message-passing mechanism of FMs, we now focus on the gradient descent operation over a single node $v \in \mathcal{E}$, referred to as the *central node* in the GNN literature. Recalling Equation 4.8, we have:

$$\text{GD}(\phi, \mathcal{T})[v] = \phi[v] + \alpha \sum_{(v,r,w) \in \mathcal{T}} \frac{\partial \log P(\mathbf{w} \mid \mathbf{v}, \mathbf{r})}{\partial \phi[v]}, \tag{4.10}$$

which aggregates the information stemming from the updates presented in the edge view. The next theorem describes how this total information flow to a particular node can be recast as an instance of message passing (cf. Section 4.3.2). We defer the full proof to Appendix B.1.1 and present a proof sketch here.

**Theorem 4.4.1** (Message passing in FMs)**.** *The gradient descent operator* GD *(Equation 4.10) on the node embeddings of a DistMult model (Equation 4.4) with the maximum* *likelihood objective (Equation 4.3) and a multi-relational graph $\mathcal{T}$ defined over entities $\mathcal{E}$ induces a message-passing operator whose composing functions are:*

$$q_{\text{M}}(\phi[v], r, \phi[w]) = \begin{cases} \phi[w] \odot g(r) & \text{if } (r, w) \in \mathcal{N}_+^1[v], \\ (1 - P_\theta(v|w, r))\phi[w] \odot g(r) & \text{if } (r, w) \in \mathcal{N}_-^1[v]; \end{cases} \quad (4.11)$$

$$q_{\text{A}}(\{m[v, r, w] : (r, w) \in \mathcal{N}^1[v]\}) = \sum_{(r,w) \in \mathcal{N}^1[v]} m[v, r, w]; \quad (4.12)$$

$$q_{\text{U}}(\phi[v], z[v]) = \phi[v] + \alpha z[v] - \beta n[v], \quad (4.13)$$

*where, defining the sets of triplets $\mathcal{T}^{-v} = \{(s, r, o) \in \mathcal{T} \ : \ s \neq v \wedge o \neq v\}$,*

$$n[v] = \frac{|\mathcal{N}_+^1[v]|}{|\mathcal{T}|} \mathbb{E}_{P_{\mathcal{N}_+^1[v]}} \mathbb{E}_{u \sim P_\theta(\cdot|v,r)} g(r) \odot \phi[u] + \frac{|\mathcal{T}^{-v}|}{|\mathcal{T}|} \mathbb{E}_{P_{\mathcal{T}^{-v}}} P_\theta(v|s, r) g(r) \odot \phi[s], \quad (4.14)$$

*where $P_{\mathcal{N}_+^1[v]}$ and $P_{\mathcal{T}^{-v}}$ are the empirical probability distributions associated to the respective sets.*

**Proof Sketch** (Proof Sketch for Theorem 4.4.1). *We outline how a single step of gradient descent (Equation 4.10) on the node embeddings of a DistMult model (Equation 4.4) with a softmax-based likelihood (Equation 4.3) induces a message-passing operator.*

*Setup and Assumptions.* *We consider a multi-relational graph $\mathcal{T}$ over entities $\mathcal{E}$ and relations $\mathcal{R}$. Each entity $v \in \mathcal{E}$ is associated with an embedding $\phi[v]$. The DistMult model defines the conditional probability of a tail entity given a head and relation as:*

$$P(w \mid v, r) = \frac{\exp(\Gamma(v, r, w))}{\sum_{u \in \mathcal{E}} \exp(\Gamma(v, r, u))},$$

*where $\Gamma(v, r, w) = \langle \phi[v], g(r), \phi[w] \rangle$. We assume no self-loops (i.e., $(v, r, v)$ not in $\mathcal{T}$).*

*Gradient Decomposition.* *The gradient of the log-likelihood w.r.t. $\phi[v]$ is a sum over the triples comprising the training graph*

$$\sum_{(\text{v},\text{r},\text{w}) \in \mathcal{T}} \frac{\partial \log P(\text{w} \mid \text{v}, \text{r})}{\partial \phi[v]},$$

*which splits into:*

- *Outgoing edges of $v$: $(v, r, \text{w})$ yield terms pulling $\phi[v]$ toward $\phi[\text{w}] \odot g(r)$. At the*

*same time, the partition function induced by the denominator of the 1vsAll loss yields terms pushing $\phi[v]$ away from global pseudo-negative entities $g(r) \odot \phi[u]$ for $u \in \mathcal{E}$ modulated by $P(u|v, r)$.*

- *Incoming edges of $v$: $(\mathrm{w}, r, v)$ yield terms pulling $\phi[v]$ towards $g(r) \odot \phi[\mathrm{w}]$ modulated by $1 - P(v|\mathrm{w}, r)$.*

- *Non-local edges: $(s, r, o)$ Triplets not involving $v$ but in the training graph still affect $\phi[v]$ due to $v$'s appearance in the partition function, producing a term proportional to $-P(v|s, r) g(r) \odot \phi[s]$.*

***Message-Passing Form.*** *Collecting these categories and regrouping them based on if the term comes from $v$'s neighbourhood, yielding:*

- *A message function $q_\mathrm{M}(\phi[v], r, \phi[w])$ from local neighbours, where messages along outgoing edges and incoming edges have different forms.*

- *An aggregation function $q_\mathrm{A}$ summing all messages from neighbourhood producing $z[v]$*

- *A correction term $n[v]$ from the global partition of outgoing edges and the non-local edges.*

- *An update rule:*

$$\phi[v] \leftarrow q_\mathrm{U}(\phi[v], z[v]) = \phi[v] + \alpha z[v] - \beta n[v],$$

*with step sizes $\alpha, \beta$.*

*This establishes equivalence between DistMult's gradient update and a message-passing architecture with global context.*

What emerges from the equations is that each GD step contains an explicit information flow from the neighbourhood of each node, which is then aggregated with a simple summation. Through this direct information path, $t$ GD steps cover the $t$-hop neighbourhood of $v$. As $t$ goes towards infinity or in practice as training converges, FMs capture the global graph structure. The update function (Equation 4.13) somewhat deviates from classic message passing frameworks as $n[v]$ of Equation 4.14 involves global information. However, we note that we can interpret this mechanism under the framework of augmented message passing [Veličković, 2022] and, in particular, as an instance of *graph rewiring*, where $n[v]$ represents rewired edges to global nodes that are not in the local neighbourhood.

Based on Theorem 4.4.1 and Equation 4.7, we can now view $\phi$ as the transient node states $h$ (cf. Section 4.3.2) and GD on node embeddings as a message-passing layer. This dualism sits at the core of the ReFactor GNN model, which we describe next.

## 4.5 REFACTOR GNN: Inductivising Factorization based Models

FMs are trained by minimising the objective (Equation 4.3), initialising both sets of parameters ($\phi$ and $\psi$) and performing GD until approximate convergence (or until early stopping terminates the training). The implications are twofold: $i$) the initial value of the entity lookup table $\phi$ does not play any major role in the final model after convergence, and $ii$) if we introduce a new set of entities, the conventional wisdom is to retrain[2] the model on the expanded knowledge graph. This can be computationally rather expensive and operationally complex, compared to the "inductive" models that require no additional training and can leverage node features like entity descriptions.

However, as we have just seen in Theorem 4.4.1, the training procedure of FMs may be naturally recast as a message-passing operation, which suggests that it is possible to use FMs for inductive learning tasks. In fact, we envision that there is an entire novel spectrum of model architectures interpolating between pure FMs and (various instantiations of) GNNs. Here we propose one simple implementation of such an architecture which we dub REFACTOR GNNs. Figure 4.2 gives an overview of REFACTOR GNNs.

**The REFACTOR Layer**   A REFACTOR GNN contains $L$ REFACTOR layers, that we derive from Theorem 4.4.1. Aligning with the notations in Section 4.3.2, given a knowledge graph $\mathcal{T}$ and entity representations $h^{l-1} \in \mathbb{R}^{|\mathcal{E}| \times K}$, the REFACTOR layer computes the

---

[2]Typically, until convergence and possibly by partially warm-starting the model parameters.

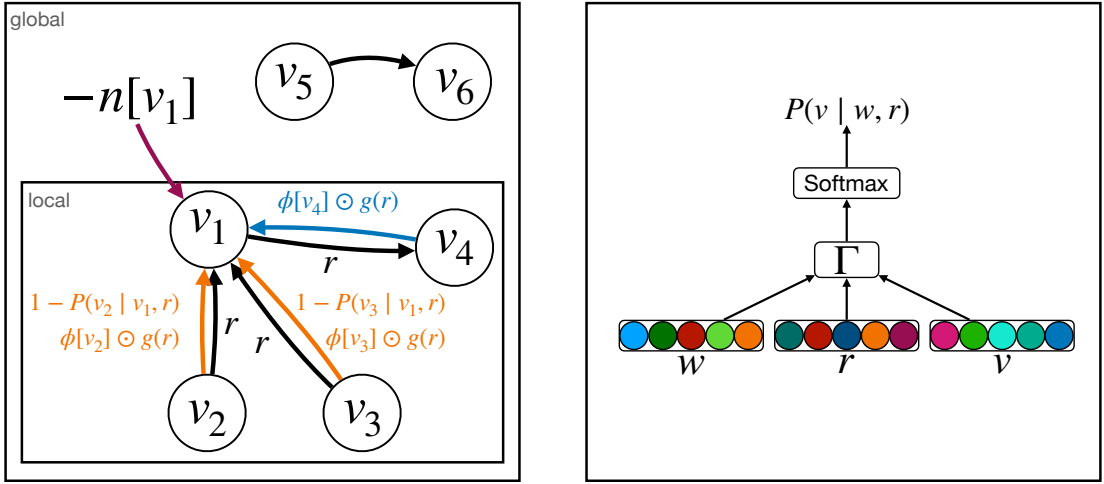

Figure 4.2: ReFactor GNN architecture. The left figure describes messages from the local neighbourhood $\{(v_2, r_1, v_1), (v_3, r_2, v_1), (v_1, r_3, v_4)\}$ (the orange and blue edges, which depend on the type of relationship of the edges) and a global normaliser term induced by the partition function (the purple arrow); The right figure describes the computation graph for calculating $P(v \mid w, r)$, where $v, w \in \mathcal{E}$ and $r \in \mathcal{R}$: the embedding representations of $w$, $r$, and $v$ are used to score the edge $(w, r, v)$ via the scoring function $\Gamma$, which is then normalised via the SoftMax function.

representation of a node $v$ as follows:

$$h^l[v] = q^l(\mathcal{T}, h^{l-1})[v] = h^{l-1}[v] - \beta n^l[v] + \alpha \sum_{(r,w) \in \mathcal{N}^1[v]} q_M^l(h^{l-1}[v], r, h^{l-1}[w]), \quad (4.15)$$

where the terms $n^l$ and $q_M^l$ are derived from Equation 4.14 and Equation 4.11, respectively. We note that REFACTOR GNNs treat incoming and outgoing neighbourhoods differently instead of treating them equally as in for example the R-GCN, the first GNN on multi-relational graphs [Schlichtkrull et al., 2018].

Equation 4.15 describes the full batch setting, which can be expensive if the KG contains many edges. Therefore, in practice, whenever the graph is big, we adopt a stochastic evaluation of the REFACTOR layer by decomposing the evaluation into several mini-batches. We partition $\mathcal{T}$ into a set of computationally tractable mini-batches. **For each mini-batch, we restrict the neighbourhoods to the subgraph induced by it and readjust the computation of $n^l[v]$ to include only entities and edges present in it**.

We leave the investigation of other stochastic strategies (e.g. by taking Monte Carlo estimations of the expectations in Equation 4.14) to future work. Finally, we cascade the mini-batch evaluation to produce one full layer evaluation (i.e. one message-passing round over the entire graph).

**Training** The learnable parameters of ReFactor GNNs are the relation embeddings $\psi$, which parameterise the $g(r)$ in the message function $q_M^l, l \in [1, L]$. Inspired by Fey et al. [2021], You et al. [2020], we learn $\psi$ by layer-wise (stochastic) gradient descent. This is in contrast to conventional GNN training, where one needs to backpropagate through all the message-passing layers $l \in [1, L]$. A (full-batch) GD training dynamic for $\psi$ can be written as

$$\psi_{t+1} = \psi_t - \eta \nabla \mathcal{L}_t(\psi_t)$$

with:

$$\mathcal{L}_t(\psi_t) = \sum_{\mathcal{T}} -\frac{1}{|\mathcal{T}|} \log P_{\psi_t}(w|v, r)$$

where $P_{\psi_t}(w|v,r) = \text{Softmax}(\Gamma(v, r, \cdot))[w] \quad \Gamma(v, r, w) = \langle h^t[v], h^t[w], g_{\psi_t}(r) \rangle$.

$h^t[\cdot]$ denotes the node state of a particular node at iteration $t$ and the node state is updated recursively as

$$\begin{aligned} h^0 &= X, \text{initial node features} \\ h^t &= q^l(\mathcal{T}, h^{t-1}) \text{where } l = t \bmod L, t \geq 1. \end{aligned} \quad (4.16)$$

This dynamic ensures that at each step $t$, only the current layer $l = t \bmod L \in [1, L]$ is activated and participates in the backpropagation. Early layers $< l$ are truncated from the computational graph by treating $h^{t-1}$ as a fixed (non-differentiable) input for the current layer, bounding the gradient path to a single layer per training step. Such truncation of the computational graph to reduce memory usage is not uncommon and have been used in meta-learning algorithms [Chen et al., 2019] and for GNN scaling techniques [Fey et al., 2021].

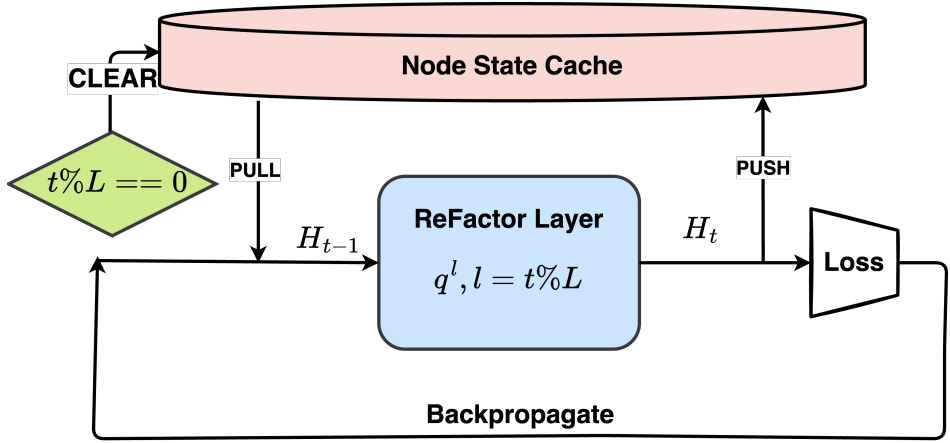

Figure 4.3: Illustration of the external node state cache used during training.

**External cache, its push, pull, and clear.** Implementation-wise, such a training dynamic equals to maintaining an external **memory** for storing and retriving historical node states $h^{t-1}$ to compute $h^t$ using Equation 4.15. Figure 4.3 illustrates the external cache. During the model optimisation, the historical node states are fixed. After each training step, newly computed node states are pushed to update the historical cache. But this push occurs after gradient computation, and these historical vectors are not part of the current backpropagation path. After every $L$ full batches, we clear the cache by resetting all node states in the cache to their initial input values $X$ (e.g., textual or random features). This procedure of push, pull, and clear, emulates an unrolling of the message-passing dynamic up to $L$ layers, and forces the model to predict based on on-the-fly $L$-layer message-passing. After training, we obtain $\psi^*$ and perform inference by running $L$-layer message-passing with $\psi^*$. In general, $L$ determines the number of effective message-passing layers in REFACTOR GNNs. A larger $L$ enables REFACTOR GNNs to fuse information from more hops of direct neighbourhoods into the final node representations. In the meantime, it reduces the inductive applicability of REFACTOR GNNs due to over-smoothing and computational requirements. In the extreme case of $L = \infty$, *where we never clear the node state cache during training*, the final cached node states will be used for inference. Note that this latter inference regime is inherently transductive since there will be no cached states for new nodes. Future work may explore a more streamlined implementation by simply resetting the entity embeddings periodically as in Chen et al. [2023].

**Relation to prior work**  While our use of caching is inspired by AutoScale [Fey et al., 2021], our model diverges in key ways. Unlike Fey et al. [2021], where the historical node states are only used for out-of-batch neighbour nodes, we use historical node states for all nodes. Fey et al. [2021] define only the "push" and "pull" operations for the memory. We additionally define a "clear" operation for the memory. This cache-clearing mechanism acts as a form of *active forgetting*, which we introduce to promote inductive capability. Work in the spirit of active forgetting has been extensively explored in the continual learning literature as a mechanism for improving adaptability and reduce overfitting to past learnings. For instance, neural pruning removes low-activity neurons to free capacity for future tasks [Golkar et al.]; episodic backward updates selectively discard outdated gradients to favor recent learning [Lee et al., 2019]; and meta-experience replay strategies [Riemer et al.] reduce gradient interference, effectively suppressing conflicting knowledge. Our cache reset parallels these approaches by clearing outdated node embeddings, thereby preventing over-specialization and supporting generalization to unseen entities. These modifications are essential in adapting static factorisation models into a dynamic, message-passing framework suitable for both transductive and inductive link prediction tasks.

## 4.6 Experiments

We perform experiments to answer the following questions regarding ReFactor GNNs:

- **RQ1.** ReFactor GNNs are derived from a message-passing reformulation of FMs: do they also inherit FMs' predictive accuracy in *transductive* KGC tasks? (Section 4.6.1)

- **RQ2.** ReFactor GNNs "inductivise" FMs. Are they more statistically accurate than other GNN baselines in *inductive* KGC tasks? (Section 4.6.2)

- **RQ3.** The term $n[v]$ involves nodes that are not in the 1-hop neighbourhood. Is such *augmented message passing* [Veličković, 2022] necessary for good KGC performance? (Section 4.6.3)

For transductive experiments, we used three well-established KGC datasets: *UMLS*, *CoDEx-S*, and *FB15K237* [Kemp et al., 2006, Safavi and Koutra, 2020, Toutanova and Chen, 2015]. For inductive experiments, we used the inductive KGC benchmarks introduced by GraIL [Teru et al., 2020], which include 12 pairs of knowledge graphs:

- (*FB15K237_vi*, *FB15K237_vi_ind*),

- (*WN18RR_vi*, *WN18RR_vi_ind*),

- (*NELL_vi*, *NELL_vi_ind*),

where $i \in [1, 2, 3, 4]$, and (*_vi*, *_vi_ind*) represents a pair of graphs with **a shared relation vocabulary and non-overlapping entities**. Note that the GraIL setup is different from a completely inductive setup, where both the relations and entities are unseen at test time.

We follow the standard KGC evaluation protocol by fully ranking all the candidate entities and computing two metrics using the ranks of the ground-truth entities: Mean Reciprocal Ranking (MRR) and Hit Ratios at Top K (Hits@$K$) with $K \in [1, 3, 10]$. For the inductive KGC, we additionally consider the partial-ranking evaluation protocol used by GraIL for a fair comparison. Empirically, we find full ranking more difficult than partial ranking, and thus more suitable for reflecting the differences among models on GraIL datasets. In fact, we would like to call for future work on GraIL datasets to also adopt a full ranking protocol on these datasets.

**Our *transductive* experiments used $L = \infty$, i.e. node states cache is never cleared, as we wanted to see if ReFactor GNNs ($L = \infty$) can reach the performance of the FMs (Section 4.6.1); on the other hand, in our *inductive* experiments, we used ReFactor GNNs with $L \in \{1, 2, 3, 6, 9\}$, since we wanted to test their performances in inductive settings akin to standard GNNs (Section 4.6.2).** We used a hidden size of 768 for the node representations. All the models are trained using $[128, 512]$ in-batch negative samples and one global negative node for each positive link. We performed a grid search over the other hyperparameters and selected the best configuration based on the validation MRR. Since training deep GNNs with full-graph message passing might be slow for large knowledge graphs, we follow the literature [Hamilton et al., 2017, Zou et al., 2019, Zeng et al., 2020] to sample sub-graphs for training GNNs as indicated by Equation 4.6. Considering that sampling on the fly often prevents high utilisation of GPUs, we resort to a two-stage process: we first sampled and serialised sub-graphs around the target edges in the mini-batches; we then trained the GNNs with the serialised sub-graphs. To ensure that we have sufficient sub-graphs for training the models, we sampled for 20 epochs for each knowledge graph, i.e. 20 full passes over the full graph. The sub-graph sampler we currently used is LADIES [Zou et al., 2019].

Table 4.1: Test MRR for transductive KGC tasks.

| Entity Encoder | UMLS | CoDEx-S | FB15K237 |
| --- | --- | --- | --- |
| R-GCN | – | 0.33 | 0.25 |
| Lookup (FM, specif. DistMult) | 0.90 | 0.43 | 0.30 |
| ReFactor GNNs ($L = \infty$) | 0.93 | 0.44 | 0.33 |

### 4.6.1 RQ1: ReFactor GNNs for Transductive Learning

ReFactor GNNs are derived from the message-passing reformulation of FMs. We expect them to approximate the performance of FMs for transductive KGC tasks. To verify this, we perform experiments on the datasets UMLS, CoDEx-S, and FB15K237. For a fair comparison, we use Equation 4.2 as the decoder and consider i) lookup embedding table as the entity encoder, which forms the FM when combined with the decoder (Section 4.3.1), and ii) ReFactor GNNs as the entity encoder. Note that the equivalence between ReFactor GNNs and the standard FMs are only obtained when ReFactor GNNs are trained with $L = \infty$, i.e. we never clear the node state cache. This is different from inductive setups, where ReFactor GNNs are trained with a finite $L$. Since transductive KGC tasks do not involve new entities, the node state cache in ReFactor GNNs can be directly used for link prediction. Table 4.1 summarises the result. We observe that ReFactor GNNs achieve a similar or slightly better performance compared to the FM. This shows that ReFactor GNNs are able to capture the essence of FMs and thus remain competitive at transductive KGC.

### 4.6.2 RQ2: ReFactor GNNs for Inductive Learning

Despite FMs' good empirical performance on transductive KGC tasks, they fail to be as inductive as GNNs. According to our reformulation, this is due to the infinite message-passing layers hidden in FMs' optimisation. Discarding infinite message-passing layers, REFACTOR GNNs enable FMs to perform inductive reasoning tasks by learning to use a finite set of message-passing layers for prediction similarly to GNNs.

Here we present experiments to verify REFACTOR GNNs's capability for inductive reasoning. Specifically, we study the task of inductive KGC and investigate whether REFACTOR GNNs can generalise to unseen entities. Following Teru et al. [2020], on GraIL datasets, we trained models on the original graph, and run 0-shot link prediction on the _ind test graph. Similar to the transductive experiments, we use Equation 4.2 as the decoder and vary the entity encoder. We denote three-layer REFACTOR GNNs as ReFactor(3) and six-layer REFACTOR GNNs as ReFactor(6). We consider several baseline entity encoders: i) no-pretrain, models without any pretraining on the original graph; ii) GAT(3), three-layer graph attention network [Veličković et al., 2018]; iii) *GAT(6)*, six-layer graph attention network; iv) GraIL, a sub-graph-based relational GNN [Teru et al., 2020]; v) NBFNet, a path-based GNN [Zhu et al., 2021], current SoTA on GraIL datasets. In addition to randomly initialised vectors as the node features, we also used textual node features, RoBERTa [Liu et al., 2019a] Encodings of the entity descriptions, which are produced by SentenceBERT [Reimers and Gurevych, 2019]. Due to space reason, we present the results on (*FB15K237_v*1, *FB15K237_v*1*_ind*) in Figure 4.4. Results on other datasets are similar and can be found in the appendix. We can see that without textual node features, REFACTOR GNNs perform better than GraIL (+23%); with textual node features, REFACTOR GNNs outperform both GraIL (+43%) and NBFNet (+10%), achieving new SoTA results on inductive KGC.

**Performance vs Parameter Efficiency as #Message-Passing Layers Increases** Usually, as the number of message-passing layers increases in GNNs, the over-smoothing issue occurs while the computational cost also increases exponentially. REFACTOR GNNs avoid this by layer-wise training and sharing the weights across layers. Here we compare REFACTOR GNNs with $\{1, 3, 6, 9\}$ message-passing layer(s) with same-depth GATs. Results are summarised in Figure 4.5. We observe that increasing the number of message-passing layers in GATs does not necessarily improve the predictive accuracy – the best results were obtained with 3 message-passing layers on *FB15K237_v1* while using 6 and 9 layers leads to performance degradation. On the other hand, REFACTOR GNNs obtain consistent improvements when increasing #Layers from 1 to 3, 6, and 9. REFACTOR

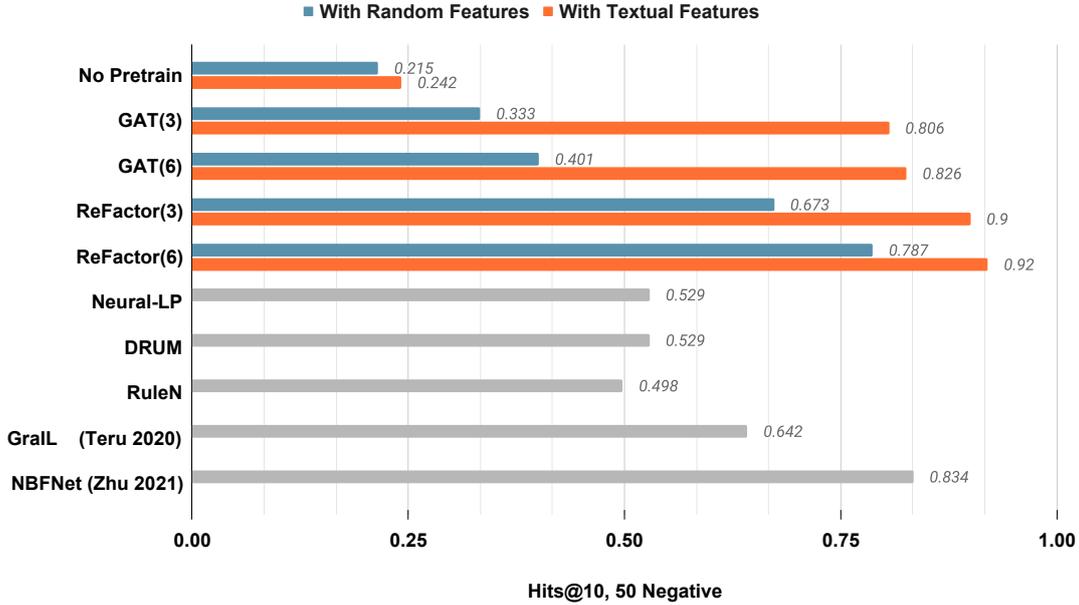

Figure 4.4: Inductive KGC performance, trained on the KG *FB15K237_v1* and tested on another KG *FB15K237_v1_ind*, where the entities are completely new. The results of GraIL and NBFNet are taken from Zhu et al. [2021]. The grey bars indicate methods that are not devised to incorporate node features.

GNNs $(6, 6)$ and $(9, 9)$ clearly outperform their GAT counterparts. Most importantly, REFACTOR GNNs are more parameter-efficient than GATs, with a constant #Parameters as #Layers increases.

### 4.6.3 RQ3: Beyond Message-Passing

As shown by Theorem 4.4.1, REFACTOR GNNs contain not only terms capturing information flow from the 1-hop neighbourhood, which falls into the classic message-passing framework, but also a term $n[v]$ that involve nodes outside the 1-hop neighbourhood. The term $n[v]$ can be treated as *augmented message-passing* on a dynamically rewired graph [Veličković, 2022]. Here we perform ablation experiments to measure the impact of the $n[v]$ term. Table 4.2 summarises the ablation results: we can see that, without the term $n[v]$, REFACTOR GNNs with random vectors as node features yield a 2% lower MRR, while REFACTOR GNNs with RoBERTa textual encodings as node features produce a 7% lower MRR. This suggests that augmented message-passing also plays a significant role in REFACTOR GNNs' generalisation properties in downstream link prediction tasks. Future work might gain more insights by further dissecting the $n[v]$ term.

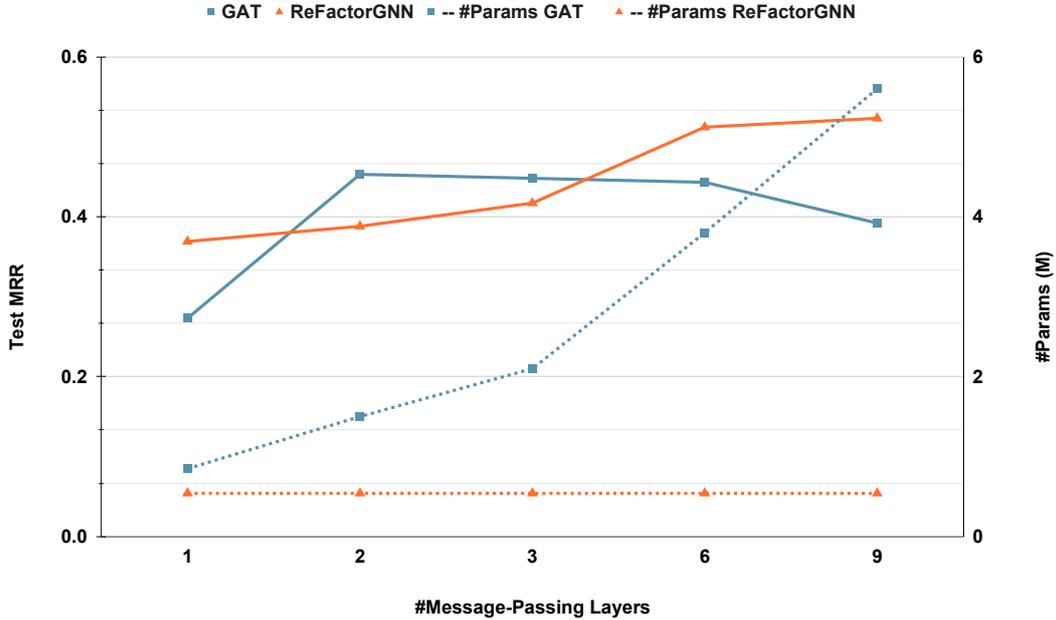

Figure 4.5: Performance vs parameter efficiency on *FB15K237_v1*. Left axis is Test MRR while right axis is #parameters. The solid lines and dashed lines indicate the changes of Test MRR and the changes of #parameters.

Table 4.2: Ablation on $n[v]$ for REFACTOR GNNs (6) trained on *FB15K237_v1*.

| Test MRR | With Random Features | With Textual Features |
|---|---|---|
| with $n[v]$ | 0.425 | 0.486 |
| without $n[v]$ | 0.418 | 0.452 |

## 4.7 Discussion

**Summary.** The task of multi-relational link prediction forms the cornerstone of constructing useful knowledge graphs, which, in turn, underpin modern knowledge engines. Factorization Models (FMs) and Graph Neural Networks (GNNs) are two prominent ap-proaches for this task. FMs excel in transductive settings, while GNNs are better suited for inductive scenarios. Despite the sharp differences in their analytical forms, our work establishes a link between FMs and GNNs. By reformulating FMs as GNNs, we address a critical question: why are FMs superior transductive multi-relational link predictors but fail in inductive scenarios? The answer lies in FMs performing excessive message-passing in standard training, capturing excessive global structures, and producing overly rigid representations.

Building on this insight, we propose ReFactor GNNs, a novel GNN variant that incorporates an *active forgetting* mechanism into the message-passing process of FMs. ReFactor GNNs periodically reset the cache of prior message-passing computations, enabling the model to focus on local neighbourhood information instead of over-relying on the entire training graph. Empirical experiments demonstrate that ReFactor GNNs achieve significantly higher accuracy than GNN baselines on inductive link prediction tasks, bridging the gap between the strengths of FMs and GNNs.

**Limitations.** Since we adopted a two-stage (sub-graph serialisation and then model training) approach instead of online sampling, there can be side effects from the low sub-graph diversity. In our experiments, we used LADIES [Zou et al., 2019] for sub-graph sampling. Experiments with different sub-graph sampling algorithms, such as GraphSaint [Zeng et al., 2020] might affect the downstream link prediction results. Furthermore, it would be interesting to analyse decoders other than DistMult, as well as additional optimisation schemes beyond SGD and AdaGrad. We do not dive deeper into the expressiveness of ReFactor GNNs. Nevertheless, we offer a brief discussion in Section B.1.1.

**Implications.** The most direct future work would be using the insight to develop more sophisticated models at the intersection between FMs and GNNs, e.g. by further parameterising the message/update function. One implication from our work is that reformulating FMs as message-passing enables the idea of "learning to factorize". This might broaden the usage of FMs, going beyond link prediction, to tasks such as graph classification. Another implication comes from our approach of unpacking embedding updates into a series of message-passing operations. This approach can be generalised to other dot-product-based models that use embedding layers for processing the inputs, lend-ing it naturally to understanding complicated attention-based models like Transformers. Although Transformers can be treated as GNNs over fully-connected graphs, where a sentence would be a graph and its tokens would be the nodes, the message-passing is limited to within each sentence under this view. We instead envision cross-sentence message-passing by reformulating the updates of the token embedding layer in transformers. In general, the direction of organising FMs, GNNs, and transformers under the same framework will allow a better understanding of all three models. While FMs and GNNs excel in the structured paradigm, transformers are often the default choice for the unstructured paradigm. Unveiling the connections among these models can facilitate the seamless integration of the structured and unstructured paradigm, paving the way for building universal knowledge engines.

## Chapter 5

# Improving Language Plasticity via Pretraining with Active Forgetting

*A version of this work was previously presented at a peer-reviewed conference. Please refer to [Chen et al., 2023] for full citation.*

Reality is full of constantly changing details. To navigate such dynamism, intelligent agents must adapt to new information in *real time*. This requires mechanisms that support flexible knowledge integration. *Active forgetting* (Chapter 4) appears to be one such mechanism: by actively forgetting historical node states resulted from previous message-passing computation, factorization-based models – representatives of the structured paradigm – can learn to accommodate new entity nodes in knowledge graphs, weaving them into the fabric of existing knowledge. At its core, active forgetting manifests an emergent principle of *destructuring*:

> To remain adaptable in changing environments, intelligent units (e.g., agents, models, humans) must not only construct knowledge, but also deliberately dismantle parts of it.

Increasingly, similar manifestations of such intentional destructuring have been identified across domains including but not limited to psychology, neuroscience, education, and artificial intelligence [Levy et al., 2007, Barrett and Zollman, 2009, Hardt et al., 2010, 2013, Anderson and Hulbert, 2021, Nikishin et al., 2022, Zhou et al., 2022,

Ramkumar et al., 2023], reinforcing the idea that intelligence, especially its fluid side [Cattell, 1963, Horn and Cattell, 1966, Brown, 2016, Kent, 2017], relies as much on destructuring as on structuring. Structuring provides the foundations for consistent reasoning and repeatable knowledge serving. Destructuring, on the other hand, overcomes outdated and overly-rigid structures.

One of the key challenges in materializing the destructuring principle is to find the targets to dismantle. For natural intelligence, the targets of destructuring can be both cognitive and psychic structures. For instance, dismantling entrenched associative thinking patterns can lead to novelty in idea generation [Horan, 2009], while breaking down rigid psychic structures increases mental mobility, turning behavioural rigidity into feeling, thinking, and action [Sandell, 2019]. Similarly, inhibition of linguistic structures from one's native language plays an important role in acquiring a second language [Levy et al., 2007, MacWhinney, 2005, Schmid, 2017].

For artificial intelligence, the targets of destructuring remain understudied. Partially because scaling model sizes is the focus right now as it is more prominent in improving benchmark numbers. However, as more and more inappropriate behaviours by these models are exposed [Farquhar et al., 2024, Shumailov et al., 2024], it becomes more and more important to underpin these inappropriate structures inside the models. Chapter 2 and 3 show that certain structures are stored in the embeddings and their interactions with other layers in both the structured and unstructured learning paradigms. This perspective offers tangible structural underpinnings to the embedding layer. Chapter 4 further explains the role of embedding and chose them as the targets for destructuring, with evidences showing this helps models accommodate new entities in the knowledge graphs. While the findings from Chapter 4 are limited to the structured learning paradigm, an important question arises: can similar destructuring techniques benefit models operating in the unstructured paradigm. Specifically, we ask *can pretrained language models, the predominant tools for constructing knowledge engines from unstructured data sources, benefit from destructuring techniques?*

## 5.1 Towards Language Model Plasticity

Pretrained language models (PLMs) have been swiftly reshaping the landscape of natural language processing (NLP) by improving upon standardized benchmarks across the board [Radford and Narasimhan, 2018, Devlin et al., 2019, Liu et al., 2019b, Brown et al., 2020]. They are often regarded as the Swiss Army knife of the unstructured paradigm for building general knowledge engines. At their core, they acquire knowledge by ingesting large datasets and store this knowledge in their parameters during pretraining. Using finetuning or prompting [Brown et al., 2020], such knowledge can then be applied to downstream applications, such as semantic analysis, question answering, writing assistance, coding companion, and many others.

Despite their success, PLMs still have a number of shortcomings [Weidinger et al., 2021, 2022]. In particular, it requires massive data and computation to pretrain them [Gururangan et al., 2020, Kaplan et al., 2020, Hernandez et al., 2021, Hu et al., 2021, Touvron et al., 2023b]. Naively retraining a new PLM to accommodate every lingual space shift[1] would be prohibitively expensive. This makes it a highly relevant research target to create PLMs that can be efficiently adapted to new lingual spaces.

While forgetting in the context of both human and machine learning is often perceived as something negative (for instance in the case of catastrophic forgetting where learning new tasks overwrites the old knowledge [McCloskey and Cohen, 1989, Ratcliff, 1990, Kirkpatrick et al., 2017]), recent works have shown that for artificial neural networks, forgetting can also play a *positive* role in increasing their "plasticity", such as improving generalization to unseen data [Zhou et al., 2022, Chen et al., 2022, Igl et al., 2021], enabling learning in low-data regimes [Alabdulmohsin et al., 2021, Taha et al., 2021], or counteracting primacy bias [Nikishin et al., 2022, D'Oro et al., 2023]. Although these pioneering works in continual learning do not explicitly define model plasticity, they in essence share a common goal across different tasks and models: improving a model's ability to remain stable while adapting flexibly to drastically changing inputs, addressing the *stability-plasticity dilemma*. Given these developments and their insights, in this work, we explore if we can draw upon forgetting techniques as a mechanism to improve *pretraining* and imbue PLMs with similar benefits in model plasticity.

It is well established in the NLP community that models struggle to generalize across languages without substantial intervention [Conneau et al., 2020, Pfeiffer et al., 2020, 2022, Ansell et al., 2022], which is especially true for low-resources languages. We thus see this as a promising testing ground for forgetting techniques. Our focus is on the input layer of the PLM, the *token embedding layer*, as learning it has been shown to be highly effective when adapting between languages [Artetxe et al., 2020].

---

[1] We use the term *lingual space shift* to describe changes in language usage between pretraining and the target downstream application, caused by factors such as language change, time evolution, or domain variation. A model with high *language plasticity* would quickly adapt to these shifts.

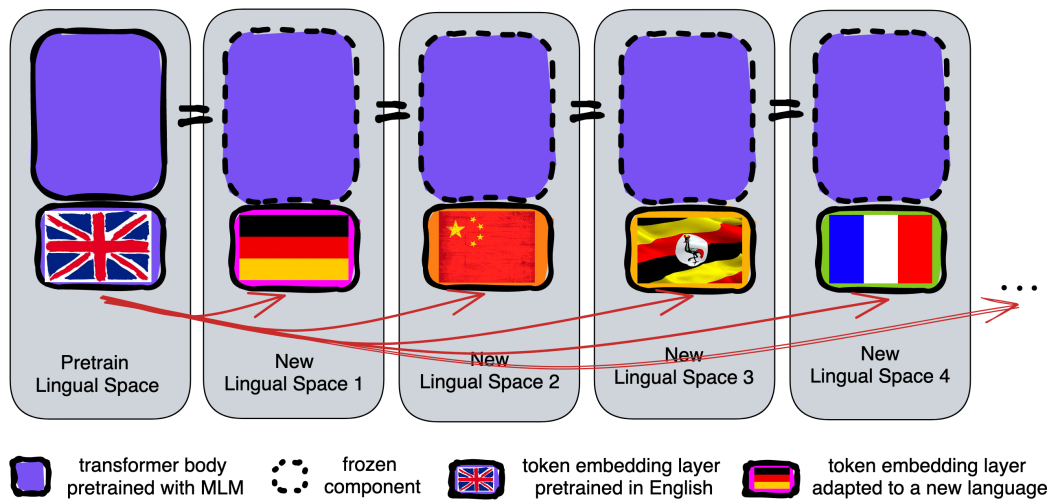

Figure 5.1: *Rewiring* via relearning token embeddings: where the transformer body (the purple part) is "frozen" and reused for a new language, but the token embeddings are relearned to suit the new language.

Concretely, we propose to introduce *active forgetting* mechanism into the pretraining phase, which resets token embeddings at regular intervals, while leaving all other parameters untouched throughout pretraining. We study whether this forgetting approach creates a PLM that can easily *rewire* (Figure 5.1) to an unseen (possibly distant) language. Intuitively, resetting embeddings forces the transformer body to re-derive reasoning each time instead of relying on memorized shortcuts. Through repetition, the body learns more abstract, high-level reasoning. A model with greater abstraction can easily transfer across languages, since high-level reasoning is more language-agnostic.

Our zero-shot evaluations on several cross-lingual transfer benchmarks show that for cases where unlabeled adaptation corpus for the unseen language has as few as 5 million tokens (a low-data regime), forgetting PLMs outperforms the baseline by large margins: average gains of $+21.2\%$ on XNLI, $+33.8\%$ on MLQA, and $+60.9\%$ on XQuAD. In addition, models pretrained using active forgetting converge faster during language adaptation. Finally, we find that active forgetting is especially beneficial for languages that *are* *distant from* English, such as Arabic, Hindi, Thai, and Turkish. Implementation-wise, the method does not introduce significant overhead to the already complex pretraining process, making it a cost-efficient way to promote a meta-learning-like effect. For those interested in details, the code is available at `https://github.com/facebookresearch/language-model-plasticity`.

## 5.2 Literature Review: Forgetting, its Positive Roles, and Cross-lingual Transfer

### 5.2.1 Forgetting and its Positive Role

The common perception of forgetting is that it implies weak memory and a loss of acquired knowledge, thus it is often regarded as a sign of *un-intelligence* or an undesirable property. In neural networks, *catastrophic forgetting* [McCloskey and Cohen, 1989, Ratcliff, 1990, Kirkpatrick et al., 2017] is portrayed as a forgetting phenomenon where neural networks lose the ability to predict old patterns after new inputs alter their weights. Forgetting, in this context, has negative consequences, as the new knowledge overwrites the prior valuable knowledge. Plenty of prior research strives to overcome catastrophic forgetting and enable continual learning [Schmidhuber, 2013, Kirkpatrick et al., 2017, Lopez-Paz and Ranzato, 2017, Shin et al., 2017, Schwarz et al., 2018, Mallya and Lazebnik, 2018, Parisi et al., 2019, Rolnick et al., 2019, Beaulieu et al., 2020, Veniat et al., 2020, Gaya et al., 2023, Khetarpal et al., 2022].

Our work differs from the above ones in that our subject is *intentional forgetting* rather than passive forgetting and its associated negative impact. To put it in another way, we seek to understand how forgetting – if purposely incorporated as an active process during training – might *help* new learning. Similar positive roles of forgetting have been discussed in the literature. Specifically, Pastötter et al. [2008] demonstrate forgetting enhances the learning of new information by resetting the encoding process and holding the attention at high levels; Levy et al. [2007] show that it helps second language acquisition by inhibiting the native language; Barrett and Zollman [2009] find it promote the emergence of an optimal language by preventing partial success from reinforce suboptimal practice. Nørby [2015] further suggests forgetting serves adaptive functions, helping people regulate emotions, acquiring knowledge and staying attuned to the context. More recently Anderson and Hulbert [2021] reviews evidence on active forgetting by prefrontal control and shows how it can adapt the memory to suit either emotional or cognitive goals.

### 5.2.2 Forgetting via Partial Neural Weights Reset

In neural networks, forgetting can be instantiated in many forms. A simple way is to reset subsets of parameters before the next round of learning. Iterations of such resetting have been shown to benefit generalization with low compute and low data for computer vision tasks [Frankle and Carbin, 2019, Alabdulmohsin et al., 2021, Taha et al., 2021, Ramkumar et al., 2023]. More recently, Zhou et al. [2022] demonstrate a similar forgetting strategy helps image classification and language emergence. Closely linked to the method in this chapter, Chapter 4 forget node embeddings in order to truncate infinite message-passing among nodes and thereby aid new graph reasoning with new nodes. Our work uses similar forgetting mechanism over token embeddings, improving new language reasoning with new tokens. As far as we know, *we are the first to bring forgetting into pretraining and demonstrate that forgetting pretraining boosts linguistic plasticity*. A relevant thread in reinforcement learning (RL) research studies the plasticity loss phenomenon [Lyle et al., 2023, Nikishin et al., 2023]. Recent work explores similar forgetting approaches to improve plasticity. Igl et al. [2021] periodically reset the current policy by distilling it into a reinitialised network throughout training. Intuitively, this releases network capacity storing suboptimal policies and opens up space for the yet-to-be-discovered optimal (final) policy. Simpler methods just reset an agent's last layers [Nikishin et al., 2022], preventing overfitting to early experiences and *primacy bias*. Resetting parameters also improves sample efficiency by allowing more updates per environment interaction [D'Oro et al., 2023].

### 5.2.3 Cross-lingual Transfer for Pretrained Language Models

Pretraining on multilingual data makes PLMs multilingual [Conneau et al., 2020] but has downsides like needing large multilingual corpus with appropriate mixing, potential interference among languages, and difficulty of covering all languages. Alternatively, the line of research on cross-lingual transfer makes PLMs multilingual by extending

English-only PLMs to other languages. Artetxe et al. [2020] demonstrate possibility of such extension by relearning the embedding layer with unsupervised data from the new language. Marchisio et al. [2023] further increase computational efficiency using a mini-model proxy. Liu et al. [2023a] use a similar partial reset-reinit approach in vision-language settings. Approaches based on adapters and sparse finetuning have also been proposed [Pfeiffer et al., 2020, 2022, 2021, Ansell et al., 2022]. Adapters are bottleneck layers (usually placed after the feedforward layers) that add extra capacity when adapting to a different task or language. Our proposed forgetting mechanism can be applied to adapter-based methods as we can allow forgetting to happen in the adapter layers. The current choice of forgetting embeddings keeps the architecture intact and incurs no additional hyperparameter tuning, allowing us to understand the fundamental capability of forgetting pretraining.

## 5.3 Rewiring PLMs for New Languages

Using unlabeled data, Artetxe et al. [2020] demonstrates possibility of rewiring a monolingual PLM to a new language; they propose to relearn the embedding layer for the new language while keeping all the other parameters frozen. The underlying assumption is that the token embedding layer and the transformer body (the non-token-embedding parameters) divide up the responsibility in a way that the former handles language-specific lexical meanings, while the latter deals with high-level general reasoning. Hence, rewiring an English PLM for a new language boils down to separately adapting the former with unlabelled data in the new language and the latter with English task data. The procedure can be summarized as follows:

1. Pretrain: A transformer-based model is pretrained on an *English* corpus. In our experiments, we choose to pretrain RoBERTa-base Liu et al. [2019b], a 12-layer transformer-based model, on English CC100 [Conneau et al., 2020].

2. Language Adapt: The token embedding layer is finetuned using unlabelled data in the new language, while the transformer body is frozen.

3. Task Adapt: The transformer body is finetuned using downstream task data in English, while the token embedding layer is frozen.

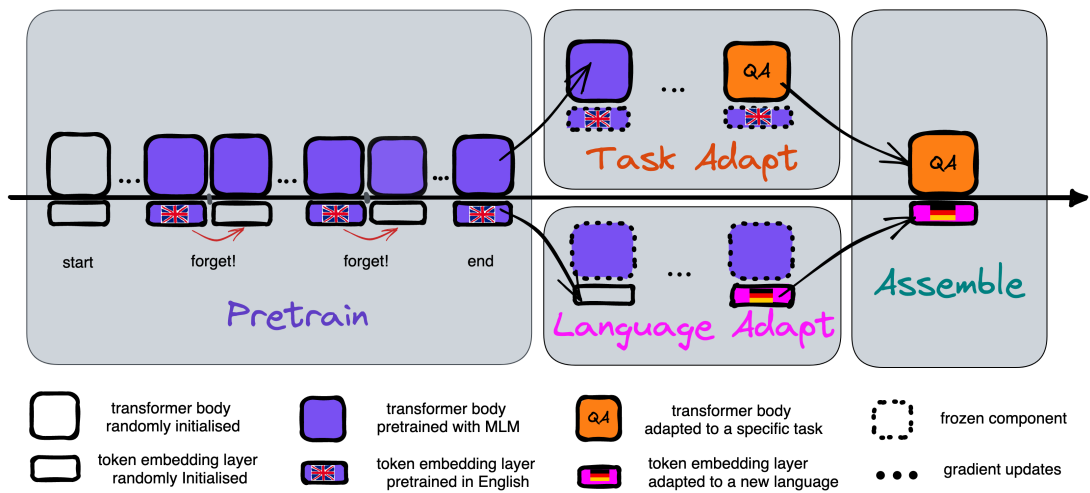

Figure 5.2: Unsupervised zero-shot cross-lingual transfer. **Left**: in the pretrain stage, we compare standard pretraining with forgetting pretraining, where the token embeddings are actively forgotten at a regular interval while the transformer body is learned as the standard pretraining. **Middle**: the task adapt stage and the language adapt stage separately adapt the transformer body using English task data and the token embeddings using unlabelled data in the new language. **Right**: the assemble stage reassemble the adapted body and token embedding layer into a usable PLM.

4. Assemble: The final model is assembled by taking the adapted token embedding layer from stage 2 and the adapted transformer body from stage 3.

## On The Difficulty of Rewiring PLMs via Relearning Token Embeddings

While the above procedure [Artetxe et al., 2020] offers a general framework for rewiring a monolingual PLM with unlabelled data in the new language, it is unclear how efficient such rewiring can be, including both sample efficiency and computational efficiency. To better understand the difficulty of rewiring PLMs via relearning the token embeddings, we design an experiment where we relearn the token embedding layer using varying amounts of adaptation data. For illustration purpose, we pick English as the pseudo "adaptation language" as its dataset is large enough to bootstrap a series of sub-datasets with varying quantity.

We create subsets with $[1K, 10K, 100K, 1M, 5M, 10M, 100M, 1B, 10B]$ tokens and relearn the English embeddings while keeping the transformer body frozen.

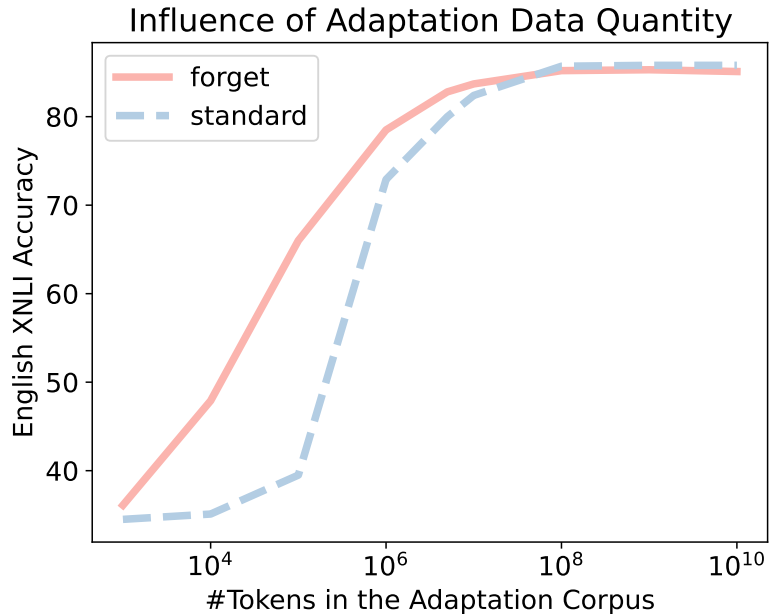

Figure 5.3: The rewiring performance for standard PLMs (blue dashed line) drops drastically if the adaptation tokens $\leq$ 10M.

The dashed blue line in Figure 5.3 summarizes the influence of the adaptation data quantity on the quality of the rewired PLMs (relearned embeddings assembled with the English NLI task body). We can see that the standard PLMs are easy to rewire if there is enough adaptation data. However, if the adaptation corpus contains fewer than 10 million tokens, the performance of the rewired standard PLMs (the blue dashed line in the figure) drops drastically as the adaptation data quantity goes down, from near 80 to around 35, a random-guessing level for NLI tasks. This motivates us to create more rewirable PLMs, i.e. PLMs with more plasticity so that the rewiring process can be faster and consume less data.

## 5.4 Pretraining with Active Forgetting

Recent works have shown that incorporating forgetting through iterative weights resetting can increase the "plasticity" of neural networks, enabling them to learn from small data and generalize better to unseen data in supervised learning [Alabdulmohsin et al., 2021, Taha et al., 2021, Zhou et al., 2022]. Building on these efforts, we study if we can bring such forgetting into the pretrain stage so that the resulting PLM would have more plasticity, allowing easier rewiring to new languages.

**Our Hypothesis.** In effect, when Artetxe et al. [2020] relearned the token embedding layer, the reinitialisation of the embeddings can be seen as forgetting applied *once* at the start of the language adapt stage. However, the PLM (specifically the transformer body) has never encountered forgetting before this stage and may struggle to handle this new situation. Without early exposure to forgetting, the PLM might suffer from slow recovery caused by forgetting before eventually benefiting from it. This inefficiency also implies a lack of plasticity in the Transformer architecture. During standard pretraining, token embeddings in the Transformer can encode excessive structures tied to the specifics of their training languages so that other parts of these models become overly rigid to the linguistic characteristics of the training language. The learning of a new lexical embedding layer in a PLM henceforth consumes lots of data in new languages along with long training horizons as shown in Section 5.3. In this chapter, to ensure swift learning of the new languages with both high sample efficiency and convergence rate, we argue that the PLM must be exposed to forgetting during pretraining, allowing itself to maximize the positive impact of forgetting and minimizing the cost of recovery.

**Our Method.** With this hypothesis in mind, we propose to add an *active forgetting* mechanism to the pretraining procedure, which resets the token embedding layer periodically as described in Algorithm 5. Concretely, the forgetting mechanism operates by intentionally clearing the weights of the embedding layer, which stores the static representations for all tokens, and reinitialising them to a new set of random values every $K$ gradient updates. Since pretraining involves advanced training strategies, like optimizers with states and learning rate schedulers, we also reset them together with the token embedding layer. We refer to language models pretrained with such active forgetting mechanism as *forgetting PLMs*, in contrast to *standard PLMs* which are pretrained in a standard way. The pretraining loss curve of a forgetting PLM is episodic (Figure 5.4), like in reinforcement learning or meta-learning. This episodic learning demonstrates that the active forgetting mechanism can introduce diversity without requiring

---
**Algorithm 5:** Active Forgetting Mechanism. The learning of token embedding layer is reset every $K$ updates.
---
**Input:** $K$: interval between two consecutive forgetting;
$n_{\text{body/emb}}$: current effective number of updates for the body or the token embedding layer;
$\alpha_{\text{body/emb}}$: current learning rate for the body or the token embedding layer;
$P^n_{\text{body/emb}}$: parameters after the $n^{\text{th}}$ update for the body or the token embedding layer;
$O^n_{\text{body/emb}}$: optimizer states after the $n^{\text{th}}$ update for the body or the token embedding layer;
$\Theta$: randomly initialised embedding parameters, each element drawn from $\mathcal{N}(0, 0.02)$;
$f$: function that computes the gradients w.r.t. the parameters using the sampled data;
$g$: function that updates the parameters based on the gradients (e.g., one step in Adam optimizer);
$s$: function that updates the learning rate (e.g., one step in the polynomial learning rate scheduler).
**Output:** The updated parameters and optimizer states:
$P^{(n)} = \{P^{(n)}_{\text{emb}}, P^{(n)}_{\text{body}}\}$,
$O^{(n)} = \{O^{(n)}_{\text{emb}}, O^{(n)}_{\text{body}}\}$.
$n_{\text{emb}} \leftarrow n \mod K$;
$\alpha_{\text{body}} \leftarrow s(n_{\text{body}})$ // Adjust learning rate for body based on $n$;
$\alpha_{\text{emb}} \leftarrow s(n_{\text{emb}})$;
$G^{(n)} \leftarrow f(P^{(n-1)}, \cdot)$ // Compute all gradients;
$P^{(n)}_{\text{body}}, o^{(n)}_{\text{body}} \leftarrow g(G^{(n)}_{\text{body}}, P^{(n-1)}_{\text{body}}, o^{(n-1)}_{\text{body}}, \alpha_{\text{body}}, n)$ // Update the transformer body;
**if** $n_{emb} == 0$ **then**
 $P^{(n)}_{\text{emb}} \leftarrow \Theta$ // Reset token embeddings and relevant optimizer states;
 $o^{(n-1)}_{\text{emb}} \leftarrow 0$;
$P^{(n)}_{\text{emb}}, o^{(n)}_{\text{emb}} \leftarrow g(G^{(n)}_{\text{emb}}, P^{(n-1)}_{\text{emb}}, o^{(n-1)}_{\text{emb}}, \alpha_{\text{emb}}, n_{\text{emb}})$ // Update the token embeddings;
---

actual new data. Each forgetting event kind of "branches out" a novel environment for the model to explore, as if initiating a new episode of learning.

**Research Questions.** To further examine the proposed forgetting mechanism, we compare *forgetting PLMs* and *standard PLMs* on sample efficiency and convergence speed during language adapt, two key aspects of model plasticity. Our research investigates:

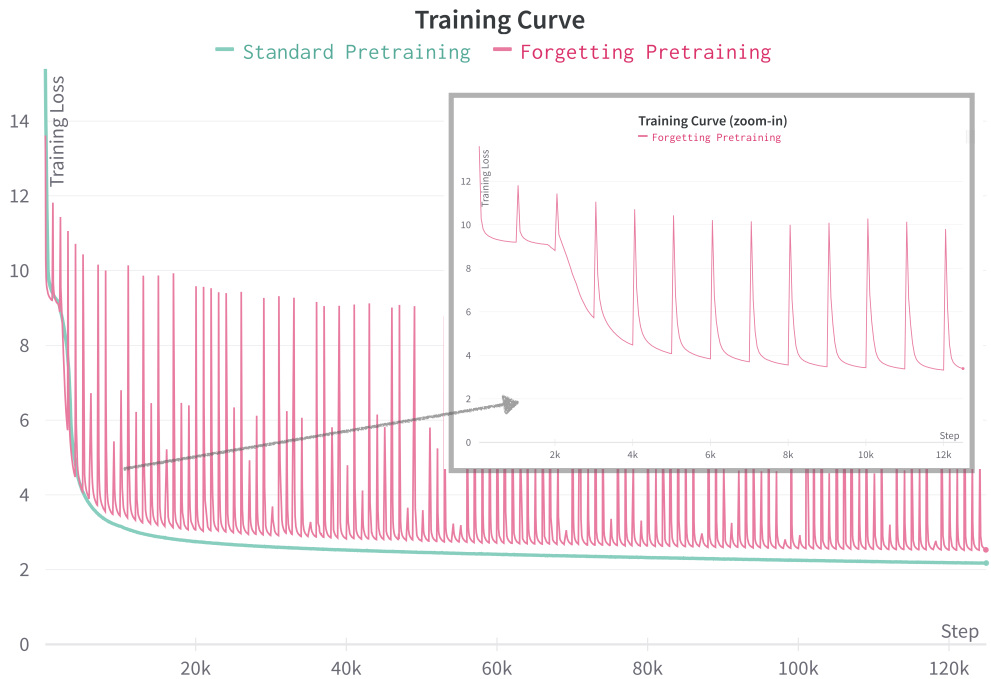

Figure 5.4: Pretraining losses of forgetting and standard language models. The forgetting mechanism brings an episodic pattern into the loss curve: every embedding forgetting produces a loss spike, from which the model learn to recover. Through such repeats of forget-relearn, the model gets used to learn new embeddings from scratch.

- RQ1: Real-world low-resource languages often have scarce data for adapting models. Does pretraining with active forgetting impart enough plasticity to forgetting PLMs, enabling them to learn new languages even with such limited data?

- RQ2: Deploying PLMs frequently encounters computational limitations. Endowed with more plasticity, can forgetting PLMs reduce adaptation time for such low-compute scenarios?

- RQ3: New languages may be very similar or different from pretraining languages. Does this similarity/difference impact the relative benefit of forgetting PLMs over standard PLMs?

## 5.5 Experiments

To evaluate the effectiveness of forgetting PLMs and address RQ1-RQ3, we conduct experiments on several cross-lingual transfer benchmarks.

### 5.5.1 Experimental Setup

In our work, we closely follow the setup in Artetxe et al. [2020] and Marchisio et al. [2023]. Our pretraining model is RoBERTa-base, a standard 12-layer transformer-based language model. We trained for each language a sentencepiece tokenizer [Kudo and Richardson, 2018] with a vocabulary size of 50K over the corresponding data subsets in CC100. The model was pretrained with the English subset of the CC-100 dataset. The pretraining process consists of 125K updates, with a batch size of 2048. We used a learning rate scheduler with linear decay and an initial learning rate of $7e-4$, with 10K warm-up updates. Checkpoints were saved every 500 updates. Since longer pretraining consistently led to better validation perplexities in our experiments, we chose the final pretraining checkpoint (step 125K) whenever possible for optimal performance. Since the final checkpoint might coincide token embeddings reset in forgetting pretraining, we instead chose the closest checkpoint that has the best validation perplexity. This ensured that we selected the best pretrained checkpoints for both approaches based on when they achieved their optimal validation perplexities. We set the frequency of forgetting K = 1000 and used a clip-norm of 0.5.

During the language adapt stage, we kept most of the hyperparameters the same as for pretraining. We finetuned the token embedding layer while keeping the others frozen, as described in Section 5.3. This differs from the pretraining setup, where all parameters are learnable to maximize learning speed. In contrast, the finetuning setup is intended to mimic how humans might typically relearn word meanings: by updating embeddings while keeping the rest of the system fixed. Note that *no* forgetting happens during this stage because we want the models to learn the new languages as well as possible. In the task adapt stage, both models were finetuned for 10 epochs on the English task data, specifically MultiNLI [Williams et al., 2018] for the NLI task and SQUAD Rajpurkar et al. [2016] for the QA task. After the assemble stage, we evaluate the zero-shot performance of the assembled model on XNLI [Conneau et al., 2018], a cross-lingual NLI task, along with XQuAD [Artetxe et al., 2020] and MLQA [Lewis et al., 2020a], two cross-lingual QA tasks. We report the NLI accuracy and QA F1 on the test sets.

Table 5.1: Accuracy comparison of forgetting and standard PLMs on the XNLI dataset (table continues).

| Method | vi | sw | es | bg | de | fr | el | ru |
|---|---|---|---|---|---|---|---|---|
| Standard | **65.8** | 55.6 | 68.0 | 65.5 | 62.2 | 63.5 | 63.1 | 56.9 |
| Forgetting | 62.8 | **59.5** | **74.0** | **71.7** | **68.5** | **71.2** | **70.8** | **65.8** |
| Gain(%) | −4.6 | +7.0 | +8.8 | +9.5 | +10.1 | +12.1 | +12.2 | +15.6 |

Our experiments were implemented using fairseq [Ott et al., 2019]. The pretraining and language adaptation experiments were conducted on 32 Tesla V100 GPUs (each with 32 GB memory) and took approximately 24-36 hours to complete. The time taken for both stages were quite close to each other even though the latter only involved tuning the embeddings. This demonstrates the importance of reducing the computational cost of the language adaptation stage.

Differing from prior work [Artetxe et al., 2020, Marchisio et al., 2023], we focus on language adapt in low-data regimes. We simulate low-resources scenarios by limiting the adaptation data for each downstream language to only 5M subword tokens from CC100. This is in contrast with conventional setups, where all the tokens in the corresponding languages in CC100 are used for language adaptation. As Table C.2 shows, such setups consume several orders of magnitude more data than our 5M-token setup; for instance, the Swahili CC100 subset contains 345M tokens, roughly 69 times larger than our corpus, and the Russian subset contains 34.9B tokens, roughly 6,980 times larger. Therefore, PLMs that can successfully learn new languages with rich data under traditional setups may struggle to do so with our limited 5M-token corpus.

### 5.5.2 RQ1: Forgetting PLMs Work Better in Low-Data Regimes

Standard PLMs struggle in low-data language adaptation, dropping from 86.1 English NLI accuracy to just 53.3 average accuracy on XNLI with limited 5M token adaptation data. Compared to prior work which uses full data from Wikipedia [Artetxe et al., 2020] or from CC100 [Marchisio et al., 2023], the average accuracy on XNLI drops about 18% (from 66.8/66.3 to 53.3). This indicates standard PLMs are not coping well with the low-data regime. In contrast, forgetting PLMs achieve decent 62.7 average XNLI accuracy, a +21.2% relative gain over standard PLMs, as shown in Table 5.2.

Table 5.2: Accuracy comparison of forgetting and standard PLMs on the XNLI dataset (table continued). On average, forgetting achieve a 21.2% relative gain in accuracy compared to standard across the languages tested, where averaged relative gain = $\frac{\sum_{x \in \{\text{languages}\}} \text{Relative Gain of } x}{\text{\#Languages}}$.

| Method | zh | ur | hi | tr | ar | th | Avg | en |
|---|---|---|---|---|---|---|---|---|
| Standard | 53.2 | 36.8 | 39.7 | 38.9 | 41.2 | 35.3 | 53.3 | **86.1** |
| Forgetting | **63.5** | **45.8** | **52.9** | **52.7** | **59.5** | **59.7** | **62.7** | 85.1 |
| Gain(%) | +19.4 | +24.5 | +33.2 | +35.5 | +44.4 | +69.1 | +21.2 | −1.2 |

Table 5.3: F1-score comparison of forgetting and standard PLMs on MLQA. On average, forgetting PLMs achieve a 33.8% relative gain in F1 compared to standard PLMs across the languages tested, where averaged relative gain = $\frac{\sum_{x \in \{\text{languages}\}} \text{Relative Gain of } x}{\text{\#Languages}}$.

| Method | es | vi | de | zh | hi | ar | Avg | en |
|---|---|---|---|---|---|---|---|---|
| Standard | 49.4 | 38.3 | 45.3 | 34.1 | 17.7 | 20.8 | 34.3 | **78.9** |
| Forgetting | **55.3** | **45.0** | **53.4** | **43.0** | **28.8** | **34.7** | **43.4** | 78.3 |
| Gain(%) | +12.0 | +17.6 | +17.8 | +26.2 | +62.5 | +67.0 | +33.8 | −0.8 |

Forgetting PLMs also outperform standard PLMs on MLQA and XQuAD, with average F1 relative gains of +33.8% and +60.9% across languages, as respectively demonstrated in Table 5.3, Table 5.4 and Table 5.5. Across NLI and QA tasks, forgetting PLMs consistently surpass standard PLMs in low-data regimes. Why do forgetting PLMs handle the low-data regime better? We hypothesize this is because forgetting PLMs are more robust to different embedding initialisations. They encode more universal knowledge in the transformer body. Standard PLMs may encode more "shortcut" knowledge relying on certain embedding initialisations. In low data, standard PLMs cannot adjust embeddings towards shortcut routes without access to enough data. Forgetting PLMs do not rely on shortcuts so perform better.

Table 5.4: F1-score comparison of forgetting and standard PLMs on XQuAD (table continues). On average, forgetting PLMs achieve a $60.9\%$ relative gain in F1 compared to standard PLMs across the languages tested, where averaged relative gain $= \frac{\sum_{x \in \{\text{languages}\}} \text{Relative Gain of } x}{\#\text{Languages}}$.

| Method | vi | es | ru | de | el | zh |
|---|---|---|---|---|---|---|
| Standard | 49.7 | 57.7 | 49.4 | 50.9 | 48.5 | 32.4 |
| Forgetting | **52.9** | **64.6** | **56.5** | **60.9** | **59.9** | **43.7** |
| Gain(%) | +6.4 | +12.0 | +14.5 | +19.7 | +23.6 | +34.6 |

Table 5.5: F1-score comparison of forgetting and standard PLMs on XQuAD (table continued). On average, forgetting PLMs achieve a $60.9\%$ relative gain in F1 compared to standard PLMs across the languages tested, where averaged relative gain $= \frac{\sum_{x \in \{\text{languages}\}} \text{Relative Gain of } x}{\#\text{Languages}}$.

| Method | hi | ar | th | tr | Avg |
|---|---|---|---|---|---|
| Standard | 21.4 | 22.2 | 15.4 | 13.0 | 36.1 |
| Forgetting | **33.3** | **38.7** | **38.4** | **41.4** | **49.0** |
| Gain(%) | +55.8 | +74.2 | +149.7 | +218.8 | +60.9 |

### 5.5.3 RQ2: Rewiring Forgetting PLMs Requires Fewer Updates

We are also interested in how quickly forgetting PLMs and standard PLMs can learn new languages. Figure 5.5 summarizes adaptation curves on XNLI, MLQA and XQuAD, with each point representing the averaged performance across all languages. In just $5K$ steps (4% of full adaptation), forgetting PLMs reach $57.8$ accuracy on XNLI while standard PLMs struggle at random guessing levels of $37.2$. Similar trends hold for MLQA and XQuAD. After $5K$ steps, forgetting PLMs achieve 92% of their full performance on XQuAD versus just 53% for standard PLMs (see the last plot in Figure 5.5).

Why do forgetting PLMs converge faster? We hypothesize it is because periodical embedding resetting forces the body to gradually locate itself on a particular manifold, where it can easily cooperate with new embeddings. This makes the body encourage larger embedding updates when adapting to new languages. Active forgetting simulates language switching during pretraining[2] introducing diversity without new data. This allows faster adaptation to real new languages.

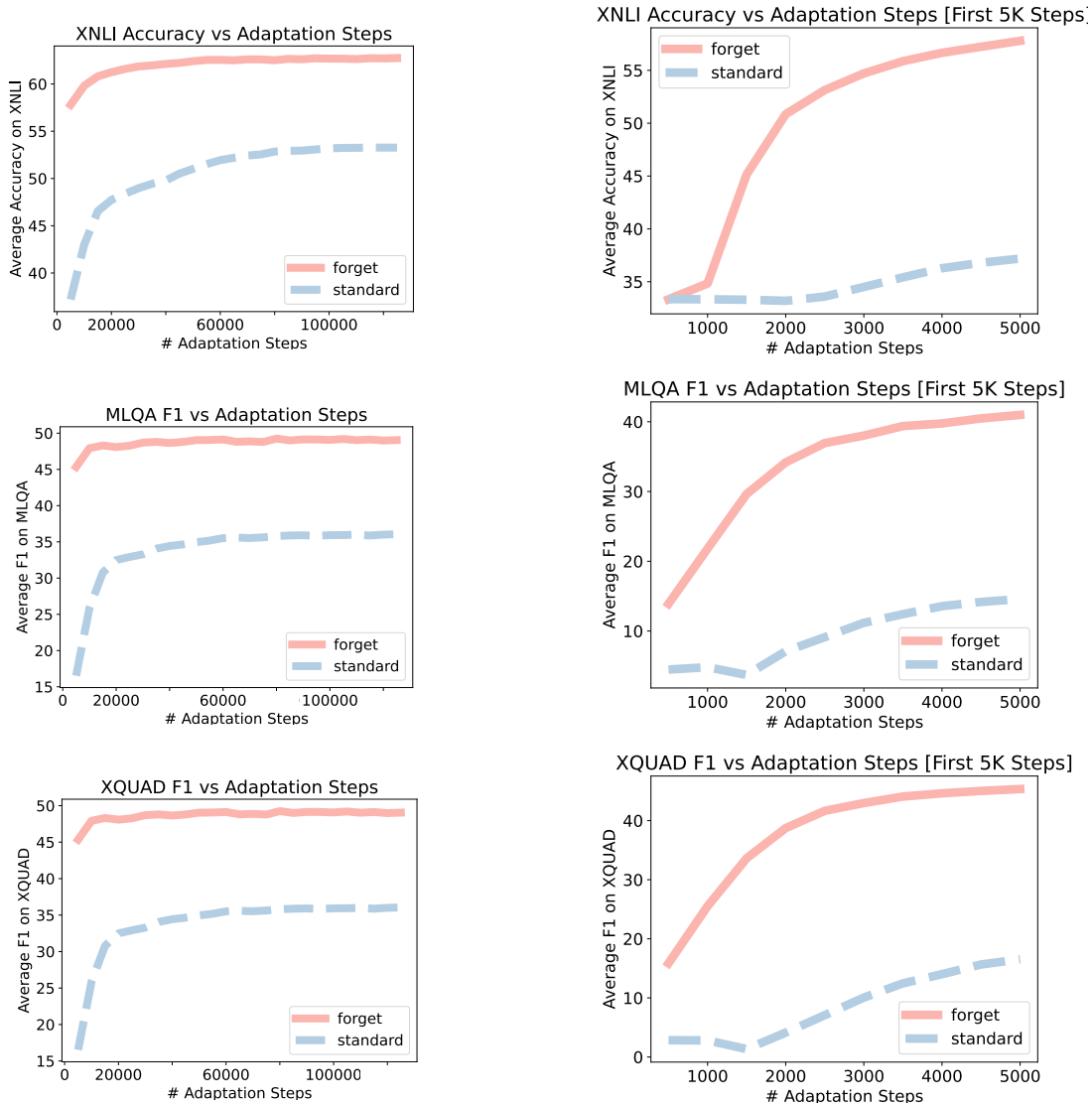

Figure 5.5: Adaptation curves on XNLI, MLQA, and XQuAD. Numbers aggregated across languages. The first row contains the full adaptation curves, which comprises 125K adaptation steps. The second row contains the zoom-in versions of curves for the first 5K adaptation steps. Forgetting PLMs converge faster than standard PLMs; for instance, on XQuAD (the last plot), forgetting PLMs reach 92% of their final performance within 5K updates, while standard PLMs only reached 53% of their final performance at that point.

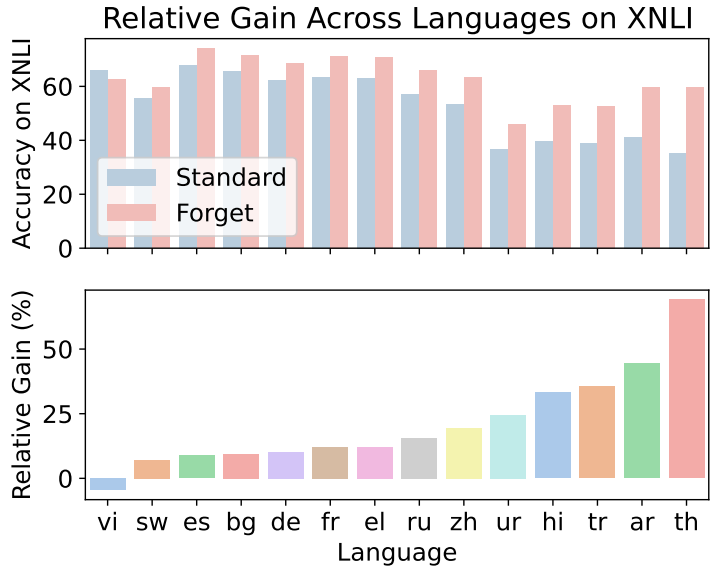

Figure 5.6: Relative gains of forgetting PLMs over standard PLMs across languages for XNLI. Forgetting yields substantial relative gains for languages like Arabic, Hindi, Thai, Turkish, and Urdu.

### 5.5.4  RQ3: Distant Languages Benefit From Forgetting PLMs

We have primarily focused on discussing the averaged performance in the previous sections (Sec 5.5.2 and 5.5.3). In this section, we provide a more detailed comparison of language-specific performances between forgetting PLMs and standard PLMs on XNLI, MLQA, and XQuAD. To gain a deeper insight into which languages benefit the most from the use of forgetting, we present the relative performance changes across the languages in Figure 5.6 for XNLI and in Figure 5.7 for MLQA. For space reason, the results of XQuAD can be found in Figure C.1 in the appendix.

Across the spectrum of languages (Table C.1), we observe that forgetting provides greater benefits for languages distant to the pretraining language (English) in terms of language family, script and morphology. Specifically, forgetting brings large rela-tive gains for languages such as *Arabic*, *Hindi*, *Thai*, *Turkish*, and *Urdu* compared to closer languages like *German*. Script seems important - forgetting helps Vietnamese and Swahili less despite their distance from English, likely due to the shared Latin script.

---
[2]Precisely, it simulates vocabulary swappings, causing drastic changes to the input of the body.

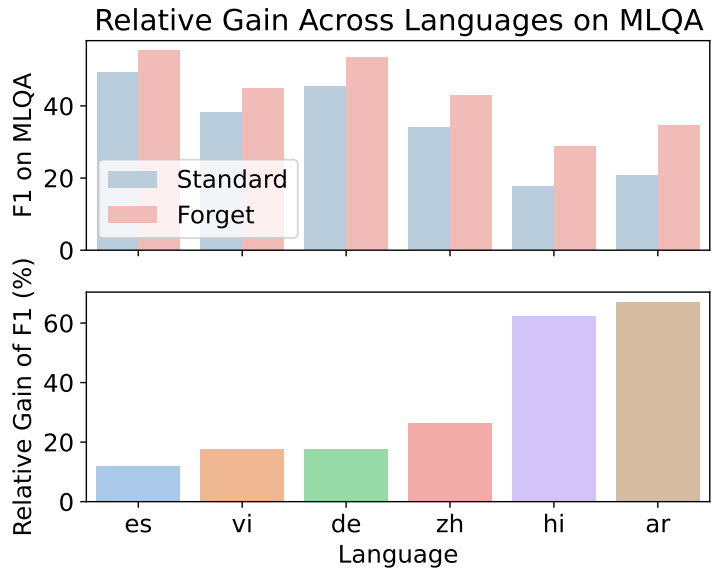

Figure 5.7: Relative gains of forgetting over standard across languages for MLQA. For languages closely related to English, such as German, the relative gains from forgetting are modest.

Languages that share a script with the pretraining language (e.g., English and German) tend to share subword tokens, enabling models to reuse learned embeddings and lexical patterns. This facilitates transfer and reduces the need to relearn low-level representations. In contrast, languages with different scripts (e.g., Arabic, Hindi, Thai) have minimal subword overlap and lack orthographic familiarity, making tokenization and representation learning more difficult. Script similarity, therefore, narrows the representational gap in cross-lingual transfer. Forgetting is more beneficial for script-divergent languages, as it enables the model to construct new, script-specific representations without interference from English.

Examining adaptation curves within the first 5K steps, forgetting PLMs reach substantially superior performance over standard PLMs for almost all languages except Urdu, while standard PLMs struggle at random guess levels (see Figure 5.8 and Section C.2). This demonstrates forgetting PLMs' ability to efficiently adapt to new languages,

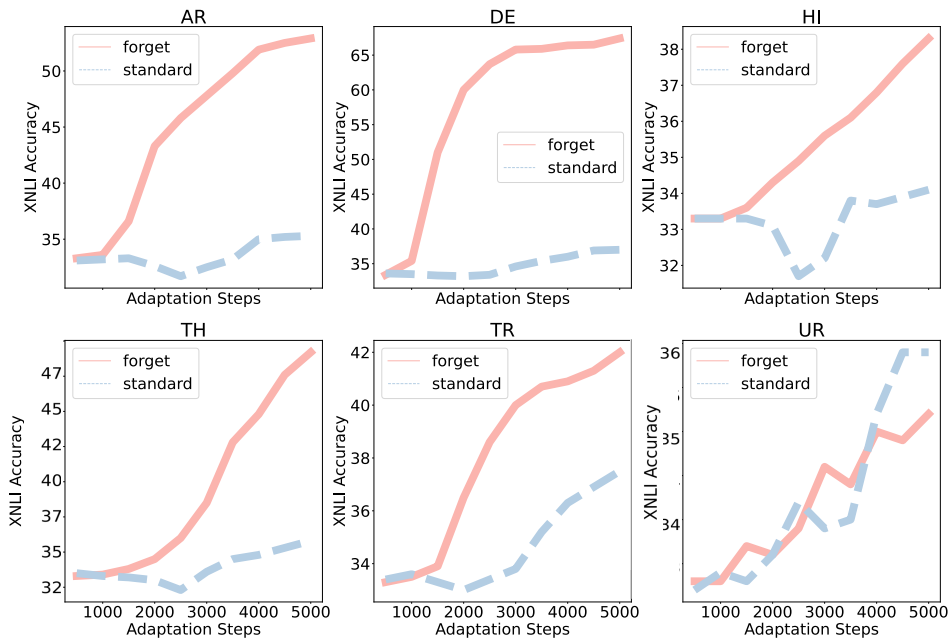

Figure 5.8: Adaptation curves on XNLI within 5K updates for individual languages: Bulgaria, Greek, Spanish, French, Russian, Swahili, Vietnamese and Chinese. For all languages except Urdu, the forgetting PLMs converge faster than the standard PLMs during the language adaptation stage.

particularly dissimilar ones, in low-data settings.

## 5.6 Discussion

**Summary** This chapter expands on the idea of *active forgetting*, a manifestation of the destructuring principle, and its potential impact on AI models. While Chapter 4 demonstrated the value of active forgetting in the structured paradigm for building general knowledge engines, this chapter applies it to unstructured paradigms, showing that *active forgetting* can improve pretrained language models by imbuing them with more linguistic plasticity. Experiments with RoBERTa show that models pretrained via active forgetting can better learn from **small data** while enjoying faster convergence during language adaptation, particularly for languages that are distant from English.

Most current efforts to build knowledge engines in the unstructured paradigm have been focusing on ingesting more data into larger models [Kaplan et al., 2020]. Accel-erating techniques on both hardware and software sides are being developed to help us achieve such *structuring* of the reality (whether real or synthetic) into machine computation. On the other side, we as a community seem to have far fewer ideas on how we can **rewire** inappropriate structures from the models safely, timely, and relevantly [Weidinger et al., 2021, 2022, Kirk et al., 2024]. This chapter stands at the crossroad of structuring and destructuring, where we highlight the necessity of *destructuring* in its role for "machine plasticity" – a kind of freedom to delete built-in structures and rewire model behavior whenever needed. We **speculate** that destructuring may reduce the model's reliance on shortcut learning, where models depend on superficial cues rather than deeper structure [Geirhos et al., 2020]. By disrupting these shortcuts, destructuring could encourage the model to focus on more abstract patterns, potentially improving its ability to generalize to new environments.

The conclusion of this chapter, a dual focus on structuring and destructing, is surprising while providing a promising alternative to the scaling approach [Kaplan et al., 2020]. Destructuring can drive model evolution and rewire models to adapt to the dynamic world. Without this capacity for machine plasticity, we risk creating rigid AI systems that potentially trap their human users in outdated or biased "knowledge". A balance between structuring and destructuring opens the door to create more natural and flexible knowledge engines, ultimately supporting diverse AI applications that blend into our everyday life.

**Implications** Going beyond language adaptation, we argue that pretrained language models with more plasticity are a promising direction for future research, as they allow easier adaptation to various tasks, domains, languages and can evolve faster as the real world changes. Unlike symbolic methods, such as knowledge graphs, which can easily rewire a fact by modifying the corresponding knowledge triplet, current static PLMs are harder to rewire since changing one fact by updating model weights may disrupt multiple other facts without substantial post-hoc intervention. Improving the rewirability via forgetting pretraining thus can be seen as one way of imbuing PLMs with similar benefits as symbolic methods (making the resulted model more controllable i.e. can be modified with tiny cost), complementing the line of post-hoc model editing research [Mitchell et al., 2021, 2022].

**Limitations** This chapter uses one of the simplest forgetting approach - directly resetting embeddings to random initialisation. Advanced techniques like gradually injecting noise could be explored. We focus on masked language modelling pretraining with language-specific tokenizers. Applying active forgetting to autoregressive LMs, other pretraining methods (e.g. DeBerta pretraining [He et al., 2021b,a]), and various tokenization strategies is promising future work. More broadly, current large language models need more plasticity to expand across tools, tasks, and domains. Our work takes an initial step, showing that directly resetting embeddings can significantly improve model plasticity. Further research on more sophisticated forgetting techniques during pretraining could unlock additional gains.

On the theory front, potential connections can be made between forgetting and meta-learning [Schaul and Schmidhuber, 2010, Thrun and Pratt, 2012, Andrychowicz et al., 2016, Finn et al., 2017] since both attempt to learn solutions that can quickly adapt themselves to new inputs. Another possible theoretical explanation for why active forgetting works so well might be related to the flatness of the solution in the loss landscape [Alabdulmohsin et al., 2021]. Flatter minima tend to enjoy better generalization [Liu et al., 2023b]. Thus, it might be worthwhile to study the flatness of the transformer body during the forgetting pretraining.

Beyond methodology, it would be valuable to more deeply investigate how this periodic resetting of embeddings affects the internal dynamics of the Transformer architecture itself. For instance, how does the reset influence attention patterns, layer activations, or representational drift across training epochs? Such analysis could shed light on whether active forgetting encourages more modular or adaptive representations. Additionally, while this work focuses on input embeddings, the same principle could be extended to other components such as attention heads or feedforward layers to improve plasticity further.

# Summary of *Destructure*

Explicitly or implicitly, both structured and unstructured AI paradigms rely on structures to represent knowledge succinctly and effectively, as shown in Part I: *Structure*. However, a shared challenge for both paradigms lies in the fact that structures can be overly rigid when faced with unseen environments. What were once the foundation for efficient and consistent reasoning may turn into an outdated lens, distorting the model's ability to perceive beyond the familiar and adapt to novel scenarios – what we call model plasticity. Part II: *Destructure* addresses this shared challenge by integrating destructuring techniques into training. Experiments in both the structured paradigm and unstructured paradigm show this method helps models adapt to new knowledge graphs and languages.

We begin by examining how structures are carved into models (Chapter 4). The embedding layer, often overlooked in modern models for knowledge engines, is nevertheless key to understanding this process. Rather than viewing embeddings as a set of isolated vectors, we unfold their gradient decent traces and interpret these traces as message-passing between corresponding symbols. Similar to graph neural networks, the message-passing propagates information along "edges", which can be triples in knowledge graphs or sentences in text corpora.

This reinterpretation explains transductiveness, or why models wrapped by embeddings on both ends fail with unfamiliar inputs. During long training horizons, embeddings, as repositories of all message-passing computation, store excessive global structural information about the training symbols; the non-embedding model body ends up maintaining contextual knowledge to be triggered by specific known embeddings. While excessive global information benefits transductive tasks – where all symbols are known – it leaves non-embedding model body ill-equipped to handle unseen symbols. Models thus fail for inductive scenarios, where new symbols lack embedding values that can trigger the body's knowledge accordingly.

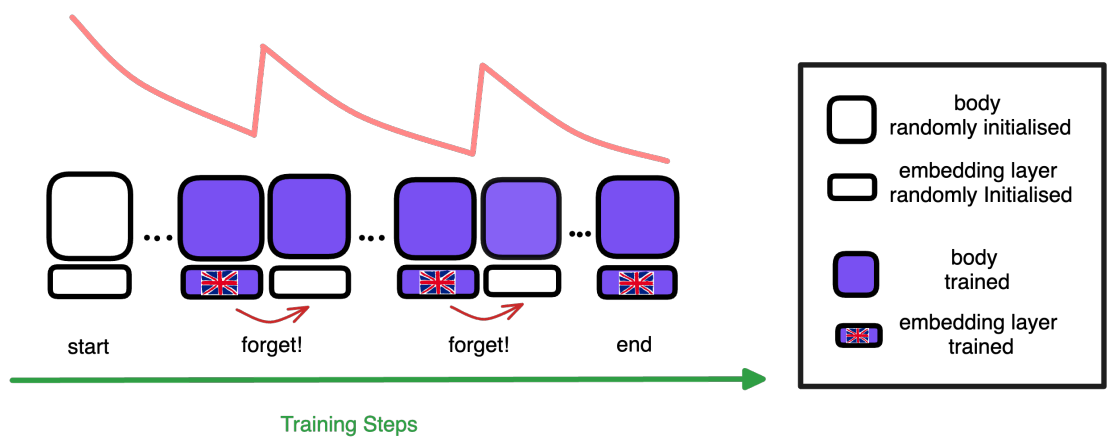

Figure 5.9: The active forgetting mechanism periodically resets the embedding layer during training. Pink curve illustrates the loss changes. Whenever forgetting happens, the loss curve spikes and then recovers to a normal downward trend.

To address this, we propose to introduce destructuring techniques into standard training. Specifically, the active forgetting mechanism, resets the embedding learning periodically while keeping the rest of the model training intact. This technique allows the model body to learn to regrow embeddings from scratch after each embedding resetting. The new training procedure derives a bi-level learning system: a fast inner loop for regrowing embeddings and a slow outer loop for learning a robust, stable body. Regular destructuring of embeddings forces the body to *"re-view" the data with a pair of fresh eyes*,[3] in a more abstract way that does not pertain to embedding value nuances but focus more on relationships between symbols. Empirical studies on inductive inference over graphs and languages demonstrate that this mechanism improves generalization to unseen symbols, such as new entities in knowledge graphs (Chapter 4) and new tokens from an unfamiliar language (Chapter 5).

The beauty of reality lies in its potential infiniteness.[4] On one hand, to simplify and understand reality with limited cognitive resources, human brains sketch it with conceptual structures. On the other hand, the human brain's neuroplasticity allows us to revise outdated conceptual structures. Balancing between structuring and destructuring

---

[3]In non-scientific texts, studying the old with a pair of new lenses is sometimes known as Onkochishin. Embeddings in a transformer or a factorization based models can be thought of as the "lens/eyes" for the non-embedding model body.

[4]The future is unknown but never affects its beauty.

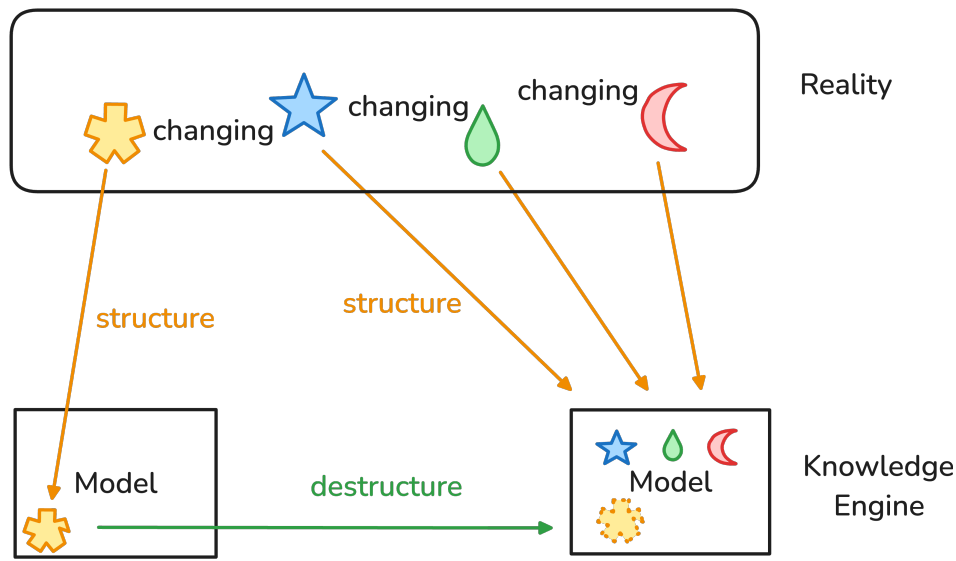

Figure 5.10: The reality is always changing. We use different shapes to indicate the observed structures of reality. These structures change as time flows. To faithfully capture reality, the knowledge engine, no matter in the structured paradigm or unstructured paradigm, must be capable of balancing the force of structuring with the force of destructuring so that it captures necessary structures but also does not get trapped by these structures.

leads to the plasticity crucial for surviving, navigating, and thriving in a dynamic reality.

Similarly, when building knowledge engines, the first attempt is to model reality comprehensively with computational structures. These structures are reusable across applications, providing knowledge efficiently without the need to recompute things from scratch. They model the *known* aspects of the reality quite well. However, we must recognize that such structures can become inaccurate as reality continually evolves. Therefore, it is crucial for knowledge engines to possess the ability to delete outdated structures and relearn new ones. In other words, they must be equipped to handle the unknown dimensions of reality. Destructuring, in this context, provides knowledge engines with the capacity to discard rigid frameworks and free up resources to address new environments. This capability helps model the unknown more effectively by accommodating randomness and avoiding excessive reliance on past solutions.

Following this line of thought, Part II: *Destructure* examines the counterforce to Part I: *Structure*. This section emphasizes the potential of destructuring through active forgetting as a mechanism to enhance model plasticity. By incorporating this approach, we can develop resilient and robust knowledge engines that evolve alongside the ever-changing world. This approach opens exciting questions for future research, such as whether components beyond embeddings should be subject to forgetting and which tasks might further benefit from active forgetting.

# Closing

## Chapter 6

# Conclusions and Critical Reflections

We have now arrived at the closing part, where we will summarize this thesis and discuss its limitations, significance, and potential future directions. This chapter will present the main conclusions of this thesis along with critical and systematic reflections on the limitations of the thesis.

## 6.1 Conclusions and Contributions

While intelligence has long been a quest for human beings, we now stand at a critical point in time. We are experiencing an intelligence revolution, where in the envisioned future, intelligence can be packed into units that can be disseminated easily across time and space, akin to how Industrial Revolution packs our physical capabilities into units. In this revolution, knowledge plays a crucial role as it serves as the interface between our cognition and the reality. An accurate knowledge interface allows intelligent agents to conceptualize and model the reality effectively (even though it remains uncertain if humans experience the reality directly). The knowledge of actions and their consequences further allows the intelligent agents to intervene and transform their environments. Hence, building knowledge engines are essential to both natural and artificial intelligence.

There are two conventional paradigms to constructing knowledge engines: the structured and unstructured paradigm, exemplified respectively by knowledge graphs and large language models. Which one is better? This thesis aims to discuss this age-old debate in the context of recent findings in knowledge graphs and language models. We argue that the presence of structures is inevitable regardless of whether data is explicitly structured (as in knowledge graphs) or implicitly structured (as in large language models). Furthermore, we assert that not all structures are of positive roles. If focusing narrowly on structure formation, we can make models that are overly rigid. Thus, we motivate the necessity of destructuring, which improves models' plasticity so that they can learn rapidly with few examples in new environments.

In summary, this thesis presents a scientific journey to discover the commonalities between the two mainstream paradigms for building knowledge engines. Although these paradigms initially appear distinct, often perceived as separate approaches, this thesis demonstrates that deeper connections can be established through a functional examination of model training dynamics and analytical reformulation of model computations. The contributions of this thesis are fourfold as detailed below:

**First**, the thesis identifies new connections between the two seemingly disjoint paradigms as summarized in Table 6.1:

- The language modelling objective induces latent structures within model computations, supporting tasks such as knowledge base completion and the interpretation of large language models.

- Active forgetting enables inductive reasoning in *both* paradigms, facilitating efficient generalization to unseen entities in knowledge graphs and new languages in pretrained language models.

**Second**, we provide new insights into the role of structures in building general knowledge engines:

- Structures are indispensable for knowledge engines, though they can manifest in various forms – explicit in data or implicit within models. The structured paradigm explicitly specifies the structures in the data. For the unstructured paradigms, latent structures about the relationships among tokens can be directly extracted from model computations post-training (Chapter 3).

- However, overly encoding structures within models can hinder their ability to generalize to unseen scenarios, highlighting the importance of balancing structures and flexibility (Chapter 4).

Table 6.1: Comparison of the structured and unstructured paradigm through the dual forces of structure and destructure.

| Force | Structured Paradigm | Unstructured Paradigm |
|---|---|---|
| **Structure** | Language modelling objectives induce structure into factorization models *(Chapter 2)* | Language modelling objectives induce structure into Transformers *(Chapter 3)* |
| **Destructure** | Active forgetting enhances generalization to unseen graphs *(Chapter 4)* | Active forgetting enhances generalization to unseen languages *(Chapter 5)* |

**Third**, we obtain new understandings about the embedding layer, the often-overlooked components in *both* paradigms:

- The concept of the *embedding sandwich* emerges as a suitable architectural abstraction for models in both the structured and unstructured paradigm, e.g. transformers (Chapter 3 and Chapter 5).

- Embeddings should not be examined in isolation but rather in conjunction with their optimization dynamics. They serve as dynamic repositories where gradient descent accumulates, propagates, and stores symbolic interactions. (Chapter 4).

**Finally**, our findings advocate for a shift in focus from the surface-level distinctions of structured versus unstructured data to the underlying dynamics of structure formation and destructuring as indicated by Table 6.2:

- Rather than focusing solely on whether data is structured or unstructured, especially given modern datasets often contain data exhibiting varying degrees of structures, we highlight the need to study both the forces driving structure formation (Part I) and their opposing force, destructuring (Part II).

- Structuring promotes structure encoding in the models. Destructuring mitigates the rigidity of excessive structuring, enabling AI systems to adapt and reason effectively in dynamic, unseen environments—an ability we term *model plasticity*.

In conclusion, this thesis highlights that structure formation and its dual force, destructuring, are both essential components for building general knowledge engines.

Table 6.2: The unified paradigm seen through the mechanistic forces of structure and destructure.

| Force | The (Un)Structured Paradigm |
| --- | --- |
| **Structure** | Language modelling induces structure in model computation |
| | *(Part I)* |
| **Destructure** | Active forgetting helps address unseen symbols and adapt to new environments |
| | *(Part II)* |

## 6.2 Limitations and Flaws

While this thesis provides new insights into the bridging of structured and unstructured learning, the thesis contains several limitations and flaws in its current form. While individual chapters already provide discussion on their own limitation, this section acknowledges global limitations related to the topic of structured and unstructured learning so that the readers can have a rigorous assessment of the thesis.

### 6.2.1 Theoretical Scope

We discussed several key constructs and concepts used in our thesis, where broader notions of them are combed through.

**Structure**

The core concept in this thesis is *structure*. In the traditional discussion of structured and unstructured learning, the concept of structure mainly centres around the structures in the data. This thesis takes a step further to discuss the relationship between the structures in the data and the structures in the computational model: language modelling objective can induce the former into the latter; the latter can in turn be recovered to the former by rearranging model computation. In this sense, the thesis considers primarily structures in the context of relational learning and language modelling.

However, structures have other broader notions which the thesis could have engaged with. We enumerate a number of them to better contextualize our notion of structures.

First, structures are the obsession object for philosophers, psychologist, educators who strive to understand human minds. In this context, structures typically refer to mental or cognitive structures, with knowledge being perhaps one of the most important of these structures. We review several famous notions of structures under this category. As early as 1781, Kant discussed how the mind structures experience and how such abstractions form the basis for humans to understand the world [Kant, 1781]. Clinical methods were used to study how these cognitive structures form in children by Piaget in 1920s [Piaget, 1929]. Using controlled observations, Vygotsky further highlighted the mental structures are highly impacted by external factors such as language and culture [Vygotsky, 1934]. Bruner examined the role of mental structures in the learning process and showed how abstract thinking is necessary to organize new experiences and knowledge [Bruner, 1960]. More recently, Deleuze further argued that mental structures are not static but dynamic and emergent in his masterwork, *Difference and Repetition* [Deleuze and Patton, 1994]. This thesis can be seen as an effort towards implementing such a dynamic notion of mental structures in AI systems (in fact, to a certain extent, one can see our destructure process as Deleuze's difference process and our structure formation as Deleuze's repetition process), while more understandings into the difference and repetition processes are required to fully realize the flexible structures described by Deleuze.

Secondly, in programming, structures often mean data structures, the abstract models for organizing and storing data [Knuth, 1997]. In this case, structures refer to an abstraction where the physical implementation is often hidden, and developers interact with abstract representations of the data. In our thesis, the "structure" in the structured and unstructured paradigm refer to the structures in the training data. Specifically, in the case of structured paradigm, the structures are relational structures formatted in subject-relation-object triples [Ji et al., 2020]; in the unstructured paradigm, the texts are without such formatting, e.g. the first few tokens are not necessarily the subject rather they can play various grammar roles depending on the contexts.

Thridly, structures in mathematics, such as sets, groups, and graphs, are abstractions that represent relationships [Dummit et al., 2004, Hausdorff, 2021, Deisenroth et al., 2020]. These mathematical constructs are essential in modelling relationships and complex systems. In the structured paradigm, knowledge bases can be represented using graphs, known as knowledge graphs. The graph representation of knowledge bases en-able easy visualization, comprehension, and efficient querying, reasoning [Noy et al., 2019, Ji et al., 2020]

Finally, structures have ample notions in machine learning. Most of these notions explcitly incorporate structures into learning and learnt structures are used to help reasoning. We discuss a couple of works with such explicit structure integration. In Bayesian learning, structures manifest as Bayesian networks, which are probabilistic graphical models consisting of variables and their conditional dependencies expressed by a directed acyclic graph [Neal, 2012]. Compared to the usual neural networks, Bayesian networks can be used for prediction with uncertainty. Similarly, in causal inference, structures often refer to causal graphs or structural causal models (SCMs) [Pearl, 1998]. These causal models describe causal relationships between variables, distinguishing causation from correlation [Pearl and Shafer, 1995, Pearl and Mackenzie, 2018]. Another branch of work, neuro-symbolic AI [Besold et al., 2021], integrates symbolic structures with neural networks. Since neural networks excel at pattern recognition and symbolic reasoning excels at abstract concepts and logic rules, neuro-symbolic AI aims to combine their strengths [Garcez et al., 2019]. In this domain, structures often refer to knowledge representations such as knowledge graphs and logic rules [Hamilton et al., 2024, Colelough and Regli, 2024]. Thus, the knowledge graphs based learning methods in this thesis can apply to some neuro-symbolic AI systems while more research are needed to extend the methods to complicated structures like logic rules. For example, it would be interesting to explore how active forgetting (Part II) could help models adapt logic rule templates by flexibly substituting different entities (i.e., performing variable instantiations) depending on the task. This could make reasoning systems more adaptable and task-specific.

All the above notions of structures are also meaningful structural objects to extend our methods with. A more comprehensive treatment of the broader notions of structures would require integrating toolkits from causal machine learning, general Bayesian learning, and neuro-symbolic AI.

**Destructure**

The concept of destructure introduced in this thesis, with active forgetting as one potential implementation, is not without limitations. While active forgetting specifically targets the embeddings, actively removing the structures captured in them, other components of the model could also be considered for forgetting. However, we have chosen embeddings as the primary target for this process, leaving other model components unexplored. One key limitation is the lack of automatic selection for which component should undergo forgetting during the pretraining process. Although it would be more convenient for users if such a mechanism were automated, this would add significant overhead to an already computationally expensive pretraining phase. Additionally, the idea of automating the schedule of forgetting frequency rather than treating it as a hyperparameter introduces further complexity that may increase the computational burden.

It is also worthwhile to compare with techniques such as dropout [Baldi and Sadowski, 2013, Srivastava et al., 2014] and iterative pruning [Frankle and Carbin, 2019]. These methods periodically erase weights, providing regularization and helping to prevent overfitting. However, they are not designed to specifically address generalization to unseen environments, which is a key goal of our proposed destructuring method. Theoretically, active forgetting and similar techniques could be linked to frameworks like Invariant Risk Minimization (IRM) [Arjovsky et al., 2019], which aims to reduce risks across different environments and improve generalization to unseen data points. However, further investigation is required to fully establish this connection.

Another limitation arises from the lack of study on one-time destructuring methods, which focus on removing unwanted structures in the models, directly patching problematic model behaviors. These methods, such as DPR [Karpukhin et al., 2020], RAG [Lewis et al., 2020b], model editing [Meng et al., 2022], and model unlearning [Liu et al., 2025], address specific issues like hallucinations or toxicity in LLM generations. However, they lack the ability to systematically and globally address these issues across the model in an integrated way. Instead, they rely on external interventions, which may not lead to the same depth of control over the model's learning and forgetting processes as the proposed approaches in this thesis. That said, these external methods are reactive and easy to deploy on-the-fly.

### 6.2.2 Methodological Constraints

While this thesis attempts to bridge structured and unstructured learning paradigms, it centres on the embedding and its role in caching symbolic relationships. The unifi-cation is limited to reframing neural embedding optimization as structural operations (message-passing over graphs). This unification is implicit rather than explicit. The alternative direction can be to estabish a mathematically rigorous framework or an empirical system that directly mixes both structured and unstructured inputs synthesizing neural networks and symbolic reasoning [Colelough and Regli, 2024]. For example, directly injecting structured data into the unstructured paradigm, making language models structure-aware [Li et al., 2023, Wu et al., 2024].

### 6.2.3 Evaluation, Scalability, and Computation Efficiency

In the destructuring experiments presented in Part II, we focused on evaluating the model's performance on unseen entities in knowledge graphs and unseen languages for pretrained language models. However, in real-world applications, there are many other potentially unseen scenarios that need to be considered. For pretrained language models, there exists a wide spectrum of linguistic shifts to which the model must adapt. These shifts include domain shifts [Gururangan et al., 2020], temporal evolution [Liska et al., 2022], task/tool changes [Lu et al., 2024], and personalization for different users [Kirk et al., 2024]. All of these factors are important to test when applying active forgetting techniques to ensure that they are effective across a variety of scenarios.

Although latent structures in large language models were analysed in Chapter 3, extracting transparent higher-order n-grams remains a partially unsolved challenge. This issue requires further scaling up of our methods to handle more complex structures. Similarly, the analysis of computational paths can be extended to explore the cascading effects across multiple self-attention modules, a task that demands increased computational resources. We focused on extracting n-grams for a selected set of large language models. To gain a deeper understanding of more models and their internal knowledge, we must systematically examine a broader range of models. This would include potentially verifying the extracted structures against data distributions or real-world knowledge graphs to validate the models' generalization capabilities and alignment with external knowledge.

## 6.3 Significance and Implications

This thesis carries several important implications that extend beyond its theoretical contributions, shedding light on practical applications for both artificial and human cognitive systems.

### 6.3.1 Applications for Machine Minds

By bridging the structured and unstructured paradigms, this work enables cross-paradigm learning, allowing us to borrow strengths from both approaches.

One key application lies in improving the *controllability* of large language models (LLMs). For instance, by training models on relevant n-gram paths identified through our method in Chapter 3, we might be able to enhance their functional flexibility. This can be particularly useful in managing *tooling plasticity* [Lu et al., 2024], where tools can be interpreted as "neural circuits" that activate only under specific conditions. By identifying model paths related to particular tool usages, we could apply active forgetting from Chapter 5 to adjust or reset those paths as needed. Additionally, this work offers pathways to improve the *interpretability* and transparency of LLMs. As shown in Chapter 3, our n-gram interpretability requires only CPU-based post-training processing and no curated external datasets, making it more computationally efficient than other approaches. This method allows institutional actors to systematically audit LLMs, enhancing transparency and user trust in generative AI applications.

### 6.3.2 Applications for Human Minds

This thesis also implies a broader scientific inquiry into how humans build knowledge engines in their minds.

First, the research on embeddings and their function as "memory banks" during training (Chapter 4) provides insights into how human memory might work and be regulated, potentially linking to *engram cells* in neuroscience [Tonegawa et al., 2015, 2018, Ryan and Frankland, 2022, Guskjolen and Cembrowski, 2023]. Second, *active forgetting* techniques that help pretrained language models generalize to new languages with less data (Chapter 5) could inspire new research in language acquisition. This is especially relevant to studying phenomena such as the *critical period* for language learning [Constantinescu et al., 2024]. Third, this thesis highlights potential applications for *digital intervention*. Many recommender systems and social media platforms maintain persistent embeddings for individual users. While these technologies have become integral to daily life, they often accumulate implicit user preferences in embeddings, leading to issues like "brain rot" addiction and echo chamber. If platforms periodically reset user embeddings, we might be able to mitigate digital addiction, and promote healthier online interactions.

# Chapter 7

# Looking Forward

This thesis opens several promising avenues for future research and applications:

## 7.1 Future Directions

Theoretically, this thesis emphasizes the central role of factorization – a form of computation decomposition – in learning structures (Chapter 2 and Chapter 3). A promising direction for future research is studying a unified framework for understanding the widespread presence of factorization in handling discrete symbolic interactions. This includes models such as word2vec, tensor factorization, RNNs, and Transformers. Potentially transformers themselves can be interpreted as bi-level factorization models, offering new perspectives on their internal computations. Exploring this perspective could allow us to interpolate between existing architectures while opening opportunities for designing entirely new ones. A closely related theoretical question concerns the plasticity of architectures – that is, the extent to which their behavior can be systematically controlled. Ultimately, the next generation of AI-based knowledge engines should not only be powerful and adaptive, but also controllable by individuals, ensuring human agency in their use and development.

Beyond theoretical exploration, the most compelling directions lie in applications to AI transparency and safety. First, the identified n-gram paths (Chapter 3) provide a structural handle for controlling the behavior of large language models. For instance, applying invariant learning to selected paths [Arjovsky et al., 2019] may help eliminate unwanted structures that introduce biases or undesirable outcomes. Second, research such as [Chen et al., 2024] can be extended by systematically extracting interpretable structures from different LLMs, advancing both transparency and trust in these systems. Third, the active forgetting techniques proposed in this thesis (Part II) can support the preservation of low-resource languages such as Hokkien. This direction aligns with decentralized knowledge management, where decoupling embeddings from model bodies Zhao et al. [2024b], Iacob et al. [2024] enables cultural preservation and inclusivity. Moreover, such techniques could improve biomedical applications such as epitope link prediction, where limited data hampers generalization across epitope groups [Liu et al., 2024a].

These future directions lie in the intersection between theoretical advances and practical implementations, contributing to both the foundational understanding of AI models and their real-world applications in transparent, safe, and inclusive systems.

## 7.2 Final Thoughts

The equation for building general knowledge engines likely transcends the simplistic notion of *structured + unstructured = intelligence* or scaling both of them. Structured and unstructured representations are merely two states of the engine. They are not the driving forces behind intelligent behaviour.

A more accurate formulation may be:

$$\text{Structuring} \leftrightarrow \text{Destructuring} = \text{Intelligence}$$

where intelligence emerges from balancing the dual forces while maintaining fluidity. Structuring accumulates knowledge by organizing meaningful relations into reusable forms, while destructuring mitigates rigid, outdated, and potentially harmful structures, enabling continuous learning and adaptation in ever-changing reality.

Scaling may indeed be a pathway toward *Artificial General Intelligence* (AGI v1), as digesting vast amounts of data naturally leads to learning more and more complex structures about our world. However, achieving *Artificial Good Intelligence* (AGI v2) requires more than just scaling. It demands plasticity – the ability to discard outdated knowledge structures, regrow new ones, and adapt effectively to new environments. In this view, balancing structuring and destructuring becomes a key hallmark of a truly intelligent system, capable of evolving with the ever-changing realities.

# Appendix

# Appendix A

# Relation Prediction

Here we provide additional technical details and results for *Chapter 2 Language Modelling Completes Knowledge Graph Structures*.

## A.1 Technical Details

### A.1.1 Code Snippets of Relation Prediction

Figure A.1 demonstrates how to add relation prediction to the existing implementation of ComplEx, which transform the existing 1vsAll objective into a language modeling objective.

### A.1.2 Hyperparameters

Tesla P100 and Tesla V100 GPUs were used to run the experiments. We implemented each model by PyTorch. Our codebase is based on Our codebase is based on this repository.

**Relation Prediction Hyperparameter Ranges Across Datasets**

**Kinship, Nations, and UMLS** For all small datasets (Kinship, Nations, UMLS), we trained RESCAL, ComplEx, CP and TuckER with Adagrad optimiser and N3 regularisation for at most 400 epochs. Reciprocal triples were included since they are reported to be helpful [Dettmers et al., 2018, Lacroix et al., 2018]. We performed grid searches over

```python
1   class ComplEx(KBCModel):
2       def __init__(self, sizes, rank, init_size):
3           super(ComplEx, self).__init__()
4           self.sizes = sizes
5           self.rank = rank
6
7           self.embeddings = nn.ModuleList([
8               nn.Embedding(s, 2 * rank, sparse=False)
9               for s in sizes[:2]
10          ])
11          self.embeddings[0].weight.data *= init_size
12          self.embeddings[1].weight.data *= init_size
13
14      def forward(self, x, score_rhs=True, score_rel=False, score_lhs=False, normalize_rel=False):
15          lhs = self.embeddings[0](x[:, 0])
16          rel = self.embeddings[1](x[:, 1])
17          rhs = self.embeddings[0](x[:, 2])
18
19          lhs = lhs[:, :self.rank], lhs[:, self.rank:]
20          rel = rel[:, :self.rank], rel[:, self.rank:]
21          rhs = rhs[:, :self.rank], rhs[:, self.rank:]
22
23          rhs_scores, rel_scores = None, None
24          if score_rhs:
25              to_score_entity = self.embeddings[0].weight
26              to_score_entity = to_score_entity[:, :self.rank], to_score_entity[:, self.rank:]
27              rhs_scores = (
28                  (lhs[0] * rel[0] - lhs[1] * rel[1]) @ to_score_entity[0].transpose(0, 1) +
29                  (lhs[0] * rel[1] + lhs[1] * rel[0]) @ to_score_entity[1].transpose(0, 1)
30              )
31          if score_rel:
32              to_score_rel = self.embeddings[1].weight
33              to_score_rel = to_score_rel[:, :self.rank], to_score_rel[:, self.rank:]
34              rel_scores = (
35                  (lhs[0] * rhs[0] + lhs[1] * rhs[1]) @ to_score_rel[0].transpose(0, 1) +
36                  (lhs[0] * rhs[1] - lhs[1] * rhs[0]) @ to_score_rel[1].transpose(0, 1)
37              )
38          if score_lhs:
39              to_score_lhs = self.embeddings[0].weight
40              to_score_lhs = to_score_lhs[:, :self.rank], to_score_lhs[:, self.rank:]
41              lhs_scores = (
42                  (rel[0] * rhs[0] + rel[1] * rhs[1]) @ to_score_lhs[0].transpose(0, 1) +
43                  (rel[0] * rhs[1] - rel[1] * rhs[0]) @ to_score_lhs[1].transpose(0, 1)
44              )
```

Figure A.1: Relation Prediction for ComplEx, the red region shows the lines related to using relation prediction as an auxiliary training task.

Table A.1: Hyperparameter search space for different KBC models on small datasets (Kinship, Nations, UMLS). $d$ is embedding size, $d_r$ is relation embedding size, *lr* is learning rate, *bsz* is batch size, and *reg* is regularization.

| Model | $d$ or $(d, d_r)$ | lr | bsz | reg |
|---|---|---|---|---|
| RESCAL | 50, 100, 200 | 1e−1, 1e−2 | 10, 50, 100, 500 | 0, 5e−3, 1e−2, 5e−2, 1e−1, 5e−1 |
| ComplEx | 100, 200, 500, 1000 | 1e−1, 1e−2 | 10, 50, 100, 500 | 0, 5e−3, 1e−2, 5e−2, 1e−1, 5e−1 |
| CP | 200, 400, 1000, 2000 | 1e−1, 1e−2 | 10, 50, 100, 500 | 0, 5e−3, 1e−2, 5e−2, 1e−1, 5e−1 |
| TuckER | (100, 25), (200, 25), (100, 50), (200, 50), (100, 100), (200, 100) | 1e−1, 1e−2 | 10, 50, 100, 500 | 0, 5e−3, 1e−2, 5e−2, 1e−1, 5e−1 |

hyperparameter combinations and chose the best configuration for each dataset based on validation MRR. We list the grids of hyperparameter search in Table A.1 and report the best configuration in Table A.2. As for balancing between relation prediction and entity prediction, we searched the weight of relation prediction $\lambda$ over $\{4, 2, 0.5, 0.25, 0.125\}$.

**WN18RR, FB15k-237, and Aristo-v4**  For all datasets, we trained ComplEx with an N3 regulariser and Adagrad optimiser for at most 400 epochs. Reciprocal triples were included since they are reported to be helpful [Dettmers et al., 2018, Lacroix et al., 2018]. As for the weight of relation prediction, we searched over different zones for different datasets. For WN18RR, we searched the weight of relation prediction over $\{5e−3, 1e−3, 5e−2, 1e−1, 5e−1, 1\}$. For FB15k-237 and Aristo-v4, we searched over $\{0.125, 0.25, 0.5, 1, 2, 4\}$. We did grid searches over hyperparameter combinations and chose the best configuration for each dataset based on validation MRR. We report the grids for each dataset in Table A.3, and the best configuration in Table A.4.

**Relation Prediction Hyperparameter Ranges Across Models**

We experiment with each model on FB15k-237. Note that the original TucKER [Balazevic et al., 2019] includes some training strategies which are not used in CP, ComplEx, and TuckER, like dropout, learning rate decay etc. However, for a fair comparison of Table A.2: Best hyperparameter configurations and validation MRRs on small datasets. RP = relation prediction, EP = entity prediction, $d$ = embedding size, $d_r$ = relation embedding size, $lr$ = learning rate, $bsz$ = batch size, $reg$ = regularization, $\lambda$ = weight on RP. NA = not applicable.

| Dataset | RP | EP | Model | $d$ / $(d, d_r)$ | $lr$ | $bsz$ | $reg$ | $\lambda$ | Dev MRR |
|---|---|---|---|---|---|---|---|---|---|
| KINSHIP | ✔ | ✘ | TuckER | (200, 100) | 0.10 | 10 | 0.10 | NA | 0.920 |
|  | ✘ | ✔ | CP | 2000 | 0.10 | 50 | 0.01 | NA | 0.897 |
|  | ✔ | ✔ | CP | 2000 | 0.10 | 50 | 0.05 | 4.00 | 0.918 |
| NATIONS | ✔ | ✘ | TuckER | (200, 50) | 0.01 | 10 | 0.10 | NA | 0.686 |
|  | ✘ | ✔ | CP | 2000 | 0.01 | 10 | 0.01 | NA | 0.855 |
|  | ✔ | ✔ | TuckER | (200, 25) | 0.01 | 10 | 0.10 | 0.25 | 0.865 |
| UMLS | ✔ | ✘ | CP | 1000 | 0.10 | 500 | 0.01 | NA | 0.863 |
|  | ✘ | ✔ | ComplEx | 1000 | 0.10 | 10 | 0.00 | NA | 0.968 |
|  | ✔ | ✔ | ComplEx | 1000 | 0.01 | 10 | 0.00 | 0.50 | **0.972** |

how relation prediction affects each model, we trained all the models conditioned on similar settings with Adagrad optimiser and N3 regularisation for at most 400 epochs. We performed grid searches and selected the best hyperparameter configurations according to validation MRR. We set the weight of relation prediction to 1 in this experiment. Table A.5 lists the grid of the shared hyperparameters. For RESCAL, the regularisation over predicate matrices can be normalised over the rank to achieve better results. Also, F2 regularisation empirically performed better than the N3 regulariser for RESCAL. For TuckER, the ranks for predicate and entity are different. Table A.6 lists the best hyperparameter configuration found by our search.

Table A.3: Hyperparameter search space for vanilla relation perturbation using ComplEx on three datasets. $d$ = embedding size, $lr$ = learning rate, $bsz$ = batch size, $reg$ = regularization strength.

| Dataset | d | lr | bsz | reg |
|---|---|---|---|---|
| WN18RR | 100, 500, 1000 | 1e−1, 1e−2 | 100, 500, 1000 | 5e−3, 1e−2, 5e−2, 1e−1, 5e−1, 1 |
| FB15k-237 | 100, 500, 1000 | 1e−1, 1e−2 | 100, 500, 1000 | 5e−4, 5e−3, 1e−2, 5e−2, 1e−1, 5e−1, 1, 0 |
| Aristo-v4 | 500, 1000, 1500 | 1e−1, 1e−2 | 100, 500, 1000 | 0, 5e−3, 1e−2, 5e−2, 1e−1, 5e−1, 1 |

## A.2   Additional Results

### Additional Metrics for Ablation on Rank

Figure A.2 (MRR), Figure A.3 (Hits@3) and Figure A.4 (Hits@10) show additional metrics for the experiments ablating ranks. The range of the rank is

$$\{25, 50, 100, 500, 1K, 2K, 3K, 4K\}.$$

Blue indicates training with relation prediction, while red indicates training without prediction.

Table A.4: Best hyperparameter configurations and the corresponding validation MRR for ComplEx across datasets with Weighted Relation Perturbation. RP = relation prediction, EP = entity prediction, $d$ = embedding size, $lr$ = learning rate, $bsz$ = batch size, $reg$ = regularisation strength, $\lambda$ = relation weighting. NA = not applicable.

| Dataset | RP | EP | $d$ | lr | bsz | reg | $\lambda$ | Dev MRR |
|---|---|---|---|---|---|---|---|---|
| | ✔ | ✘ | 1000 | 0.10 | 500 | 0.5 | NA | 0.2579 |
| WN18RR | ✘ | ✔ | 1000 | 0.10 | 100 | 0.10 | NA | 0.4881 |
| | ✔ | ✔ | 1000 | 0.10 | 100 | 0.10 | 0.050 | 0.4901 |
| | ✔ | ✘ | 1000 | 0.10 | 1000 | 0.0005 | NA | 0.2629 |
| FB15k-237 | ✘ | ✔ | 1000 | 0.10 | 100 | 0.05 | NA | 0.3723 |
| | ✔ | ✔ | 1000 | 0.10 | 1000 | 0.05 | 4.000 | 0.3937 |
| | ✔ | ✘ | 1500 | 0.10 | 1000 | 0.01 | NA | 0.1687 |
| Aristo-v4 | ✘ | ✔ | 1500 | 0.01 | 500 | 0.01 | NA | 0.3071 |
| | ✔ | ✔ | 1500 | 0.10 | 100 | 0.05 | 0.125 | 0.3144 |

Table A.5: Hyperparameter search for different KBC models on FB15k-237. $d$ stands for embedding size, $d_r$ for relation embedding size, $lr$ is learning rate, $bsz$ is batch size, and *reg* is regularization strength.

| Model | *d* or (*d*, $d_r$) | *lr* | *bsz* | *reg* |
|---|---|---|---|---|
| RESCAL | 128, 256, 512 | 1e−1, 1e−2 | 100, 500, 1000 | 0, 1e−3, 5e−3, 1e−2, 5e−2, 1e−1, 5e−1, 1 |
| ComplEx | 100, 500, 1000 | 1e−1, 1e−2 | 100, 500, 1000 | 0, 5e−4, 5e−3, 1e−2, 5e−2, 1e−1, 5e−1, 1 |
| CP | 64, 128, 256, 512, 4000 | 1e−1, 1e−2 | 100, 500, 1000 | 5e−3, 1e−2, 5e−2, 1e−1, 5e−1, 1 |
| TuckER | (1000, 150), (1000, 100), (400, 400), (500, 75), (300, 300), (200, 200) | 1e−1, 1e−2 | 100, 500, 1000 | 5e−3, 1e−2, 5e−2, 1e−1, 5e−1, 1 |

Table A.6: Best hyperparameter configurations and validation MRR on FB15k-237 across models. RP = Relation Prediction. $d$ is the embedding size, $d_r$ is the relation embedding size, $lr$ is the learning rate, $bsz$ is the batch size, and $reg$ is the regularisation strength.

| Model | RP | $d$ or $(d, d_r)$ | $lr$ | $bsz$ | $reg$ | Dev MRR |
|---|---|---|---|---|---|---|
| RESCAL | ✗ | 512 | 0.1 | 500 | 0.00 | 0.365 |
|  | ✓ | 512 | 0.1 | 100 | 0.00 | 0.367 |
| ComplEx | ✗ | 1000 | 0.1 | 100 | 0.05 | 0.372 |
|  | ✓ | 1000 | 0.1 | 1000 | 0.05 | 0.387 |
| CP | ✗ | 4000 | 0.1 | 100 | 0.05 | 0.364 |
|  | ✓ | 4000 | 0.1 | 1000 | 0.05 | 0.372 |
| TuckER | ✗ | (1000, 100) | 0.1 | 100 | 0.10 | 0.359 |
|  | ✓ | (1000, 100) | 0.1 | 100 | 0.50 | 0.360 |

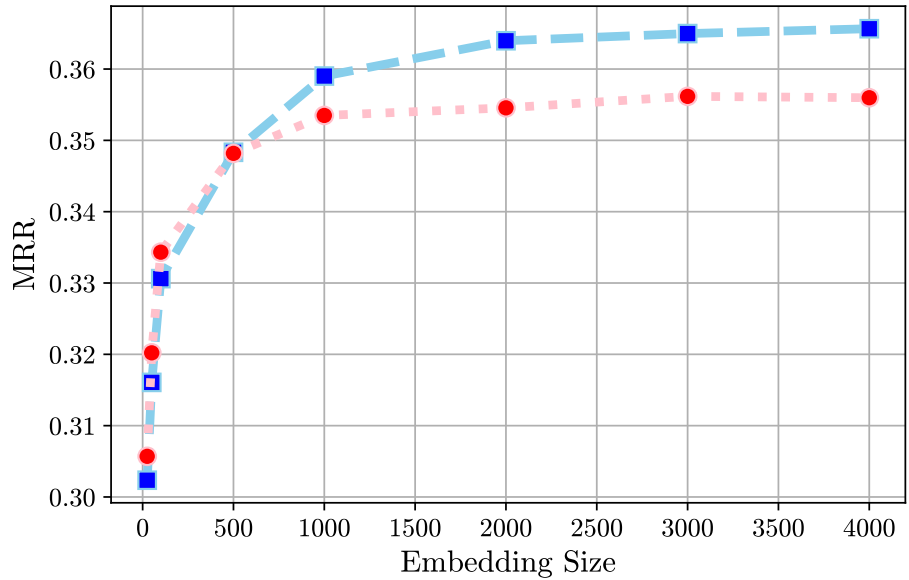

Figure A.2: MRR versus Rank for CP on FB15k-237.

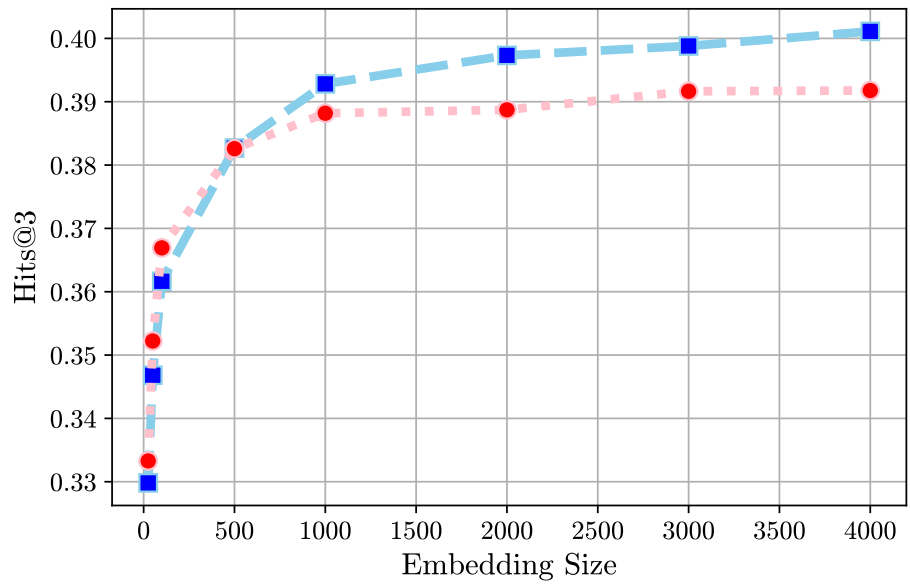

Figure A.3: Hits@3 versus Rank for CP on FB15k-237.

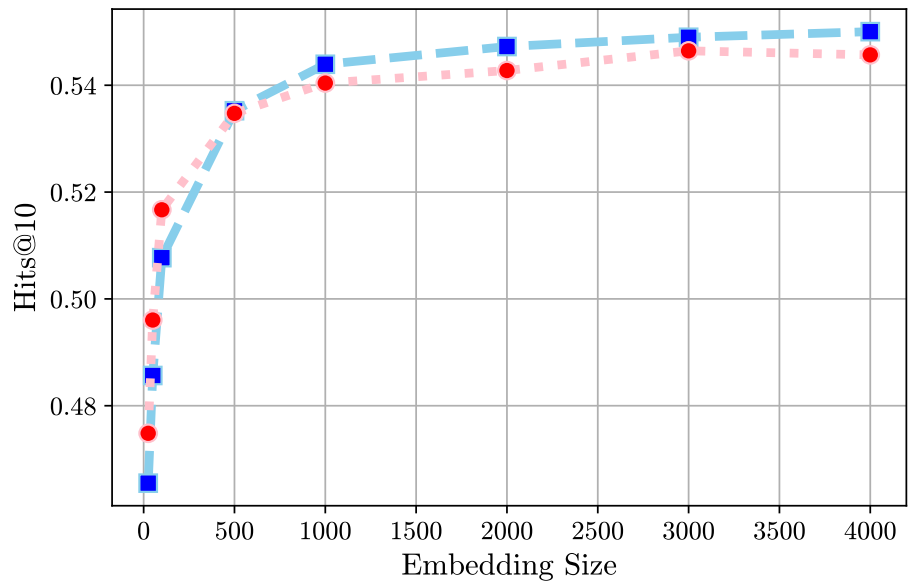

Figure A.4: Hits@10 versus Rank for CP on FB15k-237.

Table A.7: Top 20 $(s, p, o)$ test triples, based on their increase of the right-hand-side (i.e. on the task of predicting $o$ given $p$ and $s$) reciprocal rank after we introduce the relation prediction auxiliary objective.

| Subject | Predicate | Object | $\Delta$ 1/**Rank** |
|---|---|---|---|
| Paramount Pictures | /common/topic/webpage /common/webpage/category | NA | 1.000 |
| Midfielder | /sports/sports_position/players /sports/sports_team_roster/team | Gaziantepspor | 0.988 |
| The Dictator (2012 film) | /film/film/personal_appearances /film/personal_film_appearance/person | Hillary Clinton | 0.970 |
| Academy Award for Best Supporting Actress | /award/award_category/winners /award/award_honor/award_winner | Maureen Stapleton | 0.963 |
| Christopher Columbus | /user/tsegaran/random/taxonomy_subject/entry /user/tsegaran/random/taxonomy_entry/taxonomy | Library of Congress Classification | 0.950 |
| United Kingdom | /location/location/contains | Westminster | 0.933 |
| President | /organization/role/leaders /organization/leadership/organization | West Virginia University | 0.923 |
| Academy Award for Best Supporting Actor | /award/award_category/winners /award/award_honor/award_winner | Christopher Walken | 0.923 |
| President | /organization/role/leaders /organization/leadership/organization | Bryn Mawr College | 0.917 |
| President | /organization/role/leaders /organization/leadership/organization | Dickinson College | 0.917 |
| National Society of Film Critics Award for Best Actress | /award/award_category/winners /award/award_honor/award_winner | Reese Witherspoon | 0.917 |
| President | /organization/role/leaders /organization/leadership/organization | Louisiana State University | 0.917 |
| Vera Drake | /award/award_winning_work/awards_won /award/award_honor/award | Los Angeles Film Critics Association Award for Best Actress | 0.900 |
| President | /organization/role/leaders /organization/leadership/organization | University of Oklahoma | 0.900 |
| United States | /location/country/second_level_divisions | Marion County, Indiana | 0.900 |
| President | /organization/role/leaders /organization/leadership/organization | University of Southern California | 0.900 |
| Travis Tritt | /film/actor/film /film/performance/film | Blues Brothers 2000 | 0.900 |
| London Film Critics' Circle Award for Director of the Year | /award/award_category/winners /award/award_honor/award_winner | Neil Jordan | 0.900 |
| United States | /location/country/second_level_divisions | Niagara County, New York | 0.900 |
| Deion Sanders | /people/person/places_lived /people/place_lived/location | Atlanta | 0.900 |

Table A.8: Top 20 $(s, p, o)$ test triples, based on their increase of the left-hand-side (i.e. on the task of predicting $s$ given $p$ and $o$) reciprocal rank after we introduce the relation prediction auxiliary objective.

| Subject | Predicate | Object | $\Delta$ 1/**Rank** |
|---|---|---|---|
| Midfielder | /soccer/football_team/current_roster /soccer/football_roster_position/position | Wimbledon F.C. | 0.990 |
| Forward (association football) | /soccer/football_team/current_roster /sports/sports_team_roster/position | Iraq national football team | 0.989 |
| United States | /film/film/release_date_s /film/film_regional_release_date/film_release_region | Cleopatra (1963 film) | 0.975 |
| Critics' Choice Movie Award for Best Acting Ensemble | /award/award_nominee/award_nominations /award/award_nomination/award | Matt Damon | 0.975 |
| Female | /people/person/gender | Grey DeLisle | 0.968 |
| United Kingdom | /film/film/release_date_s /film/film_regional_release_date/film_release_region | New Year's Eve (2011 film) | 0.967 |
| United States dollar | /location/statistical_region/rent50_2 /measurement_unit/dated_money_value/currency | Anchorage, Alaska | 0.966 |
| United Kingdom | /film/film/release_date_s /film/film_regional_release_date/film_release_region | Killing Them Softly | 0.962 |
| Glendale, California | /people/person/spouse_s /people/marriage/location_of_ceremony | Jane Wyman | 0.962 |
| United Kingdom | /film/film/release_date_s /film/film_regional_release_date/film_release_region | ParaNorman | 0.962 |
| Streaming media | /film/film/distributors /film/film_film_distributor_relationship/film_distribution_medium | Pulp Fiction | 0.960 |
| United Kingdom | /film/film/release_date_s /film/film_regional_release_date/film_release_region | Magic Mike | 0.958 |
| United States dollar | /location/statistical_region/rent50_2 /measurement_unit/dated_money_value/currency | Napa County, California | 0.952 |
| United Kingdom | /film/film/release_date_s /film/film_regional_release_date/film_release_region | Rock of Ages (2012 film) | 0.952 |
| United Kingdom | /film/film/release_date_s /film/film_regional_release_date/film_release_region | Contact (1997 American film) | 0.950 |
| Streaming media | /film/film/distributors /film/film_film_distributor_relationship/film_distribution_medium | American History X | 0.950 |
| Los Angeles | /organization/organization/place_founded | Paramount Pictures | 0.947 |
| United Kingdom | /film/film/release_date_s /film/film_regional_release_date/film_release_region | L.A. Confidential (film) | 0.941 |
| Psychological thriller | /film/film/genre | Family Plot | 0.941 |
| United Kingdom | /film/film/release_date_s /film/film_regional_release_date/film_release_region | Moneyball (film) | 0.938 |

# Appendix B

# ReFactor GNNs

Here we provide additional technical details and results for *Chapter 4 Inductive Knowledge Graph Learning with Active Forgetting*.

## B.1 Technical Details

### B.1.1 Proof, Complexity, and Expressiveness

**Theorem 1 Proof**

In this section, we prove Theorem 1, which we restate here for convenience.

**Theorem B.1.1** (Message passing in FMs). *The gradient descent operator* GD *(Equation 4.10) on the node embeddings of a DistMult model (Equation 4.4) with the maximum likelihood objective in Equation 4.3 and a multi-relational graph $\mathcal{T}$ defined over entities $\mathcal{E}$ induces a message-passing operator whose composing functions are:*

$$q_{\text{M}}(\phi[v], r, \phi[w]) = \begin{cases} \phi[w] \odot g(r) & \text{if } (r,w) \in \mathcal{N}_{+}^{1}[v], \\ (1 - P_\theta(v|w,r))\phi[w] \odot g(r) & \text{if } (r,w) \in \mathcal{N}_{-}^{1}[v]; \end{cases} \quad \text{(B.1)}$$

$$q_{\text{A}}(\{m[v,r,w] : (r,w) \in \mathcal{N}^1[v]\}) = \sum_{(r,w)\in\mathcal{N}^1[v]} m[v,r,w]; \quad \text{(B.2)}$$

$$q_{\text{U}}(\phi[v], z[v]) = \phi[v] + \alpha z[v] - \beta n[v], \quad \text{(B.3)}$$

*where, defining the sets of triplets $\mathcal{T}^{-v} = \{(s, r, o) \in \mathcal{T} \ : \ s \neq v \land o \neq v\}$,*

$$n[v] = \frac{|\mathcal{N}^1_+[v]|}{|\mathcal{T}|}\mathbb{E}_{P_{\mathcal{N}^1_+[v]}}\mathbb{E}_{u \sim P_\theta(\cdot|v,r)}g(r) \odot \phi[u] + \frac{|\mathcal{T}^{-v}|}{|\mathcal{T}|}\mathbb{E}_{P_{\mathcal{T}^{-v}}}P_\theta(v|s,r)g(r) \odot \phi[s], \quad \text{(B.4)}$$

*where $P_{\mathcal{N}^1_+[v]}$ and $P_{\mathcal{T}^{-v}}$ are the empirical probability distributions associated to the respective sets.*

*Proof.* Remember that we assume that there are no triplets where the source and the target node are the same (i.e. $(v, r, v)$, with $v \in \mathcal{E}$ and $r \in \mathcal{R}$), and let $v \in \mathcal{E}$ be a node in $\mathcal{E}$. First, let us consider the gradient descent operator GD over $v$'s node embedding $\phi[v]$:

$$\text{GD}(\phi, \mathcal{T})[v] = \phi[v] + \alpha \sum_{(\text{v},\text{r},\text{w}) \in \mathcal{T}} \frac{\partial \log P(\text{w} \,|\, \text{v}, \text{r})}{\partial \phi[v]}.$$

The gradient is a sum over components associated with the triplets $(\text{v}, \text{r}, \text{w}) \in \mathcal{T}$; based on whether the corresponding triplet involves $v$ in the subject or object position, or does not involve $v$ at all, these components can be grouped into three categories:

1. Components corresponding to the triplets where $\text{v} = v \land \text{w} \neq v$. The sum of these components is given by:

$$\sum_{(v,\text{r},\text{w}) \in \mathcal{T}} \frac{\partial \log P(\text{w}\,|v,\text{r})}{\partial \phi[v]} = \sum_{(v,\text{r},\text{w}) \in \mathcal{T}} \left[ \frac{\partial \Gamma(v, \text{r}, \text{w})}{\partial \phi[v]} - \sum_u P(u|v,\text{r}) \frac{\partial \Gamma(v, \text{r}, u)}{\partial \phi[v]} \right]$$

$$= \sum_{(\text{r},\text{w}) \in \mathcal{N}^1_+[v]} \phi[\text{w}] \odot g(\text{r}) \ - \ \sum_{(v,\text{r},\text{w}) \in \mathcal{T}} \sum_u P(u|v,\text{r})g(\text{r}) \odot \phi[u].$$

2. Components corresponding to the triplets where $\text{v} \neq v \land \text{w} = v$. The sum of these components is given by:

$$\sum_{(\text{v},\text{r},v) \in \mathcal{T}} \frac{\partial \log P(v\,|\,\text{v}, \text{r})}{\partial \phi[v]} = \sum_{(\text{v},\text{r},v) \in \mathcal{T}} \left[ \frac{\partial \Gamma(\text{v}, \text{r}, v)}{\partial \phi[v]} - \sum_u P(u\,|\,\text{v}, \text{r}) \frac{\partial \Gamma(\text{v}, \text{r}, u)}{\partial \phi[v]} \right]$$

$$= \sum_{(\text{v},\text{r}) \in \mathcal{N}^1_-[v]} g(\text{r}) \odot \phi[\text{v}] \left(1 - P(v\,|\,\text{v}, \text{r})\right).$$

3. Components corresponding to the triplets where $\text{v} \neq v \land \text{w} \neq v$. The sum of these

components is given by:

$$\sum_{(v,r,w)\in\mathcal{T}} \frac{\partial \log P(\mathbf{w}\,|\,\mathbf{v},\mathbf{r})}{\partial \phi[v]} = \sum_{(v,r,w)\in\mathcal{T}} \left[ 0 - \sum_u P(u|\,\mathbf{v},\mathbf{r})\frac{\partial \Gamma(\mathbf{v},\mathbf{r},u)}{\partial \phi[v]} \right]$$

$$= \sum_{(v,r,w)\in\mathcal{T}} -P(v|\,\mathbf{v},\mathbf{r})\frac{\partial \Gamma(\mathbf{v},\mathbf{r},v)}{\partial \phi[v]}.$$

$$= \sum_{(v,r,w)\in\mathcal{T}} -P(v|\,\mathbf{v},\mathbf{r})g(\mathbf{r})\odot\phi[\mathbf{v}].$$

Collecting these three categories, the GD operator over $\phi[v]$, or rather the node representation update in DistMult, can be rewritten as:

$$\text{GD}(\phi,\mathcal{T})[v] \tag{B.5}$$

$$= \phi[v] + \alpha \underbrace{\sum_{\{(r,w)\in\mathcal{N}^1_+[v]\}} \phi[\mathbf{w}]\odot g(\mathbf{r}) + \sum_{(r,v)\in\mathcal{N}^1_-[v]} \phi[\mathbf{v}]\odot g(\mathbf{r})\,(1-P(v|\,\mathbf{v},\mathbf{r}))}_{v\text{'s neighbourhood}\to v} \tag{B.6}$$

$$-\alpha \underbrace{\sum_{(v,r,w)\in\mathcal{T}, v\neq v, w\neq v} P(v|\,\mathbf{v},\mathbf{r})g(\mathbf{r})\odot\phi[\mathbf{v}] - \alpha \sum_{(v,r,w)\in\mathcal{T}} \sum_u P(u|v,\mathbf{r})g(\mathbf{r})\odot\phi[u]}_{\text{beyond neighbourhood}\to v}. \tag{B.7}$$

Note that the component "$v$'s neighbourhood $\to v$" (highlighted in red) in Equation B.5 is a sum over $v$'s neighbourhood – gathering information from positive neighbours $\phi[\mathbf{w}], (\cdot,\mathbf{w})\in\mathcal{N}^1_+[v]$ and negative neighbours $\phi[\mathbf{v}], (\cdot,\mathbf{v})\in\mathcal{N}^1_-[v]$. Hence, each atomic term of the sum can be seen as a message vector between $v$ and $v$'s neighbouring node. Formally, letting $w$ be $v$'s neighbouring node, the message vector can be written as follows

$$m[v,r,w] = q_\text{M}(\phi[v],r,\phi[w]) = \begin{cases} \phi[w]\odot g(r),\text{ if }(r,w)\in\mathcal{N}^1_+[v], \\ \phi[w]\odot g(r)(1-P(v|w,r)),\text{ if }(r,w)\in\mathcal{N}^1_-[v], \end{cases} \tag{B.8}$$

which induces a bidirectional message function $q_M$. On the other hand, the summation over these atomic terms (message vectors) induces the aggregate function $q_A$:

$$\begin{aligned} z[v] &= q_A(\{m[v,r,w] \,:\, (r,w) \in \mathcal{N}^1[v]\}) \\ &= \sum_{(r,w) \in \mathcal{N}^1_+[v]} m^l[v,\mathbf{r},\mathbf{w}] + \sum_{(r,v) \in \mathcal{N}^1_-[v]} m^l[\mathbf{v},\mathbf{r},v] = \sum_{(r,w) \in \mathcal{N}^1[v]} m[v,r,w]. \end{aligned} \quad \text{(B.9)}$$

Finally, the component "beyond neighbourhood $\to v$" (highlighted in blue) is a term that contains dynamic information flow from global nodes to $v$.

If we define

$$\begin{aligned} n[v] &= \frac{1}{|\mathcal{T}|} \sum_{(v,r,w) \in \mathcal{T}} \sum_u P(u|v,\mathbf{r}) g(\mathbf{r}) \odot \phi[u] \\ &+ \frac{1}{|\mathcal{T}|} \sum_{\substack{(v,r,w) \in \mathcal{T} \\ \mathbf{v} \neq v, \mathbf{w} \neq v}} P(v|\mathbf{v},\mathbf{r}) g(\mathbf{r}) \odot \phi[\mathbf{v}], \end{aligned} \quad \text{(B.10)}$$

the GD operator over $\phi[v]$ then boils down to an update function which utilises previous node state $\phi[v]$, aggregated message $z[v]$ and a global term $n[v]$ to produce the new node state:

$$\text{GD}(\phi, \mathcal{T})[v] = q_U(\phi[v], z[v]) = \phi[v] + \alpha z[v] - \beta n[v]. \quad \text{(B.11)}$$

Furthermore, $n[v]$ can be seen as a weighted sum of expectations by recasting the summations over triplets as expectations:

$$\begin{aligned} n[v] &= \frac{|\mathcal{N}^1_+[v]|}{|\mathcal{T}|} \mathbb{E}_{(v,\mathbf{r},\mathbf{w}) \sim P_{\mathcal{N}^1_+[v]}} \mathbb{E}_{u \sim P(\cdot|v,\mathbf{r})} g(\mathbf{r}) \odot \phi[u] \\ &+ \frac{|\mathcal{T}^{-v}|}{|\mathcal{T}|} \mathbb{E}_{(\mathbf{v},\mathbf{r},\mathbf{w}) \sim P_{\mathcal{T}^{-v}}} P(v|\mathbf{v},\mathbf{r},) g(\mathbf{r}) \odot \phi[\mathbf{v}] \end{aligned} \quad \text{(B.12)}$$

where $\mathcal{T}^{-v} = \{(\mathbf{v},\mathbf{r},\mathbf{v}') \in \mathcal{T} \,|\, \mathbf{v} \neq v \wedge \mathbf{v}' \neq v\}$ is the set of triplets that do not contain $v$. $\square$

**Extension to AdaGrad and N3 Regularisation**

State-of-the-art FMs are often trained with training strategies adapted for each model category. For example, using an N3 regulariser [Lacroix et al., 2018] and AdaGrad optimiser [Duchi et al., 2011], which we use for our experiments. For N3 regulariser, we add a gradient term induced by the regularised loss:

$$\frac{\partial L}{\partial \phi[v]} = \frac{\partial L_{\text{fit}}}{\partial \phi[v]} + \lambda \frac{\partial L_{\text{reg}}}{\partial \phi[v]} = \frac{\partial L_{\text{fit}}}{\partial \phi[v]} + \lambda \text{sign}(\phi[v])\phi[v]^2$$

where $L_{\text{fit}}$ is the training loss, $L_{\text{reg}}$ is the regularisation term, $\text{sign}(\cdot)$ is an element-wise sign function, and $\lambda \in \mathbb{R}_+$ is a hyperparameter specifying the regularisation strength. The added component relative to this regulariser fits into the message function $q_{\text{M}}(\phi[v], r, \phi[w])$ as follows:

$$q_{\text{M}}(\cdot) = \begin{cases} \phi[w] \odot g(r) - \lambda \, \text{sign}(\phi[w]) \, \phi[w]^2, & \text{if } (r, w) \in \mathcal{N}_+^1[v], \\ \phi[w] \odot g(r) \, (1 - P(v|w, r)) - \lambda \, \text{sign}(\phi[w]) \, \phi[w]^2, & \text{if } (w, r) \in \mathcal{N}_-^1[v]. \end{cases}$$

(B.13)

Our derivation in Section 4.4 focuses on (stochastic) gradient descent as the optimiser for training FMs. Going beyond this, complex gradient-based optimisers like AdaGrad use running statistics of the gradients. For example, for an AdaGrad optimiser, the gradient is element-wisely re-scaled by $\frac{1}{\sqrt{s_v + \epsilon}} \nabla_{\phi[v]} L$ where $s$ is the running sum of squared gradients and $\epsilon > 0$ is a hyperparameter added to the denominator to improve numerical stability. Such re-scaling can be absorbed into the update equation:

$$\text{AdaGrad}(\phi, \mathcal{T})[v] = \phi[v] + (\alpha z[v] - \beta n[v]) * \frac{1}{\sqrt{s[v] + \epsilon}}.$$

In general, we can interpret any auxiliary variable introduced by the optimiser (e.g. the velocity) as an additional part of the entities and relations representations on which message passing happens. However, the specific equations would depend on the optimiser's dynamics and would be hard to formally generalise.

**Extensions to Other Score Functions e.g. ComplEx**

The two main design choices in Theorem B.1.1 are 1) the score function $\Gamma$, and 2) the optimization dynamics over the node embeddings. We chose DistMult and GD because of their mathematical simplicity, leading to easier-to-read formulas. We can adapt the theorem to general, smooth scoring functions $\Gamma : \mathcal{E} \times \mathcal{R} \times \mathcal{E} \to \mathbf{R}$ by replacing oc-currences of the gradient of DistMult with a generic $\nabla\Gamma$ (the gradient of DistMult w.r.t. $\phi[v]$ at $(v, r, w)$ is simply $g(r) \odot \phi[w]$). This gives us the following lemma:

**Lemma B.1.2** (Message passing in FMs). *The gradient descent operator* GD *(4.10) on the node embeddings of a general score function with the maximum likelihood objective in Equation 4.3 and a multi-relational graph $\mathcal{T}$ defined over entities $\mathcal{E}$ induces a message-passing operator whose composing functions are:*

$$q_{\text{M}}(\phi[v], r, \phi[w]) = \begin{cases} \nabla_{\phi[v]}\Gamma(v, r, w) & \text{if } (r, w) \in \mathcal{N}_+^1[v], \\ (1 - P_\theta(v|w, r))\nabla_{\phi[v]}\Gamma(w, r, v) & \text{if } (r, w) \in \mathcal{N}_-^1[v]; \end{cases} \quad \text{(B.14)}$$

$$q_{\text{A}}(\{m[v, r, w] : (r, w) \in \mathcal{N}^1[v]\}) = \sum_{(r,w) \in \mathcal{N}^1[v]} m[v, r, w]; \quad \text{(B.15)}$$

$$q_{\text{U}}(\phi[v], z[v]) = \phi[v] + \alpha z[v] - \beta n[v], \quad \text{(B.16)}$$

*where, defining the sets of triplets $\mathcal{T}^{-v} = \{(s, r, o) \in \mathcal{T} : s \neq v \wedge o \neq v\}$,*

$$n[v] = \frac{|\mathcal{N}_+^1[v]|}{|\mathcal{T}|} \mathbb{E}_{P_{\mathcal{N}_+^1[v]}} \mathbb{E}_{u \sim P_\theta(\cdot|v,r)} \nabla_{\phi[v]}\Gamma(v, r, u)$$
$$+ \frac{|\mathcal{T}^{-v}|}{|\mathcal{T}|} \mathbb{E}_{P_{\mathcal{T}^{-v}}} P_\theta(v|s, r) \nabla_{\phi[v]}\Gamma(s, r, v) \quad \text{(B.17)}$$

*where $P_{\mathcal{N}_+^1[v]}$ and $P_{\mathcal{T}^{-v}}$ are the empirical probability distributions associated to the respective sets.*

Accordingly, the node representation updating equations in Section 4.4.1 can be rewritten as follows

$$\text{GD}(\phi, \{(v, r, w)\})[v] = \phi[v] + \alpha \left( \underbrace{\nabla_{\phi[v]}\Gamma(v, r, w)}_{w \to v} - \underbrace{\sum_{u \in \mathcal{E}} P_\theta(u|v, r)\nabla_{\phi[v]}\Gamma(v, r, u)}_{u \to v} \right),$$

$$\text{GD}(\phi, \{(v, r, w)\})[w] = \phi[w] + \alpha \underbrace{(1 - P_\theta(w|v, r))\nabla_{\phi[w]}\Gamma(v, r, w)}_{v \to w},$$

$$\text{GD}(\phi, \{(v, r, w)\})[u] = \phi[u] + \alpha \left( \underbrace{-P_\theta(u|v, r)\nabla_{\phi[u]}\Gamma(v, r, u)}_{v \to u} \right).$$

$\nabla_{\phi[\cdot]}\Gamma$ can be different for different models. For example, here we offer a specific derivation for ComplEx [Trouillon et al., 2016]. Let $d = K/2$ be the hidden size for ComplEx. The ComplEx score function is given as follows

$$\begin{aligned}\Gamma(v, r, w) =&< \psi[r]_{(0:d)}, \phi[v]_{(0:d)}, \phi[w]_{(0:d)} > + < \psi[r]_{(0:d)}, \phi[v]_{(d:)}, \phi[w]_{(d:)} > \\ &+ < \psi[r]_{(d:)}, \phi[v]_{(0:d)}, \phi[w]_{(d:)} > - < \psi[r]_{(d:)}, \phi[v]_{(d:)}, \phi[w]_{(0:d)} >\end{aligned} \quad \text{(B.18)}$$

where $(0:d)$ indicates the real part of the complex vector and $(d:)$ indicates the image part of the complex vector. The gradients of the ComplEx score function with respect to the real/image node representations are given by $\frac{\partial\Gamma(v,r,w)}{\partial\phi[v]_{(0:d)}} = \psi[r]_{(0:d)} \odot \phi[w]_{(0:d)} + \psi[r]_{(d:)} \odot \phi[w]_{(d:)}$, $\frac{\partial\Gamma(v,r,w)}{\partial\phi[v]_{(d:)}} = \psi[r]_{(0:d)} \odot \phi[w]_{(d:)} - \psi[r]_{(d:)} \odot \phi[w]_{(0:d)}$, $\frac{\partial\Gamma(v,r,w)}{\partial\phi[w]_{(0:d)}} = \psi[r]_{(0:d)} \odot \phi[v]_{(0:d)} - \psi[r]_{(d:)} \odot \phi[v]_{(d:)}$, $\frac{\partial\Gamma(v,r,w)}{\partial\phi[w]_{(d:)}} = \psi[r]_{(0:d)} \odot \phi[v]_{(d:)} + \psi[r]_{(d:)} \odot \phi[v]_{(0:d)}$. Concatenating gradients for the real part and the image part, we have the gradients

$$\nabla_{\phi[v]}\Gamma(v, r, w) = \frac{\partial\Gamma(v, r, w)}{\partial\phi[v]_{(0:d)}} \| \frac{\partial\Gamma(v, r, w)}{\partial\phi[v]_{(d:)}},$$

$$\nabla_{\phi[w]}\Gamma(v, r, w) = \frac{\partial\Gamma(v, r, w)}{\partial\phi[w]_{(0:d)}} \| \frac{\partial\Gamma(v, r, w)}{\partial\phi[w]_{(d:)}}.$$

**Complexity**

We can analyse the scalability of ReFactor GNNs along three axes, the number of layers $L$, the embedding size $d$, and the number of triplets/edges in the graph $|\mathcal{T}|$. For scalability w.r.t. to the number of layers, let $L$ denote the number of message-passing layers. Since ReFactor GNNs tie the weights across the layers, the parameter complexity of ReFactor GNNs is $\mathcal{O}(1)$, while it is $\mathcal{O}(L)$ for standard GNNs such as GATs, GCNs, and R-GCNs. Additionally, since ReFactor GNNs adopt layer-wise training enabled via the external memory for node state caching, the training memory footprint is also $\mathcal{O}(1)$ as opposed to $\mathcal{O}(L)$ for standard GNNs. For scalability w.r.t the embedding size, let $d$ denote the embedding size. REFACTOR GNNs scale linearly with $d$, as opposed to most GNNs in literature where the parameter and time complexities scale quadratically with $d$. For scalability w.r.t. the number of triplets/edges in the graph, we denote the entity set as $\mathcal{E}$, the relation set as $\mathcal{R}$, and the triplets as $\mathcal{T}$. NBFNet requires $O(LT^2d + L\mathcal{T}Vd^2)$ inference run-time complexity since the message-passing is done for every source node and query relation – quadratic w.r.t the number of triplets $\mathcal{T}$ while REFACTOR GNNs are of linear complexity w.r.t $\mathcal{T}$. Extending the complexity analysis in NBFNet [Zhu et al., 2021] to all the triplets, we include a detailed table for complexity comparison in Table B.1. The inference complexity refers to the cost per forward pass over the entire graph.

Table B.1: Complexity comparison across models. All expressions are asymptotic in big-O notation.

| Model | #Param | Train Mem. | Infer. Mem. | Train Time |
|---|---|---|---|---|
| GAT | $O(Ld^2)$ | $O(L|V|d)$ | $O(L|V|d)$ | $O(L|V|d^2 + L|T|d)$ |
| R-GCN | $O(L|R|d^2)$ | $O(L|V|d)$ | $O(L|V|d)$ | $O(L|T|d^2 + L|V|d^2)$ |
| NBFNet | $O(L|R|d^2)$ | $O(L|V||T|d)$ | $O(L|V||T|d)$ | $O(L|T|^2d + L|T||V|d^2)$ |
| REFACTOR GNNs | $O(|R|d)$ | $O(|V|d)$ | $O(L|V|d)$ | $O(|T||V|d)$ |

**Expressiveness of FMs, GNNs and REFACTOR GNNs**

We envision one interesting branch of future work would be a unified framework of expressiveness for all three model categories: FMs, GNNs and REFACTOR GNNs. To the best of our knowledge, there are currently two separate notions of expressiveness, one for FMs and the other for GNNs. While these two notions of expressiveness are both widely acclaimed within their own communities, it is unclear how to bridge them and produce a new tool supporting the analysis of the empirical applications (REFACTOR GNNs) that seam the two communities.

**Fully Expressiveness for Adjacency Recovery.** In the FM community, an FM is said to be *fully expressive* [Kazemi and Poole, 2018] if, for any given graph $\mathcal{T}$ over entities $\mathcal{E}$ and relations $\mathcal{R}$, it can fully reconstruct the input adjacency tensor with an embedding size bounded by $\min(|\mathcal{E}||\mathcal{R}|, |\mathcal{T}| + 1)$. We can generalise this expressiveness analysis to the spectrum of FM-GNN models (REFACTOR GNNs). In the $L \to \infty$ limit, REFACTOR GNNs are as fully expressive as the underlying FMs. In fact, a REFACTOR GNN based on DistMult [Yang et al., 2015b] is not fully expressive (because of its symmetry); however a REFACTOR GNN based, e.g. on ComplEx [Trouillon et al., 2016, Lacroix et al., 2018] can reach full expressiveness for $L \to \infty$.

**Weisfeiler-Leman Tests for Nodes/Graphs Separation.** For GNNs, established theoretical results on expressiveness mainly focus on the separation power of induced representations in terms of Weisfeiler-Leman isomorphism tests [Xu et al., 2019, Geerts and Reutter, 2021]. However, none of these results is directly applicable to our setting (e.g. they only consider one relationship). Nevertheless, if we consider our REFACTOR GNNs in a one-relationship, simple graph setting, following the formalism of Geerts and Reutter [2021], we note that the REFACTOR Layer function cannot be written in Guarded Tensor Language since at each layer it computes a global term $n[v]$. Moreover, REFACTOR GNNs only process information coming from two nodes at one time. These two facts imply that REFACTOR GNNs have a separation power upper bound comparable to the 1-WL test, i.e. comparable to 1-MPNN (not guarded).

We are not aware of explicit connections between the two above notions of expressiveness. We think there is some possibility that we can bridge them, which itself will be a very interesting research direction, but would require a very substantial amount of additional work and presentation space and is thus beyond the scope of this chapter.

Alternatively, we can also increase the expressiveness of REFACTOR GNNs by adding more parameters to the message, aggregation and update operators. For example, introducing additional MLPs to transform the input node features or include non-linearity in the GNN update operator. This would be a natural way to increase the expressiveness of REFACTOR GNNs.

Another method for increasing expressive power for link prediction task only is to extend ReFactor GNNs from node-wise to pair-wise (Sec 4.3.2) representations like GraIL [Teru et al., 2020] and NBFNet [Zhu et al., 2021], which is more computationally intensive, but yields more powerful as node representations are not standalone but adapted to a specific query.

## B.1.2 Experimental setup, Hyperparameters, and Implementation

As we stated in the experiments section, we used a two-stage training process. In stage one, we sample subgraphs around query links and serialise them. In stage two, we load the serialised subgraphs and train the GNNs. For transductive knowledge graph completion, we test the model on the same graph (but different splits). For inductive knowledge graph completion, we test the model on the new graph, where the relation vocabulary is shared with the training graph, while the entities are novel. We use the validation split for selecting the best hyperparameter configuration and report the corresponding test performance. We include reciprocal triplets into the training triplets following standard practice [Lacroix et al., 2018].

For subgraph serialisation, we first sample a mini-batch of triplets and then use these nodes as seed nodes for sampling subgraphs. We also randomly draw a node globally and add it to the seed nodes. The training batch size is 256 while the valid/test batch size is 8. We use the LADIES algorithm [Zou et al., 2019] and sample subgraphs with depths in $[1, 2, 3, 6, 9]$ and a width of 256. For each graph, we keep sampling for 20 epochs, i.e. roughly 20 full passes over the graph.

For general model training, we consider hyperparameters including learning rates in $[0.01, 0.001]$, weight decay values in $[0, 0.1, 0.01]$, and dropout values in $[0, 0.5]$. For GATs, we use 768 as the hidden size and 8 as the number of attention heads. We train GATs with 3 layers and 6 layers. We also consider whether to combine the outputs from all the layers. For REFACTOR GNNs, we use the same hidden size as GAT. We consider whether the ReFactor Layer is induced by an SGD operator or by a AdaGrad operator. Within a ReFactor Layer, we also consider the N3 regulariser strength values $[0, 0.005, 0.0005]$, the $\alpha$ values $[0.1, 0.01]$, and the option of removing the $n[v]$, where the message-passing layer only involves information flow within 1-hop neighbourhood as most the classic message-passing GNNs do.

We use grid search to find the best hyperparameter configuration based on the validation MRR. Each training run is done using two Tesla V100 (16 GB) GPUs with, where data parallelism was implemented via the *DistributedDataParallel* component of *pytorch-lightning*. For inductive learning experiments, inference for all the validation and test queries on small datasets like FB15K237_v1 takes about 1-5 seconds, while on medium datasets it takes approximately 20 seconds, and on big datasets like

WN18RR_v4 it requires approximately 60 seconds. For most training runs, the memory footprint is less than 40% (13 GB). The training time for 20 full passes over the graph is about 1, 7, and 21 minutes respectively for small, medium, and large datasets.

Our code will be available at ReFactorGNN. We adapted the LADIES subgraph sampler from the GPT-GNN codebase [Hu et al., 2020] for sampling on knowledge graphs. The datasets we used can be downloaded from the repositories Datasets for Knowledge Graph Completion with Textual Information about Entities and GraIL - Graph Inductive Learning. We implemented REFACTOR GNNs using the *MessagePassing* API in *PyTorch Geometric*. Specially, we used *message_and_aggregate* function to compute the aggregated messages.

## B.2 Additional results

### Additional Results on Inductive KGC Tasks

In this chapter, we describe the results on FB15K237_v1_ind under some random seed. To confirm the significance and sensitivity, we further experiment with additional 5 random seeds. Due to our computational budget, for this experiment, we resorted to a coarse grid when performing the hyperparameters sweeps. Following standard evaluation protocols, we report the mean values and standard deviations of the filtered Hits@10 over 5 random seeds. Numbers for Neural-LP, DRUM, RuleN, GraIL, and NBFNet are taken from the literature [Teru et al., 2020, Zhu et al., 2021]. "-" means the numbers are not applicable. Table B.3 summarises the results. REFACTOR GNNs are able to make use of both types of input features, while textual features benefit both GAT and REFACTOR GNNs for most datasets. Increasing depth benefits WN18RR_v$i$_ind ($i \in [1, 2, 3, 4]$) most. Future work could consider the impact of textual node features provided by different pretrained language models. Another interesting direction is to investigate the impact of depth on GNNs for datasets like WN18RR, where many kinds of hierarchies are observed in the data.

In addition to the *partial ranking* evaluation protocol, where the ground-truth subject/object entity is ranked against 50 sampled entities,[1] we also consider the *full ranking* evaluation protocol, where the ground-truth subject/object entity is ranked against all the entities. Table B.4 summarises the results. Empirically, we observe that *full ranking* is more suitable for reflecting the differences between models than *partial ranking*. It also has less variance than *partial ranking*, since it requires no sampling from the candidate entities. Hence, we believe there is good reason to recommend the community to use *full ranking* for these datasets in the future.

---

[1] One implementation for such evaluation can be found in GraIL's codebase.

Table B.2: The impact of meaningful node features on *FB15K237_v1*. $\Delta$ Test MRR is computed as `test MRR (textual node features)` $-$ `test MRR (random node features)`. Larger $\Delta$ means meaningful node features bring more benefit.

| Depth | 3 | 6 | $\infty$ |
|---|---|---|---|
| $\Delta$ Test MRR | 0.060 | 0.045 | 0.016 |

## Additional Results on The Impact of Meaningful Node Features

To better understand the impact that meaningful node features have on REFACTOR GNNs for the task of knowledge graph completion, we compare REFACTOR GNNs trained with RoBERTa Encodings (one example of meaningful node features) and REFACTOR GNNs trained with Random Vectors (not meaningful node features). We perform experiments on *FB15K237_v1* and vary the number of message-passing layers: $L \in \{3, 6, \infty\}$. Table B.2 summarises the differences. We can see that meaningful node features are highly beneficial if REFACTOR GNNs are only provided with a few message-passing layers. As more message-passing layers are allowed, the benefit of REFACTOR GNNs diminishes. The extreme case would be $L = \infty$, where the benefit of meaningful node features becomes negligible. We hypothesise that this might be why meaningful node features have not been found to be useful for transductive knowledge graph completion.

## Additional Results on Parameter Efficiency

Figure B.1 shows the parameter efficiency on the dataset *FB15K237_v2*.

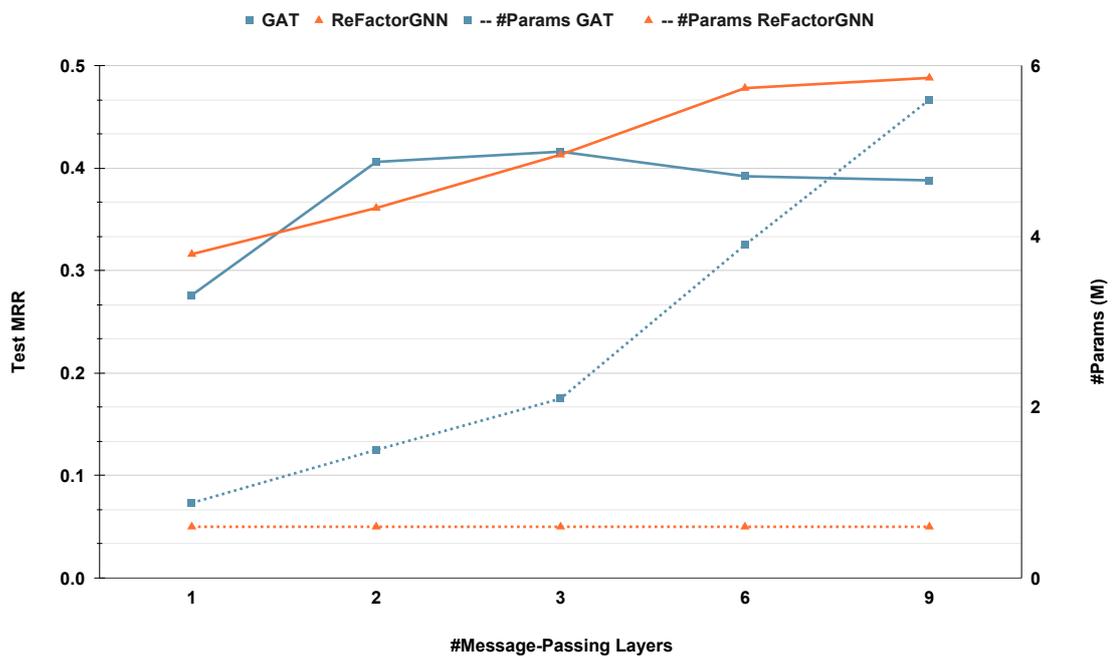

Figure B.1: Performance vs parameter efficiency on *FB15K237_v2*. The left axis is Test MRR while the right axis is #Parameters. The solid lines and dashed lines indicate the changes of Test MRR and the changes of #Parameters.

Table B.3: Hits@10 with Partial Ranking against 50 Negative Samples. "[T]" indicates using textual encodings of entity descriptions [Reimers and Gurevych, 2019] as input (positional) node features; "[R]" indicates using frozen random vectors as input (positional) node feature.

| | WN18RR | | | | FB15k-237 | | | | NELL-995 | | | |
|---|---|---|---|---|---|---|---|---|---|---|---|---|
| | v1 | v2 | v3 | v4 | v1 | v2 | v3 | v4 | v1 | v2 | v3 | v4 |
| No Pretrain [R] | 0.220±0.048 | 0.226±0.013 | 0.244±0.020 | 0.218±0.050 | 0.215±0.019 | 0.207±0.008 | 0.211±0.002 | 0.205±0.008 | 0.543±0.022 | 0.207±0.008 | 0.216±0.004 | 0.198±0.006 |
| No Pretrain [T] | 0.267±0.020 | 0.236±0.020 | 0.292±0.025 | 0.253±0.022 | 0.242±0.018 | 0.227±0.007 | 0.240±0.011 | 0.244±0.003 | 0.538±0.079 | 0.234±0.017 | 0.242±0.020 | 0.191±0.036 |
| Neural-LP | 0.744 | 0.689 | 0.462 | 0.671 | 0.529 | 0.589 | 0.529 | 0.559 | 0.408 | 0.787 | 0.827 | 0.806 |
| DRUM | 0.744 | 0.689 | 0.462 | 0.671 | 0.529 | 0.587 | 0.529 | 0.559 | 0.194 | 0.786 | 0.827 | 0.806 |
| RuleN | 0.809 | 0.782 | 0.534 | 0.716 | 0.498 | 0.778 | 0.877 | 0.856 | 0.535 | 0.818 | 0.773 | 0.614 |
| GAT(3) [R] | 0.583±0.022 | 0.797±0.002 | 0.560±0.005 | 0.660±0.015 | 0.333±0.042 | 0.312±0.036 | 0.407±0.072 | 0.363±0.050 | 0.906±0.004 | 0.303±0.031 | 0.351±0.009 | 0.187±0.098 |
| GAT(6) [R] | 0.850±0.014 | 0.841±0.001 | 0.631±0.020 | 0.802±0.004 | 0.401±0.020 | 0.445±0.018 | 0.461±0.048 | 0.406±0.143 | 0.811±0.039 | 0.670±0.055 | 0.341±0.042 | 0.301±0.002 |
| GAT(3) [T] | 0.970±0.002 | 0.980±0.001 | 0.897±0.005 | 0.960±0.001 | 0.806±0.003 | 0.942±0.001 | 0.941±0.002 | 0.954±0.001 | 0.938±0.005 | 0.839±0.005 | 0.962±0.001 | 0.354±0.002 |
| GAT(6) [T] | 0.965±0.002 | 0.986±0.001 | 0.920±0.002 | 0.970±0.003 | 0.826±0.004 | 0.943±0.001 | 0.927±0.003 | 0.927±0.001 | 0.904±0.000 | 0.811±0.001 | 0.880±0.001 | 0.297±0.003 |
| GraIL | 0.825 | 0.787 | 0.584 | 0.734 | 0.642 | 0.818 | 0.828 | 0.893 | 0.595 | 0.933 | 0.914 | 0.732 |
| NBFNet | 0.948 | 0.905 | 0.893 | 0.890 | 0.834 | 0.949 | 0.951 | 0.960 | - | - | - | - |
| ReFactorGNN(3) [R] | 0.899±0.003 | 0.842±0.004 | 0.605±0.000 | 0.801±0.002 | 0.673±0.000 | 0.812±0.002 | 0.833±0.003 | 0.877±0.002 | 0.913±0.000 | 0.913±0.011 | 0.893±0.000 | 0.838±0.002 |
| ReFactorGNN(6) [R] | 0.885±0.000 | 0.854±0.003 | 0.738±0.006 | 0.817±0.004 | 0.787±0.007 | 0.903±0.003 | 0.903±0.002 | 0.920±0.002 | 0.971±0.007 | 0.957±0.003 | 0.935±0.003 | 0.927±0.001 |
| ReFactorGNN(3) [T] | 0.918±0.002 | 0.973±0.001 | 0.910±0.003 | 0.934±0.001 | 0.900±0.004 | 0.959±0.001 | 0.952±0.002 | 0.968±0.001 | **0.955±0.004** | 0.931±0.001 | 0.978±0.001 | 0.929±0.001 |
| ReFactorGNN(6) [T] | **0.970±0.002** | **0.988±0.001** | **0.944±0.002** | **0.987±0.000** | **0.920±0.001** | **0.963±0.001** | **0.962±0.002** | **0.970±0.002** | 0.949±0.011 | **0.963±0.001** | **0.994±0.000** | **0.955±0.002** |

Table B.4: Hits@10 with Full Ranking against All Candidate Entities. "[T]" indicates using textual encodings of entity descriptions [Reimers and Gurevych, 2019] as input (positional) node features; "[R]" indicates using frozen random vectors as input (positional) node feature.

| | WN18RR | | | | FB15k-237 | | | | NELL-995 | | | |
|---|---|---|---|---|---|---|---|---|---|---|---|---|
| | v1 | v2 | v3 | v4 | v1 | v2 | v3 | v4 | v1 | v2 | v3 | v4 |
| No Pretrain [R] | 0.020±0.006 | 0.004±0.001 | 0.004±0.003 | 0.003±0.001 | 0.013±0.003 | 0.012±0.001 | 0.004±0.001 | 0.002±0.001 | 0.255±0.021 | 0.004±0.001 | 0.001±0.001 | 0.003±0.001 |
| No Pretrain [T] | 0.027±0.009 | 0.007±0.003 | 0.006±0.001 | 0.005±0.001 | 0.014±0.001 | 0.010±0.001 | 0.007±0.001 | 0.006±0.001 | 0.262±0.031 | 0.006±0.002 | 0.006±0.002 | 0.003±0.001 |
| GAT(3) [R] | 0.171±0.008 | 0.504±0.026 | 0.260±0.022 | 0.089±0.017 | 0.074±0.003 | 0.050±0.014 | 0.051±0.019 | 0.023±0.012 | 0.806±0.019 | 0.003±0.002 | 0.008±0.007 | 0.008±0.004 |
| GAT(6) [R] | 0.575±0.005 | 0.698±0.003 | 0.312±0.000 | 0.606±0.002 | 0.048±0.004 | 0.028±0.004 | 0.033±0.018 | 0.015±0.026 | 0.491±0.112 | 0.110±0.048 | 0.031±0.010 | 0.031±0.002 |
| GAT(3) [T] | 0.794±0.000 | 0.826±0.000 | 0.468±0.000 | 0.705±0.000 | 0.331±0.000 | 0.585±0.000 | 0.505±0.000 | 0.449±0.000 | 0.856±0.000 | 0.245±0.000 | 0.345±0.000 | 0.078±0.000 |
| GAT(6) [T] | 0.815±0.000 | 0.808±0.000 | 0.469±0.000 | 0.701±0.000 | 0.416±0.000 | 0.483±0.000 | 0.391±0.000 | 0.388±0.000 | 0.851±0.000 | 0.189±0.000 | 0.137±0.000 | 0.023±0.000 |
| ReFactorGNN(3) [R] | 0.826±0.000 | 0.758±0.002 | 0.374±0.000 | 0.707±0.000 | 0.455±0.010 | 0.603±0.008 | 0.556±0.003 | 0.587±0.003 | 0.907±0.004 | 0.700±0.001 | 0.630±0.001 | 0.511±0.001 |
| ReFactorGNN(6) [R] | 0.826±0.001 | 0.769±0.005 | 0.440±0.001 | 0.731±0.000 | 0.558±0.007 | 0.694±0.006 | 0.639±0.006 | 0.640±0.000 | **0.967±0.005** | **0.764±0.009** | 0.697±0.005 | **0.703±0.001** |
| ReFactorGNN(3) [T] | 0.805±0.000 | 0.796±0.003 | 0.483±0.000 | 0.682±0.000 | 0.589±0.001 | 0.672±0.001 | 0.610±0.001 | 0.611±0.001 | 0.918±0.000 | 0.629±0.001 | 0.634±0.000 | 0.305±0.000 |
| ReFactorGNN(6) [T] | **0.844±0.004** | **0.848±0.003** | **0.522±0.001** | **0.781±0.001** | **0.619±0.000** | **0.721±0.001** | **0.663±0.000** | **0.660±0.000** | 0.913±0.000 | 0.733±0.000 | **0.711±0.000** | 0.417±0.000 |

# Appendix C

# Language Model Plasticity

Here we provide additional technical details and results for *Chapter 4 Improving Language Plasticity via Pretraining with Active Forgetting*.

## C.1 Technical Details

### C.1.1 Low-data Experimental Regime

A common experimental setup for adapting to a target language is to use all the available data in that language from sources such as Wikipedia [Artetxe et al., 2020, Ansell et al., 2022] and CC100 [Marchisio et al., 2023]. In this setup, the numbers of tokens typically used for adapting each language might differ greatly, ranging from 13.9M to 70.5B, as summarised via Table C.1.

Our work, however, investigates a different setup where we control the adaptation data to 5 million tokens for each language. This can be highly relevant when studying generalisation to completely new languages, which require expanding the vocabulary. We acknowledge that dealing with real-world low-resources languages can be more challenging than such low-data setup used. And there are already rich work addressing low-resource issues: multilingual pretraining [Conneau et al., 2020, Pfeiffer et al., 2022], multilingual adapters [Pfeiffer et al., 2020, Ansell et al., 2022], multilingual adaptation [Tang et al., 2020, Alabi et al., 2022], and multilingual regularisation [Pfeiffer et al., 2021]. Nevertheless, we would like to highlight the importance of our low-data regime. The challenge of "low-resource" involve multiple *entangled factors*: the quality of the tokeniser, the amount of data, whether the script/language family is distant to the pretraining language(s) etc. Simulating a low-data regime allows us to control these factors and isolate the effects of the factor that we are interested in – the amount of data in the new language. This factor is essential to our work as our research goal is plasticity i.e. rewiring model prediction with as little new information as possible. Simulating various amount of data in the new language allows us to compare model plasticity as shown in Figure 5.3, and thus contribute a clean piece of knowledge in the line of plasticity research [Lyle et al., 2023, Nikishin et al., 2023].

### C.1.2 On the Experimental Framework Choice for Studying Pretrained Language Model Plasticity

Our motivation is to improve language models' plasticity. Plasticity of neural networks have been studied in graph learning, computer vision and reinforcement learning [Taha et al., 2021, Chen et al., 2022, Lyle et al., 2023, Nikishin et al., 2023], where forgetting-relearn methods show promise. Our goal is to study plasticity in the context of pretrained language models. We believe this is n emerging research direction and will thrive in the following years. However, translating the plasticity concept to the language model setting is not trivial due to the lack of clear experimental setups. We note that, despite the model differences, almost all language models begins with a token embedding layer. As often tied to a specific vocabulary, the token embedding layer limits the plasticity, preventing generalisation to a new vocabulary. This observation inspires us to explore the plasticity of language models by manipulating the token embedding layer. Artetxe et al. [2020] draws our attention as it offers a nice experimental framework of only manipulating the token embedding layer for adapting between languages.

## C.2 Additional Results

### Results on Distant Languages Benefits More From Forgetting

We are interested in how forgetting PLMs can improve adaptation to different languages. We compare the results of various languages on three benchmarks: XNLI, MLQA and XQuAD. We use Figure 5.6, Figure 5.7, and Figure C.1 to illustrate the relative gains from active forgetting on each benchmark. We find that languages that are less related to the pretraining language, which in this case is English, benefit more from forgetting PLMs.

Table C.1: Languages by family, script, and morphology.

| Name | Code | Family | Script | Morphology |
|---|---|---|---|---|
| Arabic | ar | Semitic | Arabic (Abjad) | Introflexive |
| Bulgaria | bg | IE:Balto-Slavic | Cyrillic | Analytic |
| German | de | IE:Germanic | Latin | Fusional |
| Greek | el | IE:Hellenic | Greek | Fusional |
| English | en | IE:Germanic | Latin | Analytic |
| French | fr | IE:Romance | Latin | Fusional |
| Hindi | hi | IE:Indo-Iranian | Devanagari | Fusional |
| Russian | ru | IE:Balto-Slavic | Cyrillic | Fusional |
| Spanish | es | IE:Romance | Latin | Fusional |
| Swahili | sw | Niger-Congo:Bantu | Latin | Agglutinative |
| Thai | th | Tai-Kadai | Thai | Analytic |
| Turkish | tr | Turkic | Latin | Agglutinative |
| Urdu | ur | IE:Indo-Iranian | Perso-Arabic | Fusional |
| Vietnamese | vi | Austroasiatic | Latin | Analytic |
| Chinese | zh | Sino-Tibetan | Chinese | Analytic |

### Results on Forgetting PLMs Converge Faster

Figure C.2 displays the adaptation curves for several languages (Arabic, German, Hindi, Thai, Turkish, and Urdu) during full training runs of 125,000 steps. This complements Figure 5.8, which focuses on the first 5,000 steps. Similar convergence patterns can be observed for additional languages, as shown in Figure C.3 and Figure C.4.

### Impact of Forgetting Frequency

We would like to elaborate on our choice of forgetting frequency $K$. In our preliminary experiments, we tried $K = 100, 1000, 5000$. We find $K = 1000$ works well and thus sticks with it. Since we don't want to over tune the hyperparameters, we just use the same $K$ for all the experiments. We include the loss curves of $K = 100$ and $K = 5000$ in Figure C.5. We can see that both forgetting too frequently and forgetting too infrequently will hurt the performance. Too frequent forgetting leaves little time for the body to learn something meaningful (the pretraining loss stuck around 11). Too sparse forgetting will make the body hard to adjust to the next forgetting, causing divergence as pretraining goes on.

Table C.2: Numbers of tokens for different languages on the two multilingual corpus, CC-100 and Wikipedia, in ascending order of CC100. The English one is used as pretraining corpus while the others are used for language adaptation.

| Language | CC-100 Tokens | Wikipedia Tokens |
| --- | --- | --- |
| sw | 345M | 13.9M |
| ur | 832M | 41.5M |
| hi | 2.13B | 54.6M |
| ar | 4.15B | 337M |
| tr | 4.19B | 157M |
| th | 6.09B | 70.7M |
| el | 6.10B | 148M |
| bg | 7.90B | 134M |
| zh | 9.72B | 584M |
| es | 11.6B | 1.17B |
| fr | 13.3B | 1.71B |
| de | 14.1B | 2B |
| vi | 28.9B | 300M |
| ru | 34.9B | 1.25B |
| **en** | **70.5B** | **4.25B** |

### Multilingual Pretraining and Forgetting Pretraining

Our work aim to have a flexible language model by pretraining with forgetting. No matter the pretraining corpus is monolingual or multilingual, this language model should easily generalise itself to unseen languages. This is different from the scenario of multilingual PLMs like XLM-R [Conneau et al., 2020], which requires seeing all the data for all languages from the scratch. Once done with pretraining and there is some new language dis-tant from the pretraining languages to support, the multilingual PLMs might still struggle with zero-shot transfer as shown in several low-resources language research [Ebrahimi et al., 2022, Adelani et al., 2021, 2022].

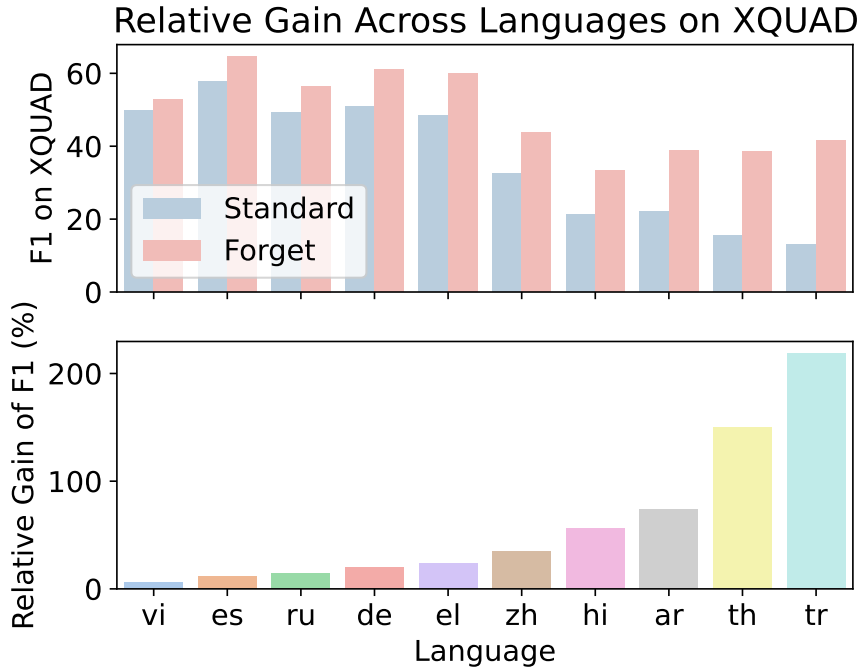

Figure C.1: Relative gains of forgetting PLMs over standard PLMs across languages on XQuAD. Languages that are less related to the pretraining language (English), such as Turkish, Thai, Arabic, Hindi, benefit more from forgetting PLMs.

Nevertheless, we ran additional experiments on multilingual pretraining with forgetting. The results are summarised in Table C.3. For a fair comparison, we trained a multilingual RoBERTa-base of the same model size as our monolingual model. Language Emb/Task Body Adaptation refers to separately adapting embeddings with 5M tokens of Thai unlabelled data and adapting body with English NLI data. Task Full Model Adaptation refers to adapting the full model with English NLI data. Note that Thai is already included in multilingual CC100 (6B tokens in the original dataset, 720M tokens in our subsampled dataset). We measure the zero-shot Thai XNLI Accuracy. We can see that multilingual pretraining indeed helps cross-lingual transfer when the language is in the pretraining data. On the other hand, we can also observe that forgetting indeed lifts the adaptation performance:

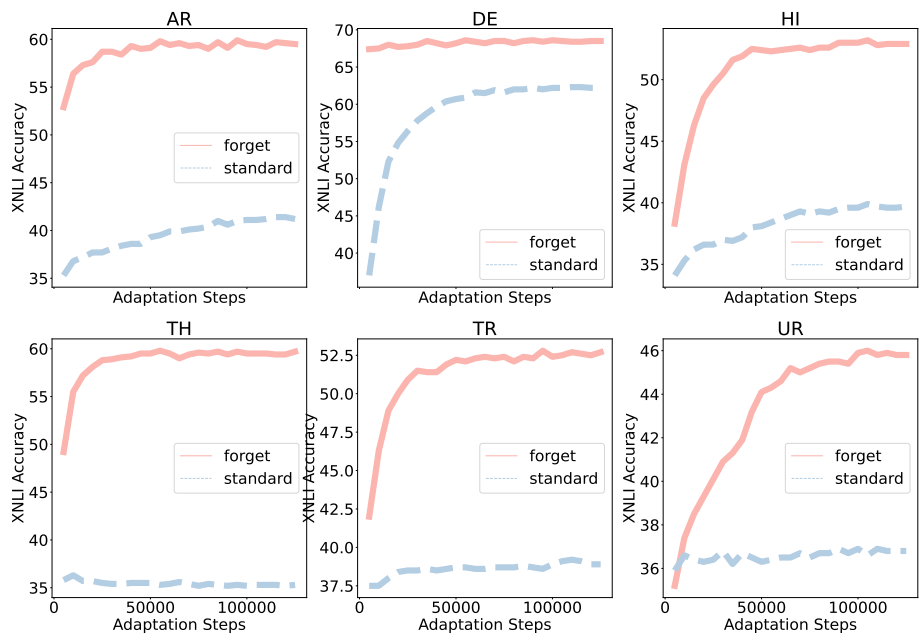

Figure C.2: Adaptation curves on XNLI for individual languages: Arabic, German, Hindi, Thai, Turkish, and Urdu. Forgetting helps more languages that are distant to English (the pretraining language).

- Comparing Row 3 and Row 4 (49.4 vs 55.0) in Table C.3, we can see that, forgetting also helps adapt multilingual pretrained models.

- Comparing Row 1 and Row 2 (35.3 vs 59.7) in Table C.3, we can see that forgetting helps monolingual pretrained models a lot.

- XLM-R (base) outperform best our multilingual pretrained baselines (72.4 vs 60.0). This is no surprise due to its large pretraining corpus (10x our multilingual corpora) and model size (2x our multilingual model).

## Full Model Task Adaptation and Forgetting

The language/task adaptation does not use any labelled data. It only uses the unlabelled data from the new language. In contrast, Standard adaptation relies on labelled data,

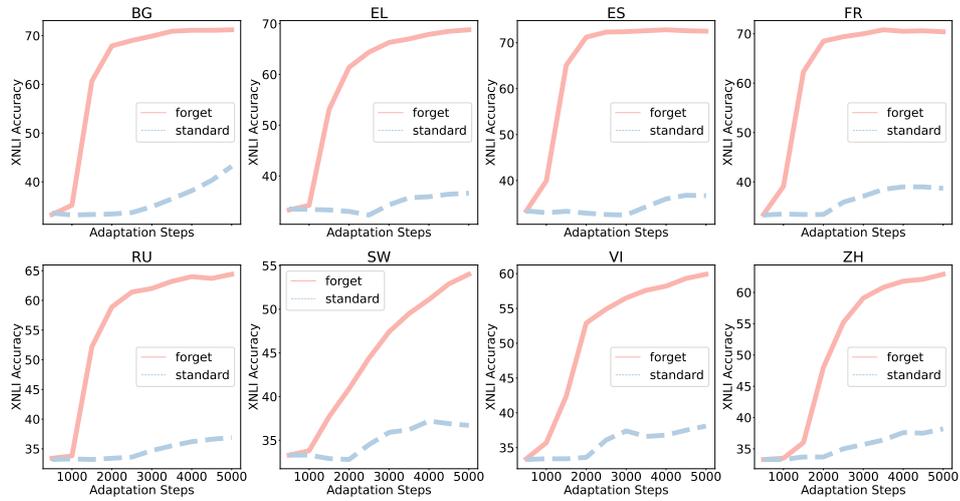

Figure C.3: Adaptation curves on XNLI within 5K updates for individual languages: Bulgaria, Greek, Spanish, French, Russian, Swahili, Vietnamese and Chinese. Forgetting PLMs converge faster than standard PLMs.

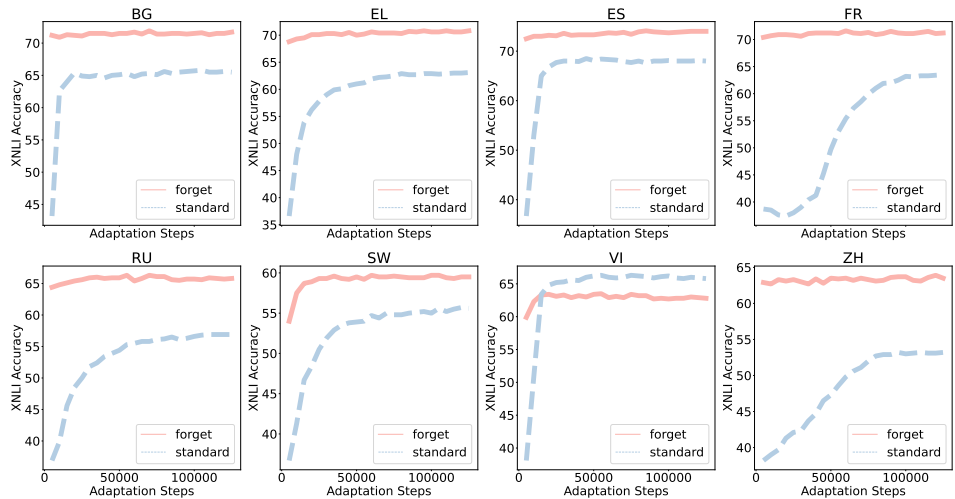

Figure C.4: Adaptation curves on XNLI for individual languages: Bulgaria, Greek, Spanish, French, Russian, Swahili, Vietnamese and Chinese. Across all the languages except on Vietnamese, the forgetting PLMs reach a better performance level than their standard counterparts.

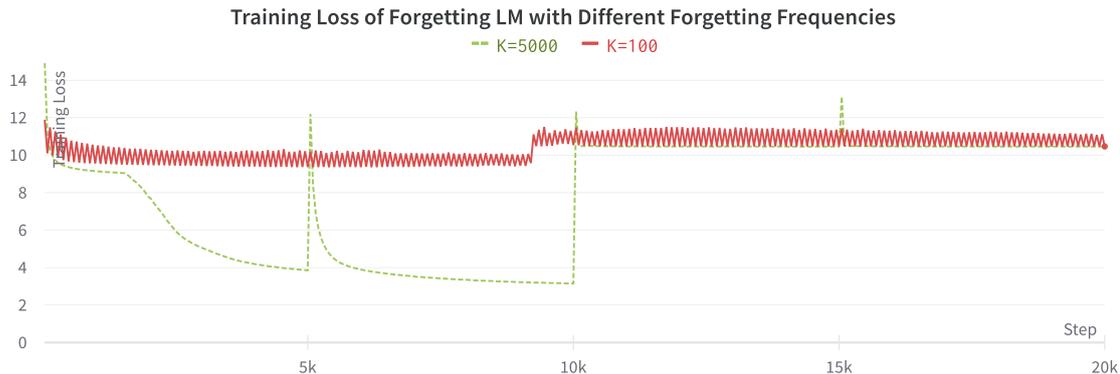

Figure C.5: Impact of Forgetting Frequency.

Table C.3: Comparing various ways to make models multilingual. The pretraining architectures are all RoBERTa (base). Standard refers to standard monolingual pretraining. Forget refers to active forgetting based monolingual pretraining. XLM refers to multilingual pretraining like the XLM work [Conneau et al., 2020]

| **Pretrain Corpus** | **#Lang** | **#Param** | **Pretrain** | **Adaptation Method** | **Acc** |
|---|---|---|---|---|---|
| 300GB English | 1 | 125M | Standard | Lang Emb/Task Body | 35.3 |
| 300GB English | 1 | 125M | Forget | Lang Emb/Task Body | 59.7 |
| 50GB Multilingual | 100 | 125M | Standard | Lang Emb/Task Body | 49.4 |
| 50GB Multilingual | 100 | 125M | Forget | Lang Emb/Task Body | 55.0 |
| 50GB Multilingual | 100 | 125M | Standard | Task Full Model | 60.0 |
| 2.5TB Multilingual | 100 | 270M | XLM | Task Full Model | 72.4 |

which is expensive for a new downstream language.

Here we present an additional experiment to compare full model adaptation and partial model adaptation. Our experimental setup follows Artetxe et al. [2020], where standard-pretraining + language/task adaptation (MonoTrans) is shown to be competitive among a few baselines for zero-shot unsupervised cross-lingual transfer. Results in Table C.4 verify this. Our proposed forgetting method can further improve the sample-efficiency of the language/task adaptation, surviving a low amount of unsupervised data in the new language. This is motivated by a practical scenario where the new languages contain only several thousands of tokens to a few millions of tokens (e.g. the corpus for the new language might contain only 2-3 books).

Table C.4: Full model adaptation vs. partial model adaptation. **Sup.** refers to the amount of supervised data (tokens), and **Unsup.** refers to the amount of unsupervised data used.

| Method | Sup. | Unsup. | Acc |
| --- | --- | --- | --- |
| standard pretraining + standard adaptation | 6.7K | 0 | 32.8 |
| standard pretraining + language/task adaptation | 0 | 5M | 41.2 |
| forget pretraining + standard adaptation | 6.7K | 0 | 34.2 |
| forget pretraining + language/task adaptation | 0 | 5M | 59.7 |

## Impact of Adaptation Data Amount

Evidence of high sample efficiency can be found by comparing the performance drop of standard PLMs and forgetting PLMs when the adaptation data change from a high-data setting [Artetxe et al., 2020, Marchisio et al., 2023] to a low-data setting that our work is considering, as shown in Table C.5.

Table C.5: Comparison of adaptation data amount.

| Method | Avg adaptation #tokens | Avg XNLI Acc |
| --- | --- | --- |
| Standard [Marchisio et al., 2023] | 10.3B | 72.0 |
| Standard [Artetxe et al., 2020] | 569M | 66.7 |
| Standard | 5M | 53.3 |
| Forgetting | 5M | 62.7 |

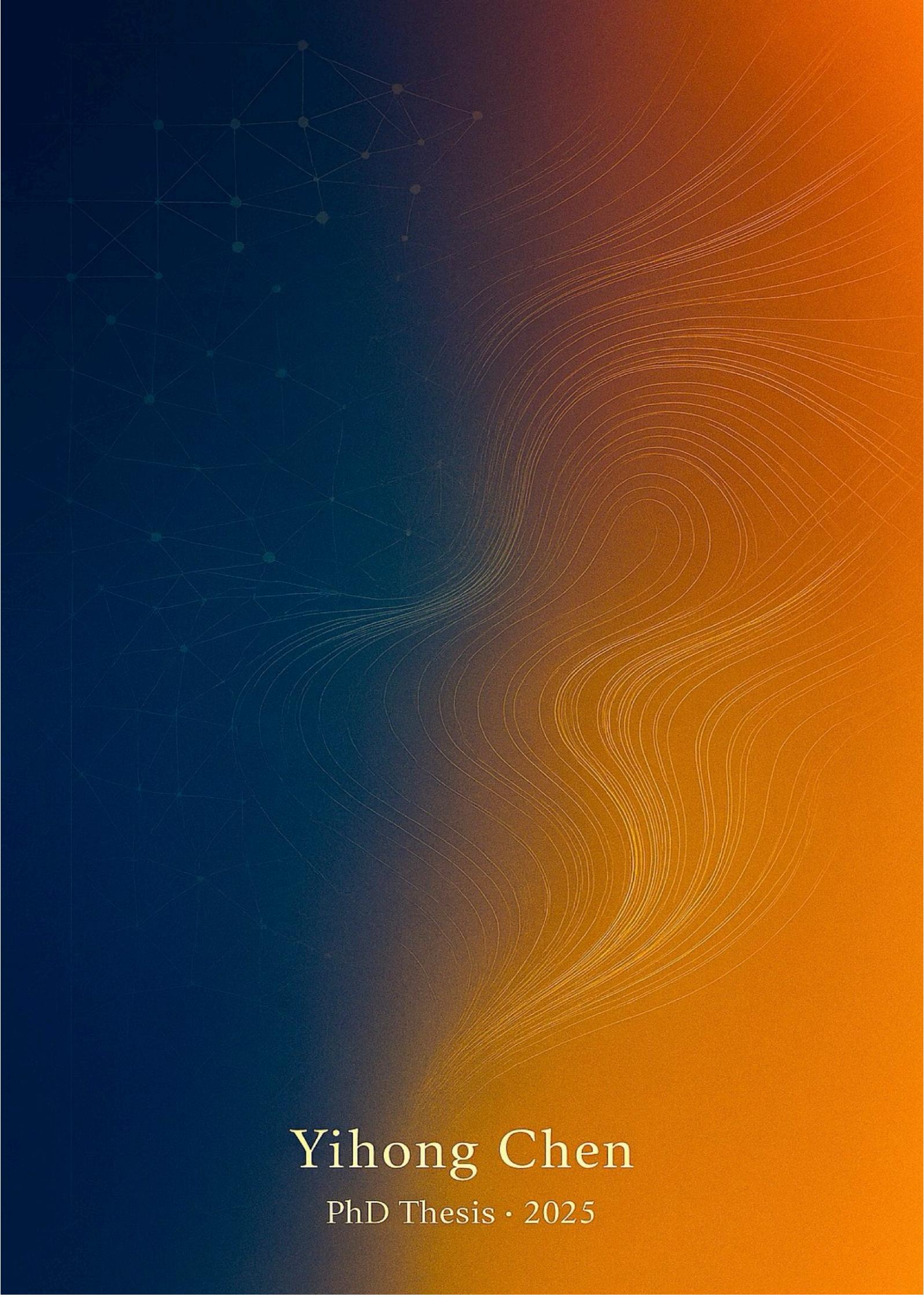
Yihong Chen
PhD Thesis · 2025